\documentclass[10pt,twocolumn,letterpaper]{article}

\usepackage[pagenumbers]{cvpr} 

\usepackage{tikz}

\definecolor{cvprblue}{rgb}{0.21,0.49,0.74}
\usepackage[pagebackref,breaklinks,colorlinks,citecolor=cvprblue]{hyperref}

\def\paperID{183} 
\def\confName{3DV\xspace}
\def\confYear{2024\xspace}

\title{Physics-Based Rigid Body Object Tracking and\\ Friction Filtering From RGB-D Videos}

\author{Rama Krishna Kandukuri \hspace{8ex} Michael Strecke \hspace{8ex} Joerg Stueckler\\
Embodied Vision Group, Max Planck Institute for Intelligent Systems, Tuebingen\\
{\tt\small \{rama.kandukuri,michael.strecke,joerg.stueckler\}@tue.mpg.de}
}

\newcommand\copyrighttext{%
	\footnotesize \textcopyright 2024 IEEE. Personal use of this material is permitted.
	Permission from IEEE must be obtained for all other uses, in any current or future
	media, including reprinting/republishing this material for advertising or promotional
	purposes, creating new collective works, for resale or redistribution to servers or
	lists, or reuse of any copyrighted component of this work in other works.
    Accepted for publication in 2024 International Conference on 3D Vision (3DV)
 DOI: 10.1109/3DV62453.2024.00111.
 
}
\newcommand\copyrightnotice{%
	\begin{tikzpicture}[remember picture,overlay]
		\node[anchor=south west,yshift=10pt,xshift=1.6cm] at (current page.south west) {\parbox{\textwidth}{\copyrighttext}};
	\end{tikzpicture}%
}

\begin{document}
\maketitle
\vspace*{-5ex}\copyrightnotice
\thispagestyle{empty}

\begin{abstract}

Physics-based understanding of object interactions from sensory observations is an essential capability in augmented reality and robotics.
It enables to capture the properties of a scene for simulation and control. 
In this paper, we propose a novel approach for real-to-sim which tracks rigid objects in 3D from RGB-D images and infers physical properties of the objects.
We use a differentiable physics simulation as state-transition model in an Extended Kalman Filter which can model contact and friction for arbitrary mesh-based shapes and in this way estimate physically plausible trajectories. 
We demonstrate that our approach can filter position, orientation, velocities, and concurrently can estimate the coefficient of friction of the objects.
We analyze our approach on various sliding scenarios in synthetic image sequences of single objects and colliding objects. 
We also demonstrate and evaluate our approach on a real-world dataset.
We make our novel benchmark datasets publicly available to foster future research in this novel problem setting and comparison with our method. 

\end{abstract}

\section{Introduction}
\label{sec:intro}
Many tasks in augmented reality and robotics require 3D scene understanding on the level of objects.
Examples are recognizing and synthesizing human interactions with objects or robotic object manipulation.
Besides the reconstruction of object shape and pose (position and orientation) over time, the estimation of physical object properties can be leveraged for simulation and planning in the perceived scene representations. 

Research on object detection and 3D pose estimation in images has achieved significant progress in recent years. 
Several works such as~\cite{gao2003_complete,lepetit2008_epnp,xiang2018_posecnn,he2020_pvn3d,he2021_ffb6d,labbe2020_cosypose,li2018_deepim,lipson2022_coupled} estimate poses of object instances with known shape and texture, while some approaches also estimate category-level object pose and shape from RGB-D images~\cite{wang2020_directshape,deng2022_icaps,irshad2022_shapo}.
To model and perceive object dynamics and their interactions, in previous works, either model-free or model-based methods have been developed.
Model-free methods (for example,~\cite{hafner2019_planet,becker2019_rkn,sanchez-gonzalez2020_gnnphys}) train deep neural network to predict future states from input states or image sequences.
This requires significant amounts of training data and the network needs to be retrained for novel scenes.
Model-based methods~\cite{jongeneel2022_model,murthy2021_gradsim} use analytical models of the object dynamics for filtering or identifying the parameters by fitting the dynamics model to trajectories.
Typically, however, the models only consider limited physical settings, have restrictions on the shapes that can be presented (e.g. only shape primitives), or do not support probabilistic state estimation.
Hybrid model- and learning-based approaches like~\cite{wu2015_galileo,kandukuri2021_physical} impose structure on the learning-based components to increase sample efficiency and generalization for the learning-based component. 

\begin{figure}
    \centering
    \includegraphics[width=.99\linewidth]{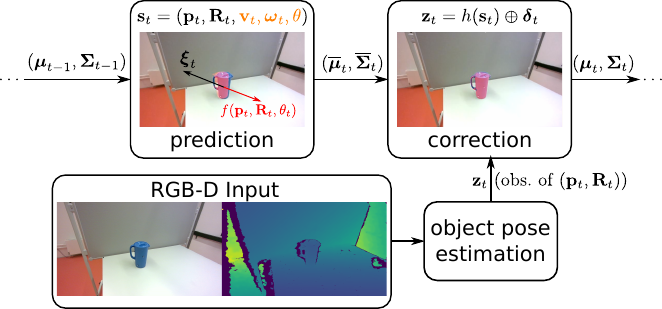} \\
    \caption{
    Overview of our method.
    Our physics-based EKF tracker can filter unobserved velocity ($\mathbf{v}_t, \boldsymbol{\omega}_t$) and friction parameters ($\theta$) (orange) as part of the state $\mathbf{s}_t$, and takes frictional forces (red arrow) into account when predicting using the current state estimate.
    Object position and orientation observations for the correction step are estimated from RGB-D images using a state-of-the-art object detector and pose estimator (CIR~\cite{lipson2022_coupled}).
    }
    \label{fig:teaser}
\end{figure}

In this paper, we propose a novel hybrid probabilistic approach for object tracking.
We use an analytical physics-based dynamics model that includes collision and friction constraints as state-transition model in an Extended Kalman Filter (EKF) framework.
The object pose is measured using a state-of-the-art approach for deep learning based object detection and pose estimation from images (e.g., Coupled Iterative Refinement, CIR~\cite{lipson2022_coupled}). 
Our approach explicitly models uncertainty in the dynamics prediction and observations to filter the object state and, importantly, the \emph{physical parameters} (\ie friction coefficient) with uncertainty estimates.
For the dynamics model we use a  custom 3D rigid body physics engine which extends~\cite{lcp_physics_2d} to 3D object dynamics with arbitrary mesh-based shapes and is end-to-end differentiable. 

We evaluate our approach using synthetic and real-world datasets with ground-truth information and analyse its accuracy in trajectory and parameter estimation.
Due to the lack of suitable datasets available for our novel problem setting, we created custom synthetic data and recorded real-world data in our lab.
We make our synthetic and real datasets publicly available\footnote{\href{https://keeper.mpdl.mpg.de/d/5ec213b655b44e40a382/}{https://keeper.mpdl.mpg.de/d/5ec213b655b44e40a382/}} to foster reproducibility and future work in this novel research direction of physics-based scene reconstruction. 

In summary, we contribute the following:
\begin{itemize}\setlength\itemsep{0em}
    \item We propose a novel approach for object-level scene understanding which not only tracks the 3D object position and orientation, but also recovers physical object properties such as friction coefficients.
    \item Our approach is based on a new way to combine Extended Kalman filtering with differentiable simulation based on a LCP-based formulation of rigid-body dynamics.
    \item We analyse our approach on new custom synthetic and real-world datasets for our novel problem setting. We make the datasets publicly available to foster future research and comparison.
\end{itemize}
In this paper, we explicitly focus on scenarios with rigid objects and leave state and property estimation for articulated and non-rigid objects for future work.

\section{Related work}
\label{sec:related_work}

\paragraph{Object pose tracking.}
In order to leverage temporal information for object tracking, an object motion model is needed.
The simplest models assume the motion between frames to be small enough such that objects can be assumed at the previous position.
They track objects based on image edges or features~\cite{kehl2017_real,vacchetti2004_combining}, or using RGB-D data aligning the last estimated models to the current depth frame in joint reconstruction and tracking of dynamic objects \cite{ruenz2017_co,ruenz2018_maskfusion,strecke2019_em,xu2019_mid}.
However, this approach tends to fail at low frame rates or with fast moving objects.
Thus, early works \cite{harris1990_rapid,evans1990_kalman} proposed to use Kalman filters to improve tracking accuracy for these cases by assuming constant velocity motion between frames.
Many works have subsequently applied Kalman or particle filters for object tracking \cite{choi2010_real,choi2012_3d} or simultaneous camera and object tracking from visual and inertial data \cite{wang2007_simultaneous,eckenhoff2019_tightly,qiu2019_tracking,eckenhoff2020_schmidt}.
A different line of research \cite{garon2017_deep,wen2020_se3,manhardt2018_deep} employs learned heuristics to refine pose estimates from previous frames.
Due to the data-driven nature of this approach, the resulting motion models are usually difficult to interpret.
Deng et al.~\cite{deng2019_poserbpf} combines CNN-based pose estimation with a Rao-Blackwellized particle filter for pose tracking.
We propose to model motion using a differentiable physics engine \cite{lcp_physics_2d}, which lets us model more complex dynamics like collisions and friction while maintaining an interpretable motion model and state representation, and also enabling estimating the physical properties such as the friction coefficient.

More closely related to our approach are the works in~\cite{jongeneel2022_model,xu2022_physicsbasedtracking}.
Jongeneel~\etal~\cite{jongeneel2022_model} model collisions and friction in an unscented particle filter framework.
In~\cite{xu2022_physicsbasedtracking} uncertainty in physical parameters is sampled in a physics-based motion model during particle filtering for 6-DoF object pose tracking.
In contrast to our work, they do not use a differentiable physics model, do not estimate physical parameters, and the works are demonstrated for simple box or cylinder shapes while we extend~\cite{lcp_physics_2d} to handle meshes of known objects.

Another work incorporating frictional contact information in a Kalman Filter setting is~\cite{varin2020_constrained}.
The authors propose to directly include contact constraints in the Kalman Filter equations, resulting in a contact constrained Kalman Filter (CCKF) formulation.
They present superior performance for tracking simulated and real objects compared to an Unscented Kalman Filter formulation.
However, they also do not demonstrate estimating object properties such as friction coefficients like in our approach.

\paragraph{Physics-aware scene reconstruction.}
Recently, several methods have been proposed that utilize prior knowledge about the physical world in scene reconstruction.
Wada~\etal~\cite{wada2020_morefusion} combine predictions for volumetric shape completion with collision-based pose refinement for physically plausible shape estimates.
The approach in~\cite{strecke2020_where} performs coarse shape completion in dynamic scenes based on physical plausibility constraints.
Some recent works utilize differentiable physics simulations to estimate physical parameters~\cite{kandukuri2021_physical,krishnamurthy2021_gradsim,cleach2022_differentiable} or shape parameters~\cite{strecke2021_diffsdfsim}, but do not filter object state and properties frame by frame like our method.
A related line of research which uses physics-based models are methods that recover and track articulated object models, for instance, for human hands or bodies~\cite{melax2013_dynamicsbasedskeletal,schmidt2015_dart,heiden2022_artrigboddyn}.

\section{Background}
\label{sec:background}

\subsection{Differentiable Physics Simulation}
\label{subsec: differentiable physics}
The differentiable physics engine in our work is a 3D extension of the work by de Avila Belute-Peres~\etal~\cite{lcp_physics_2d}.
We now briefly describe the underlying model.
The relation between time-dependent wrenches $\mathbf{f} : [0,\infty) \to \mathbb{R}^{6N}$ (\ie forces and torques) acting on $N$ objects and their motion is given by the Newton-Euler equations
    $\mathbf{f}_{ext} = \mathbf{M}\ddot{\mathbf{x}}$.
The block-diagonal matrix $\mathbf{M} \in \mathbb{R}^{6N\times 6N}$ denotes the collection of the object's mass-inertia tensor and $\mathbf{x}_t\in SE(3)^N$ their poses at time step $t$ with position $\mathbf{p}_{t,i} \in \mathbb{R}^3$ and rotation $\mathbf{R}_{t,i} \in SO(3)$. 
The twists are denoted by $\boldsymbol{\xi}$ and wrenches by $\mathbf{f}$. 
The Newton-Euler equations represent the force-acceleration formulation which is unstable under multiple contact and friction constraints~\cite{Cline_2002}. 
Therefore a velocity-impulse based formulation with contact and friction constraints is derived by time discretizing acceleration~\cite{Cline_2002},
\begin{equation}
\begin{split}
    \mathbf{M}\boldsymbol{\xi}_{t+\Delta t} - \mathbf{J}^T_c\lambda_c - \mathbf{J}^T_f\lambda_f - \mathbf{J}^T_e\lambda_e &= \mathbf{M}\boldsymbol{\xi}_{t} + \Delta t\mathbf{f}_{ext}\\
    \mathbf{J}_c\boldsymbol{\xi}_{t+\Delta t} &\geq -c\mathbf{J}_c\boldsymbol{\xi}_{t}\\
    \mathbf{J}_f\boldsymbol{\xi}_{t+\Delta t} + \mathbf{E}\lambda_f &\geq 0\\
    \mu\lambda_c - \mathbf{E}\lambda_f &\geq 0\\
    \mathbf{J}_e\boldsymbol{\xi}_{t+\Delta t} &= 0\\
\end{split}
    \label{eq: rigid body equations}
\end{equation}
\noindent
where $\mathbf{J}_c$, $\mathbf{J}_f$ and $\mathbf{J}_e$ represent contact, frictional and equality Jacobians, respectively. 
$\lambda_c$, $\lambda_f$ and $\lambda_e$ represent contact, frictional and equality Lagrange multipliers. 
$c$ and $e$ represent coefficient of restitution and coefficient of friction, respectively, and $\Delta t$ is the time step. The terms $\mathbf{J}^T_c\lambda_c$, $\mathbf{J}^T_f\lambda_f$ and $\mathbf{J}^T_e\lambda_e$ describe contact, frictional, and equality constraint forces, respectively. 
The linear complementarity problem (LCP) formulation of rigid body dynamics with contacts and friction can be written as~\cite{Cline_2002} and~\cite{lcp_physics_2d}
\begin{equation}
	\begin{pmatrix}
		0\\
		\mathbf{s}\\
		0
	\end{pmatrix}
	+
	\begin{pmatrix}
		\mathbf{M} && \mathbf{G}^\top && \mathbf{A}^\top\\
		\mathbf{G} && \mathbf{F} && 0\\
		\mathbf{A} && 0 && 0\\
	\end{pmatrix}
	\begin{pmatrix}
		-\boldsymbol{\xi}_{t+\Delta t}\\
		\mathbf{z}\\
		\mathbf{y}\\
	\end{pmatrix}
	=
	\begin{pmatrix}
		-\mathbf{q}\\
		\mathbf{m}\\
		0\\
	\end{pmatrix}\\
\label{eq: LCP formulation}
\end{equation}
\begin{equation*}
    \mathrm{subject\;to\;} \mathbf{s} \geq 0,\;\mathbf{z} \geq 0,\; \mathbf{s}^T\mathbf{z} = 0
\end{equation*}
\begin{equation*}
    \textrm{where}\;\;\mathbf{q} = \mathbf{M}\boldsymbol{\xi}_{t} + \Delta t\mathbf{f}_{ext}
\end{equation*}
\noindent
where $\mathbf{G}$ and $\mathbf{A}$ represent Jacobians of inequality and equality constraints. 
$\mathbf{F}$ represents the matrix of frictional coefficients. 
$\mathbf{z}$ and $\mathbf{y}$ are vectors of dual variables and $s$ is the vector of slack variables.
Open Dynamics Engine's (ODE)~\cite{ode2008} collision detector is used for  contact detection and resolution. 
The LCP described in equation~\eqref{eq: LCP formulation} is solved using the primal-dual interior point method~\cite{lcp_physics_2d} for $\boldsymbol{\xi}_{t+h}$. 
The pose is obtained by Euler integration, i.e., $\mathbf{p}_{t+\Delta t} = \mathbf{p}_t + \mathbf{v}_{t+\Delta t}\Delta t$ and
$\mathbf{R}_{t+\Delta t} = \exp( \boldsymbol{\omega}_{t+\Delta t}\Delta t ) \mathbf{R}_t$.

Using this formulation, the results stored while solving the forward problem can be used to calculate the analytical gradients with respect to each matrix at the solution efficiently as described in~\cite{Amos_Kolter_2017} and~\cite{lcp_physics_2d}.

\subsection{Object Detection and 6-DoF Pose Estimation}
We use Coupled Iterative Refinement (CIR, \cite{lipson2022_coupled}) to detect known objects and estimated their six degree of freedom (6-DoF) pose (\ie position and orientation) from images.
CIR first uses Mask R-CNN \cite{he2017_mask} to detect objects.
It then estimates an initial position by matching the projected 3D bounding box of the known object model with the detected mask and runs EfficientNet \cite{tan2019_efficientnet} to estimate an initial orientation.
This initial estimate is then refined by aligning the rendered view of the known object model in the current pose with the observed detected object using RAFT \cite{teed2020_raft} features to match object points.
For more details on detection and 6-DoF pose estimation of known objects, see \cite{lipson2022_coupled}.

\section{Method}
\label{sec:method}

We propose an EKF-based tracker to filter 6D pose and velocity of the object along with the coefficient of friction between object and support surface.
The state $\mathbf{s}_t$ at time $t$ in the EKF is represented by a tuple $\mathbf{s}_t = ( \mathbf{p}_t, \mathbf{R}_t, \mathbf{v}_t, \boldsymbol{\omega}_t, \theta ) \in \mathbb{R}^3 \times SO(3) \times \mathbb{R}^3 \times \mathbb{R}^3 \times \mathbb{R}$ with position $\mathbf{p}_t$, rotation $\mathbf{R}_t$, linear velocity $\mathbf{v}_t$, angular velocity $\boldsymbol{\omega}_t$ and friction coefficient $\theta$, respectively.
Position and linear velocity are represented in world frame, while rotation and angular velocity are represented in body frame.
Note that the rotation is element of the Special Orthogonal Group $SO(3)$ and, hence, cannot be represented as a vector in Euclidean space.
Thus, we represent the mean state estimate of the EKF in tuple form, and use a local state representation $\mathbb{R}^{13}$ for the covariances by expressing perturbations of the state in the original Euclidean space for the non-rotation state components and the Euclidean vector representation $\boldsymbol{\tau} \in \mathbb{R}^3$ of the tangent space of rotations $\widehat{\boldsymbol{\tau}} \in so(3)$.
The operator $\widehat{\boldsymbol{\tau}} = \left( \begin{array}{ccc} 0 & -\boldsymbol{\tau}_2 & \boldsymbol{\tau}_1\\ \boldsymbol{\tau}_2 & 0 & -\boldsymbol{\tau}_0\\ -\boldsymbol{\tau}_1 & \boldsymbol{\tau}_0 & 0  \end{array} \right)$ maps the tangent vector $\boldsymbol{\tau}$ to its corresponding Lie algebra element $\widehat{\boldsymbol{\tau}} \in so(3)$.
The tangent vector in the local state representation expresses left-multiplied perturbations to the current rotation, i.e. the perturbed rotation is $\mathbf{R}_t^{\boldsymbol{\tau}} = \exp\left( \widehat{\boldsymbol{\tau}}_t \right) \mathbf{R}_t$. 

\subsection{Motion Model}

The probabilistic motion model in our EKF
\begin{equation}
    \mathbf{s}_{t+1} = g( \mathbf{s}_{t} ) \boxplus \boldsymbol{\epsilon}_t 
\end{equation}
is the 3D differentiable physics model as described in Sec.~\ref{subsec: differentiable physics} augmented with Gaussian noise $\boldsymbol{\epsilon}_t \sim \mathcal{N}( 0, \mathbf{S} ), \mathbf{S} \in \mathbb{R}^{13 \times 13}$ to accomodate unmodeled effects.
The Gaussian noise is sampled in the local state representation and added to the predicted state using the operator
\begin{multline}
    \mathbf{s} \boxplus \boldsymbol{\epsilon} := \left( \mathbf{p}_t + \boldsymbol{\epsilon}_{1:3},  \exp( \widehat{\boldsymbol{\epsilon}}_{4:6} ) \mathbf{R}_t, \right. \\
    \left.\mathbf{v}_t + \boldsymbol{\epsilon}_{7:9}, \boldsymbol{\omega}_t + \boldsymbol{\epsilon}_{10:12}, \theta + \boldsymbol{\epsilon}_{13} \right).
\end{multline}
Function 
\begin{multline}
    g( \mathbf{s}_{t} ) := ( \mathbf{p}_t + \Delta t \mathbf{v}^*_{t+\Delta t},  \exp( \Delta t \boldsymbol{w}^*_{t+\Delta t} ) \mathbf{R}_t,\\ \mathbf{v}^*_{t+\Delta t}, \boldsymbol{w}^*_{t+\Delta t}, \theta_t  )
\end{multline}    
propagates the state according to the physics-based dynamics with the LCP solution $\boldsymbol{\xi}^*_{t+\Delta t} = \left({\boldsymbol{w}^*_{t+\Delta t}}^\top, {\mathbf{v}^*_{t+\Delta t}}^\top\right)^\top$.
For the friction coefficient estimate $\theta_t$, the motion model assumes it to be constant with Gaussian noise.
The estimated friction coefficient is used in the physics-based model to parameterize the combined friction coefficient $\mu = \theta_t^2$ between object and support surface, assuming that both object and support surface have the same friction coefficient $\theta_t$, i.e. $\boldsymbol{\xi}^*_{t+\Delta t}(\theta_t)$ is a function of $\theta_t$.
By this, we can estimate the friction coefficient alongside with the dynamic state of the object. 

\subsection{Observation Model}

Object pose observations $\mathbf{z}_t = ( \mathbf{z}_{\mathbf{p},t}, \mathbf{z}_{\mathbf{R},t} ) \in \mathbb{R}^3 \times SO(3)$ at time $t$ are determined by CIR~\cite{lipson2022_coupled} and are modeled by
\begin{equation}
    \mathbf{z}_t = h( \mathbf{s}_t ) \oplus \boldsymbol{\delta}_t = ( \mathbf{p}_t, \mathbf{R}_t ) \oplus \boldsymbol{\delta}_t,
\end{equation}
where $\boldsymbol{\delta}_t \sim \mathcal{N}( 0, \mathbf{Q} ), \mathbf{Q} \in \mathbb{R}^{6 \times 6}$.
Again, the Gaussian noise is determined in a local parameterization of the position and rotation, i.e., 
\begin{equation}
    \mathbf{z} \oplus \boldsymbol{\delta} := ( \mathbf{p}_t + \boldsymbol{\delta}_{1:3}, \exp(\widehat{\boldsymbol{\delta}}_{4:6}) \mathbf{R}_t )
\end{equation}

\subsection{EKF Prediction}

The EKF prediction follows the standard update equations
\begin{equation}
    \begin{aligned}
        \overline{\boldsymbol{\mu}}_t &= g(\boldsymbol{\mu}_{t-1}),~~~
        \overline{\mathbf{\Sigma}}_t = \mathbf{G}_t \mathbf{\Sigma}_{t-1} \mathbf{G}^\top_t + \mathbf{S}_t, \text{where}\\
    \mathbf{G}_t &:=  \left. \frac{d\left( g(\boldsymbol{\mu}_t \boxplus \boldsymbol{\tau}) \boxminus g(\boldsymbol{\mu}_t) \right)}{d\boldsymbol{\tau}} \right|_{\boldsymbol{\tau} = 0}
    \end{aligned}
\end{equation}
is the derivative of the motion model in the local state representation. 
The $\boxminus$ operator is defined as
\begin{equation}
    \mathbf{x}' \boxminus \mathbf{x} := \left( \mathbf{p}' - \mathbf{p}, \log( \mathbf{R}' \mathbf{R}^\top ), \mathbf{v}' - \mathbf{v}, \boldsymbol{\omega}' - \boldsymbol{\omega}, \theta' - \theta \right).
\end{equation}
We efficiently calculate the Jacobian using PyTorch's autograd module.
Please refer to the supplemental material for further details.

\subsection{EKF Correction}

The EKF correction step updates the predicted state with the observation,
\begin{equation}
    \begin{split}
        \mathbf{K}_t &= \mathbf{\Sigma}_t \mathbf{H}^\top_t \left(\mathbf{H}_t \overline{\mathbf{\Sigma}}_t \mathbf{H}^\top_t + \mathbf{Q}_t\right)^{-1},\\
        \boldsymbol{\mu}_t &= \overline{\boldsymbol{\mu}}_t \boxplus \mathbf{K}_t ( \mathbf{z}_t \ominus h( \overline{\boldsymbol{\mu}}_t ) ),\\
        \mathbf{\Sigma}_t &= \left(\mathbf{I} - \mathbf{K}_t \mathbf{H}_ t\right) \overline{\mathbf{\Sigma}}_t,
    \end{split}
\end{equation}
where the operator $\ominus$ is defined as
\begin{equation}
    \mathbf{z}' \ominus \mathbf{z} := \left( \mathbf{p}' - \mathbf{p}, \log( \mathbf{R}' \mathbf{R}^\top ) \right)
\end{equation}
and
\begin{equation}
    \mathbf{H}_t := \left. \frac{d\left( h(\overline{\boldsymbol{\mu}}_t \boxplus \boldsymbol{\tau}) \boxminus h(\overline{\boldsymbol{\mu}}_t) \right)}{d\boldsymbol{\tau}} \right|_{\boldsymbol{\tau} = 0}.
\end{equation}

To exclude outliers in the pose measurement which violate the Gaussian noise assumption, we apply gating by rejecting correction steps if the following condition holds
\begin{equation}
    ( \mathbf{z}_t \ominus h( \overline{\boldsymbol{\mu}}_t ) )^\top \left(\mathbf{H}_t \overline{\mathbf{\Sigma}}_t \mathbf{H}^\top_t + \mathbf{Q}_t\right)^{-1} ( \mathbf{z}_t \ominus h( \overline{\boldsymbol{\mu}}_t ) ) > \zeta
\end{equation}
where $\zeta$ is a gating threshold.

\subsection{Recursive State Estimation}

The EKF proceeds by sequentially predicting the current state estimate forward in time until the next image frame.
Note that for accurate prediction with the physics-based dynamics model, typically small time increments are needed. 
Therefore, multiple EKF prediction steps can be applied (2 in our experiments at 60\,Hz) until the next frame (30\,Hz image frame rate in our experiments), before a measurement is integrated for an image frame by the EKF correction step.

\begin{figure}
    \centering
    \includegraphics[width=\linewidth]{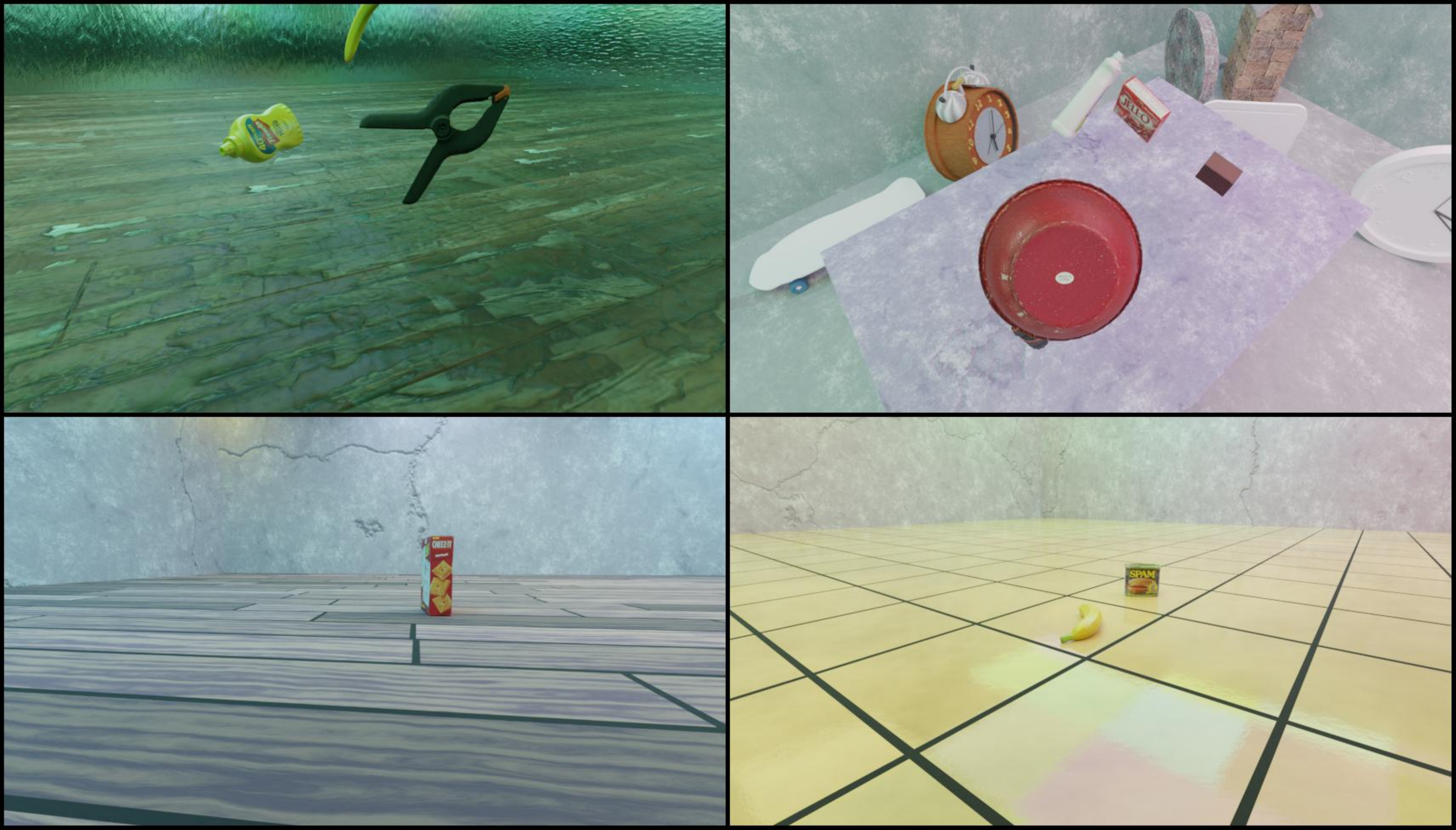}
    \caption{Top: Example frames from two sequences used for detector training (left: detector for synthetic data, right: detector for real-world data). Bottom: Example frames from two sequences with sliding objects (left: single object, right: multiple objects).}
    \label{fig:data_detector_training_sliding}
\end{figure}

\section{Experiments}
\label{sec:experiments}

\subsection{Experiment Setup}
\label{sec:data_collection}

We evaluate our approach with synthetic data and real data captured in table-top scenes.
Both kinds of data consist of RGB-D images of YCB objects~\cite{calli2015_benchmarking,calli2015_ycb} sliding on a plane
with varying initial velocity.
The RGB-D images come at a frame rate of~30\,fps with a resolution of~$848 \times 480$ pixels.

\paragraph{Synthetic Dataset}
As no suitable benchmark datasets are publicly available for our problem setting of 3D physical scene reconstruction, we render a novel synthetic dataset using the physics-based rendering (PBR) tool Blender~\cite{blender} and the PyBullet simulation engine~\cite{pybullet}.
We provide further details on our dataset in the supplementary material. 
We render two datasets with the same intrinsics as the Intel RealSense D455 RGB-D camera.
The first is solely used for fine-tuning the object detector and pose estimator~\cite{lipson2022_coupled}, where objects are placed in random poses above or on the ground and the camera moves in a circle around the static object arrangement (see example frames in Fig.~\ref{fig:data_detector_training_sliding}, top).
We provide details on the split sizes for this dataset in the supplementary material.
The second dataset contains a single object sliding on the floor in each sequence.
As mentioned earlier, the initial pose and velocity vectors of the object are sampled randomly, generating a random trajectory of 90 frames at 30 fps by downsampling to the output frame rate from 240 steps per second simulated in PyBullet.
This dataset contains 840 training, 105 validation (which we use to fit the EKF parameters), and 105 testing sequences (40 training, 5 validation and 5 test per object).
Since we do not train the detector on this dataset, we do not need as much background variety and use procedural textures that appear like concrete, wood, tiles, or wooden planks for the background.
Example frames from this dataset can be seen in Fig.~\ref{fig:data_detector_training_sliding}, bottom.
Additionally, this dataset contains 200 training, 25 validation, and 25 test sequences for two-object scenes in which one object slides and (likely) collides with another initially static object.
Statistics of the velocities in the sequences can be found in the supp. material.

\begin{figure}
    \centering
    \includegraphics[width=.99\linewidth]{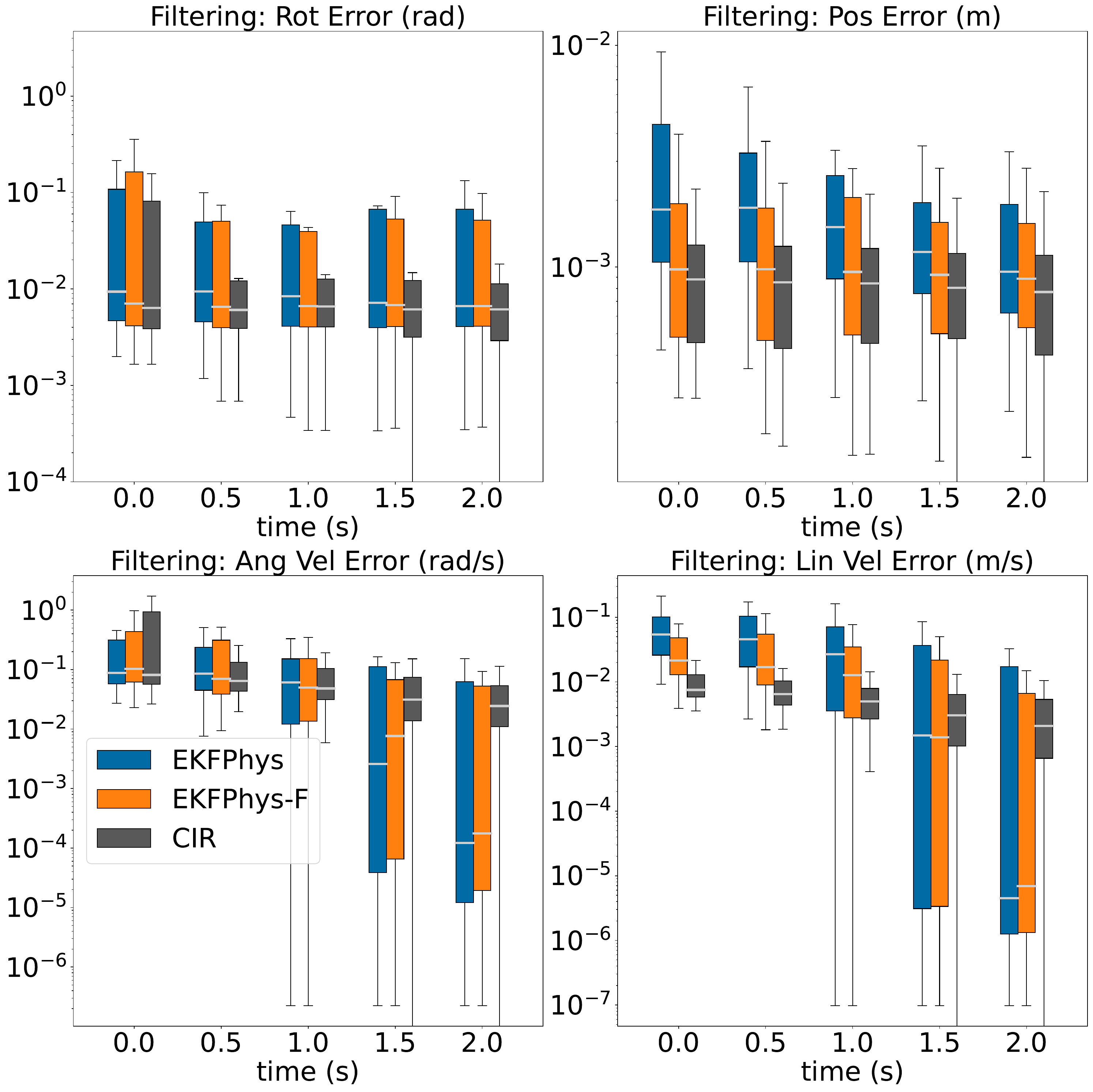}
    \caption{
    Filtering accuracy on single object sliding in synthetic sequences for non-symmetric objects.
    The box plots show the median and quartiles of average errors in rotation (top-left), position (top-right), angular (bottom-left) and linear (bottom-right) velocity over all the scenes in the dataset.
    The x-axis denotes the time step from which we evaluate the average error over all sequences.
    }
    \label{fig:trajectory_error_filtering_single_obj_nonsymm_synthetic}
\end{figure}
\begin{figure}
    \centering
    \includegraphics[width=.99\linewidth]{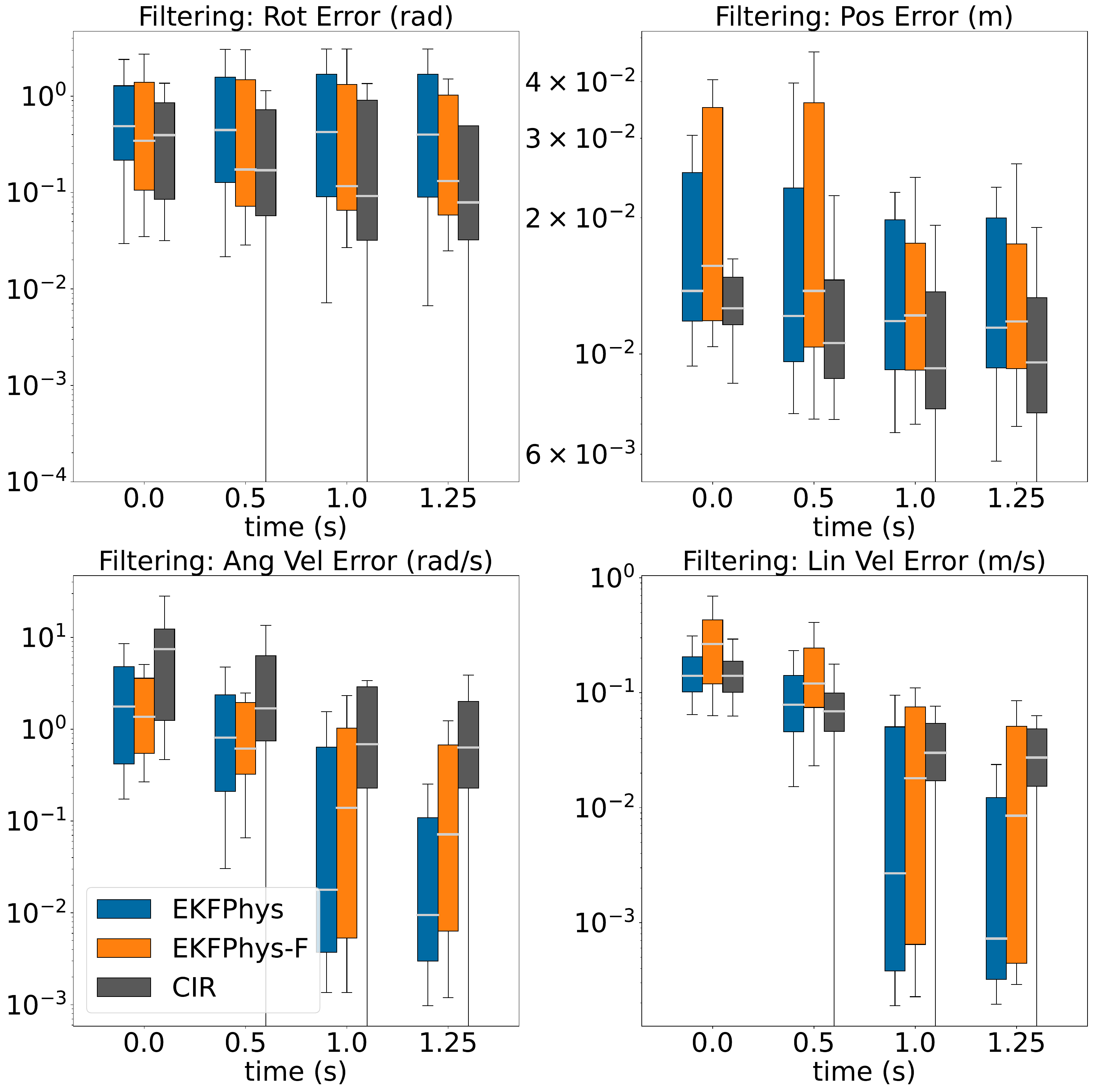}
    \caption{
    Filtering accuracy on single object sliding in real sequences for non-symmetric objects. 
    The labeling convention corresponds to Fig.~\ref{fig:trajectory_error_filtering_single_obj_nonsymm_synthetic}.
    }
    \label{fig:trajectory_error_filtering_single_obj_nonsymm_real}
\end{figure}

\paragraph{Real-World Dataset}

For our real-world experiments, we push 4 different objects from the YCB Objects~\cite{calli2015_benchmarking,calli2015_ycb} namely pitcher, bleach cleanser, mustard bottle, and mug on a table and record 5 RGB-D sequences per object with an Intel RealSense D455 camera at 60\,Hz as well as ground-truth poses for the objects at 240\,Hz using a motion capture system.
For capturing ground truth we attach MoCap markers to the object and align its CAD model with the object shape in the MoCap world frame using alignment tools provided by the MoCap system.
We also attach markers to the corners of the table to obtain an accurate estimate of the table plane.
CIR estimates the 6D object pose in the camera frame.
To get the estimated pose in the MoCap world frame, we need a transformation from the camera frame to the motion capture (MoCap) system's world frame. 
For that we attach a mount with MoCap markers to the D455 and perform a hand-eye calibration to get the transformation from the MoCap frame for the mount's markers to the optical frame of the camera.
The MoCap system measures the transformation from the mount's marker to its world frame.
The MoCap pose measurements are sent to a second PC which captures the RGB-D images. 
The average time-shift between RGB-D images and MoCap data is determined by the hand-eye calibration tool.
The pose data from MoCap system is recorded at 240\,Hz and the RGB-D image data from D455 camera is recorded at 60\,Hz using ROS. 
The dataset is generated by grouping the pose data and image data with similar timestamps and downsampling it to 30\,Hz to match our synthetic dataset.
We refer to the supplementary material for statistics of the linear and angular velocities in the sequences.

\paragraph{Evaluation Measures and Baselines}

We evaluate filtering and prediction accuracy of our physics-based EKF (EKFPhys) and compare it on synthetic scenes with a variant of our physics-based EKF for which the friction coefficient is assumed known (set to ground truth) and not estimated (EKFPhys-F).
We compare these methods against CIR as a baseline.
Since CIR only gives pose estimates, we use finite differences to estimate velocities. 
Due to the novelty of the problem setting of estimating trajectory and physical parameters by recursive filtering, no other relevant baseline methods are available (see Sec.~\ref{sec:related_work}).
We measure the accuracy in terms of the average error in position, linear, and angular velocities (average norm of the difference between estimate and ground truth), and the average rotation error in terms of the angle of the axis-angle representation of the relative rotation between estimate and ground truth.
As symmetric objects can lead to inconsistent pose estimations for consecutive frames by the pose estimator (it will just give any orientation fitting under symmetry), we separate the evaluation in symmetric and non-symmetric objects by the specifications in the BOP benchmark \cite{hodan2018_bop,hodan2020_bop} for the YCB objects.
In the following, we will only analyze scenes with non-symmetric objects and we provide additional analysis for scenes with symmetric objects in the supp. material.

\paragraph{Tracking Pipeline Details}

Our EKF filter is initialized from the second frame of the sequence, where initial poses are given by the CIR pose estimator.
The velocity is initialized using the finite difference of poses in the first two frames multiplied by the frame rate of CIR (30 Hz).
The mass is known from the dataset.
If friction coefficients are filtered, too, they are initialized at zero.
The EKF prediction step is run at 60\,Hz, whereas the correction step runs after every second prediction step at 30\,Hz. 
We train the Mask R-CNN \cite{he2017_mask} detector for CIR \cite{lipson2022_coupled} by running the code from the CosyPose \cite{labbe2020_cosypose} repository on our detector training dataset (see section \ref{sec:data_collection}), starting from the pretrained checkpoint trained on YCB Video \cite{xiang2018_posecnn} as provided by the authors of \cite{lipson2022_coupled}.
Afterwards, the RaftSE3 module from CIR \cite{lipson2022_coupled} is trained from scratch on our dataset for the final pose alignment.
We found the pretrained EfficientNet \cite{tan2019_efficientnet} model as provided by the authors of CIR \cite{lipson2022_coupled} to be sufficiently accurate for our experiments, as both EfficientNet and RaftSE3 only act on image crops generated by the initial detections.
We train the Mask R-CNN detector on different splits for the synthetic and real-world experiments (without and with the table, respectively, see supp.~material for details), but use the same EfficientNet and RaftSE3 models in both synthetic and real-world experiments.  We provide the iteration configurations of CIR we used for each dataset in the supplementary material.
We tune the hyperparameters (parameters of the covariance matrices, $\mathbf{\Sigma_0^{p}}, \mathbf{\Sigma_0^{R}}, \mathbf{\Sigma_0^{v}}, \mathbf{\Sigma_0^{\omega}}, \mathbf{\Sigma_0^{\theta}}, \mathbf{S}^{\{\mathbf{p}, \mathbf{R}, \mathbf{v}, \mathbf{\omega}\}}, \mathbf{S}^{\theta}, \mathbf{Q}^{p}, \mathbf{Q}^{R}$ and gating threshold, $\zeta$) for EKFPhys and EKFPhys-F individually using Optuna~\cite{optuna_2019} by multi-objective optimization (rotation, position, friction error). We provide the parameters in the supplementary material.

\subsection{Results}
\label{sec:results}
\begin{figure}
    \centering
    \includegraphics[width=.49\linewidth]{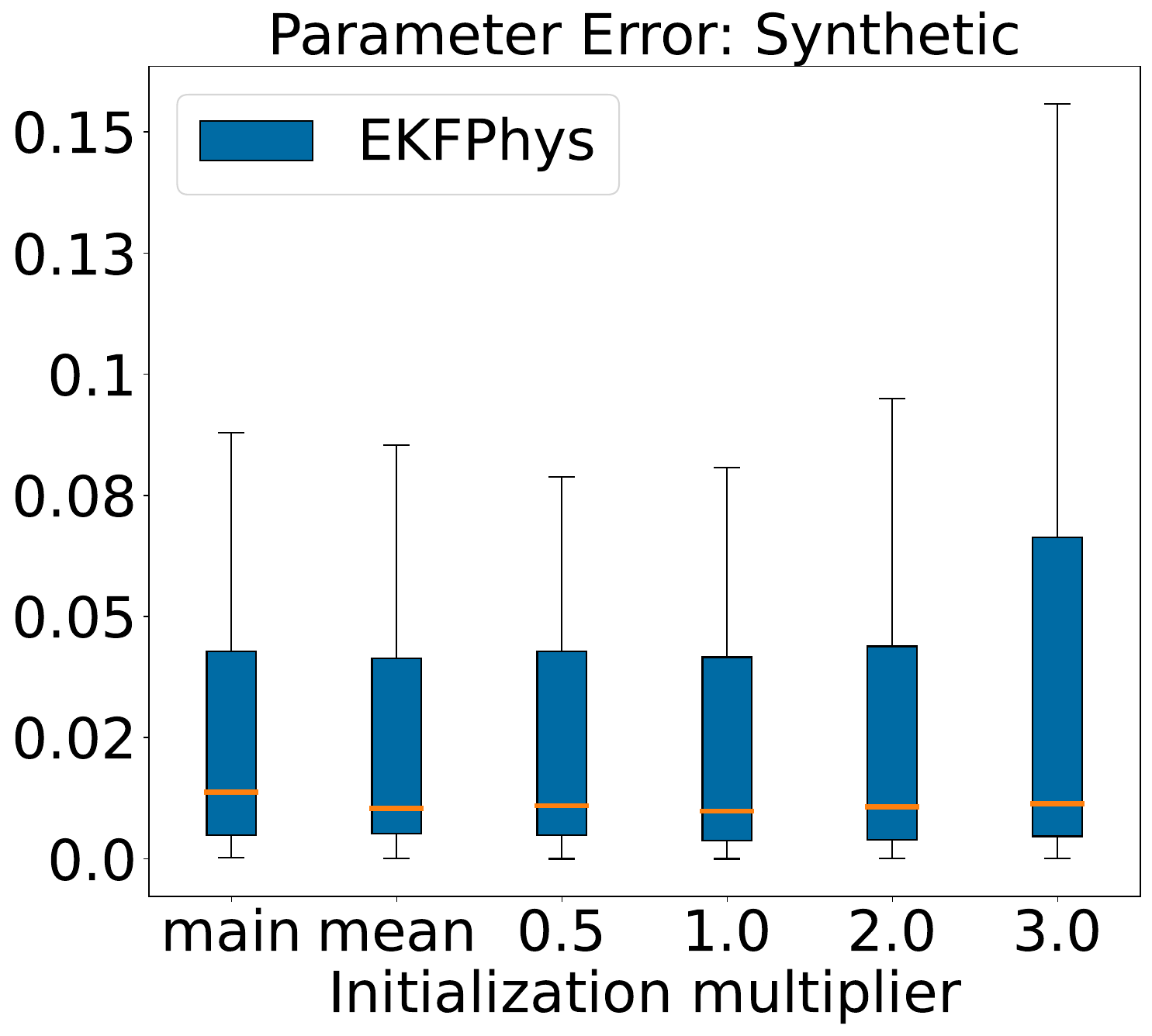}
    \includegraphics[width=.49\linewidth]{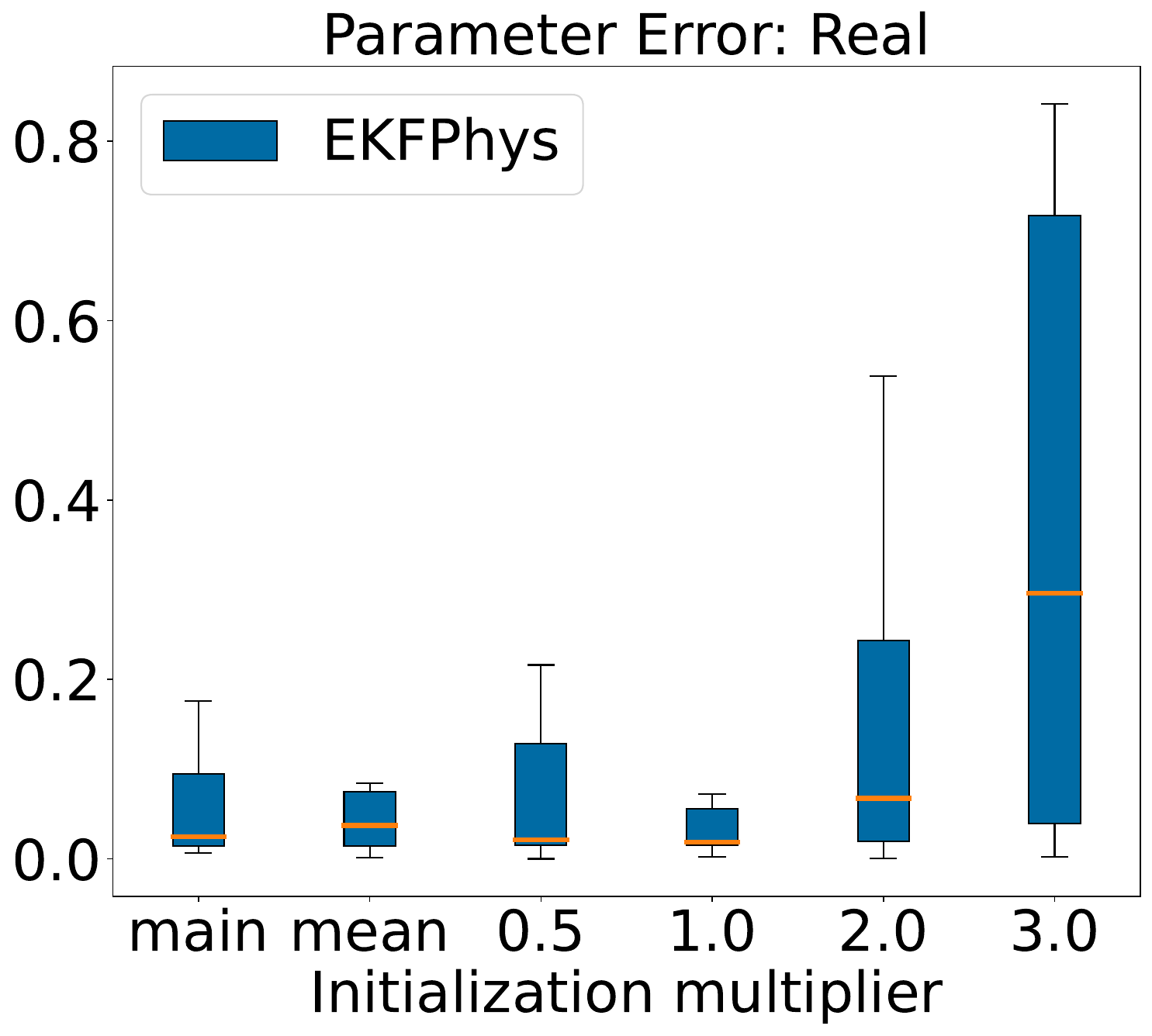} \\
    \caption{
    Friction coefficient error on single object sliding in synthetic (left) and real (right) sequences for non-symmetric objects. 
    The friction value is initialized with various multiples of gt-friction (initialization multiplier). For ``mean'' we initialize with the average value of all the friction values in the dataset (0.062 for synthetic and 0.164 for real).
    The entry ``main'' indicates the value 0.0, at which our model EKFPhys is initialized for filtering and prediction experiments.
    Our approach recovers the friction coefficients with a good accuracy in the median.
    }
    \label{fig:friction_estimation_error_single_obj_nonsymm_synphys_non_damping}
\end{figure}
\begin{table}
\centering
\footnotesize
\begin{tabular}{ccccc}
    \toprule
    & \multicolumn{2}{c}{synthetic} & \multicolumn{2}{c}{real} \\
    \cmidrule(lr){2-3} \cmidrule(lr){4-5}
    & mean & median & mean & median \\
    \midrule
    EKFPhys - main & 0.114          & \textbf{0.0137} & \textbf{0.082} & \textbf{0.0246} \\
    Zero baseline        & \textbf{0.062} & 0.0192          & 0.164          & 0.03\\
    \bottomrule
\end{tabular}
\caption{Friction error for EKFPhys - main (initialized at zero and estimates the friction through filtering) vs the baseline (constantly zero)}
\label{tab:fric_analysis}
\end{table}
\begin{figure}
    \centering
    \includegraphics[width=.99\linewidth]{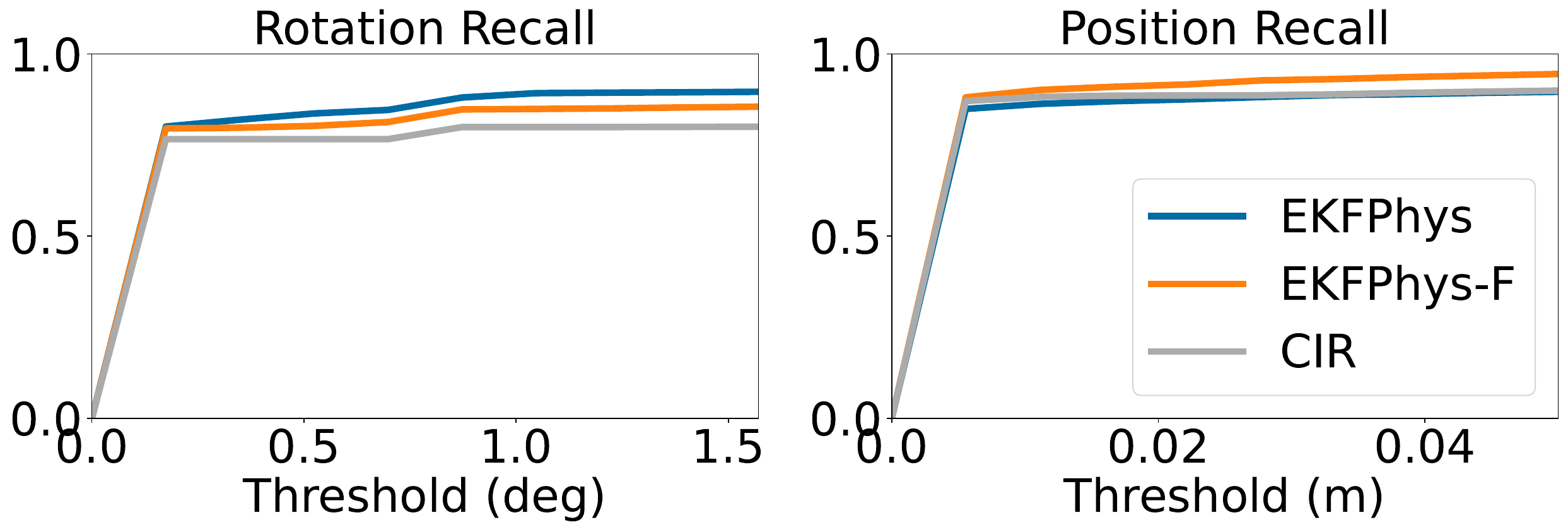} 
    \includegraphics[width=.99\linewidth]{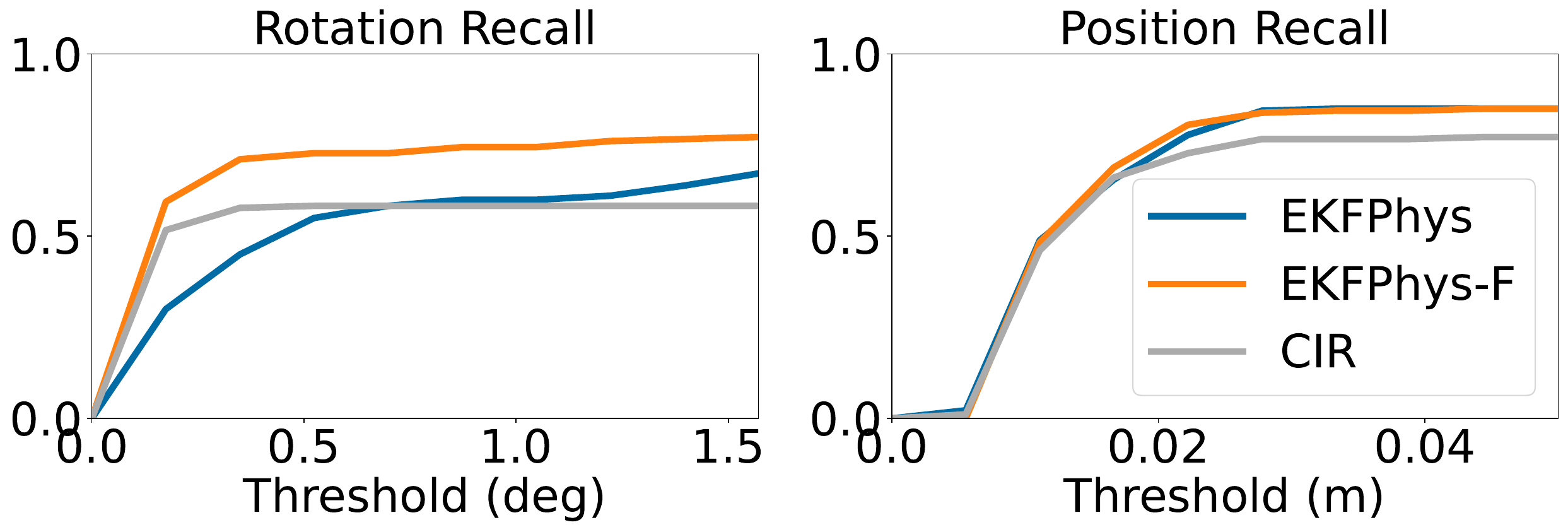} 
    \caption{
    Recall of rotation and position for single object sliding in synthetic (top) and real (bottom) sequences for non-symmetric objects.
    The x-axis denotes the threshold (in degrees for rotation and m for position) under which the detection/estimation is considered accurate.
    We achieve higher recall rates than CIR by filtering poses for frames without detections.
    }
    \label{fig:recall_filtering_single_obj_nonsymm_non_damping}
\end{figure}
\begin{figure}
    \centering
    \includegraphics[width=.99\linewidth]{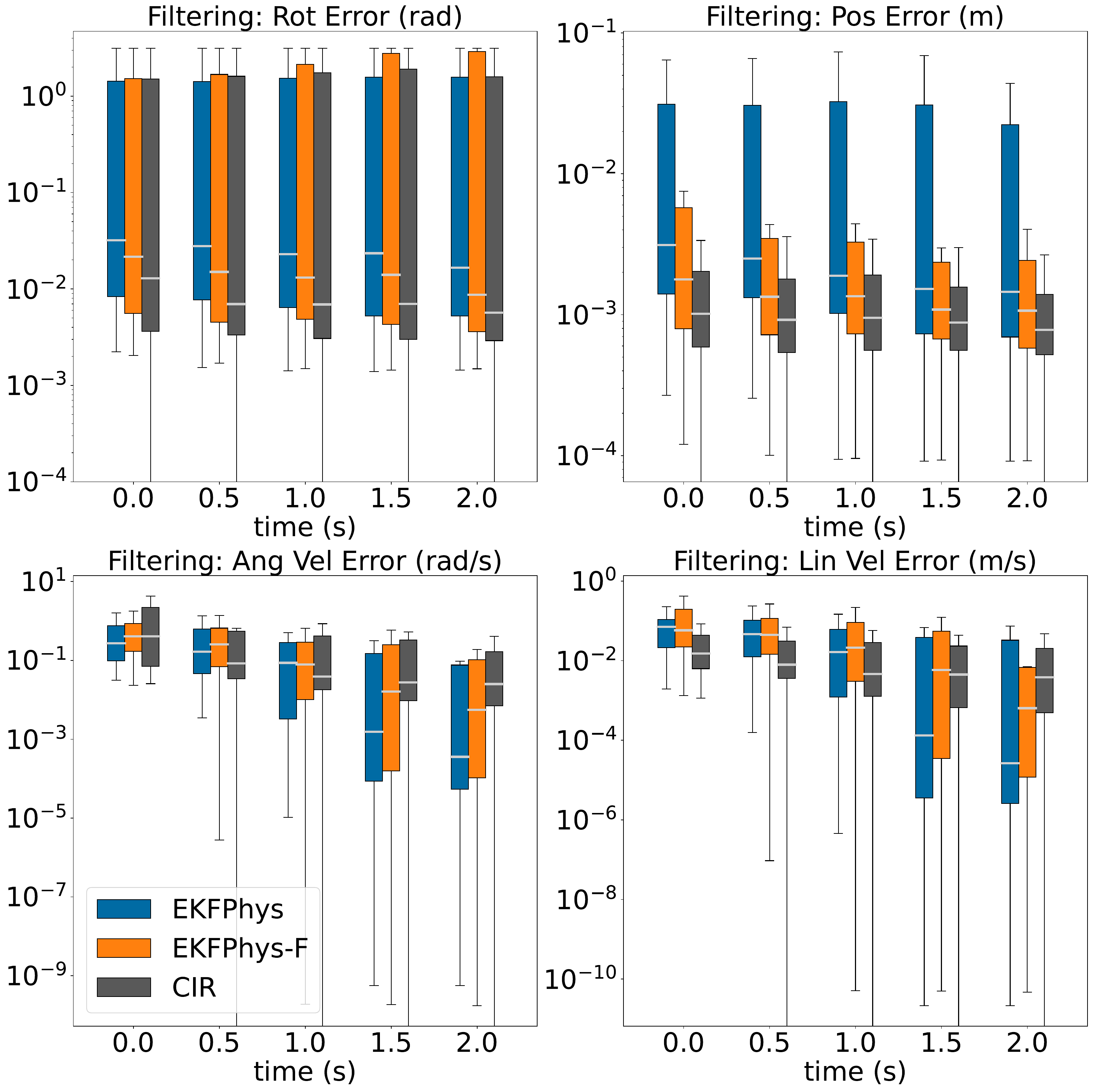}
    \caption{
    Filtering accuracy on two objects sliding and colliding in synthetic sequences for non-symmetric objects. The labeling convention corresponds to Fig.~\ref{fig:trajectory_error_filtering_single_obj_nonsymm_synthetic}.
    }
    \label{fig:trajectory_error_filtering_two_objs_nonsymm_synthetic}
\end{figure}
\begin{figure}
    \centering
    \includegraphics[width=.99\linewidth]{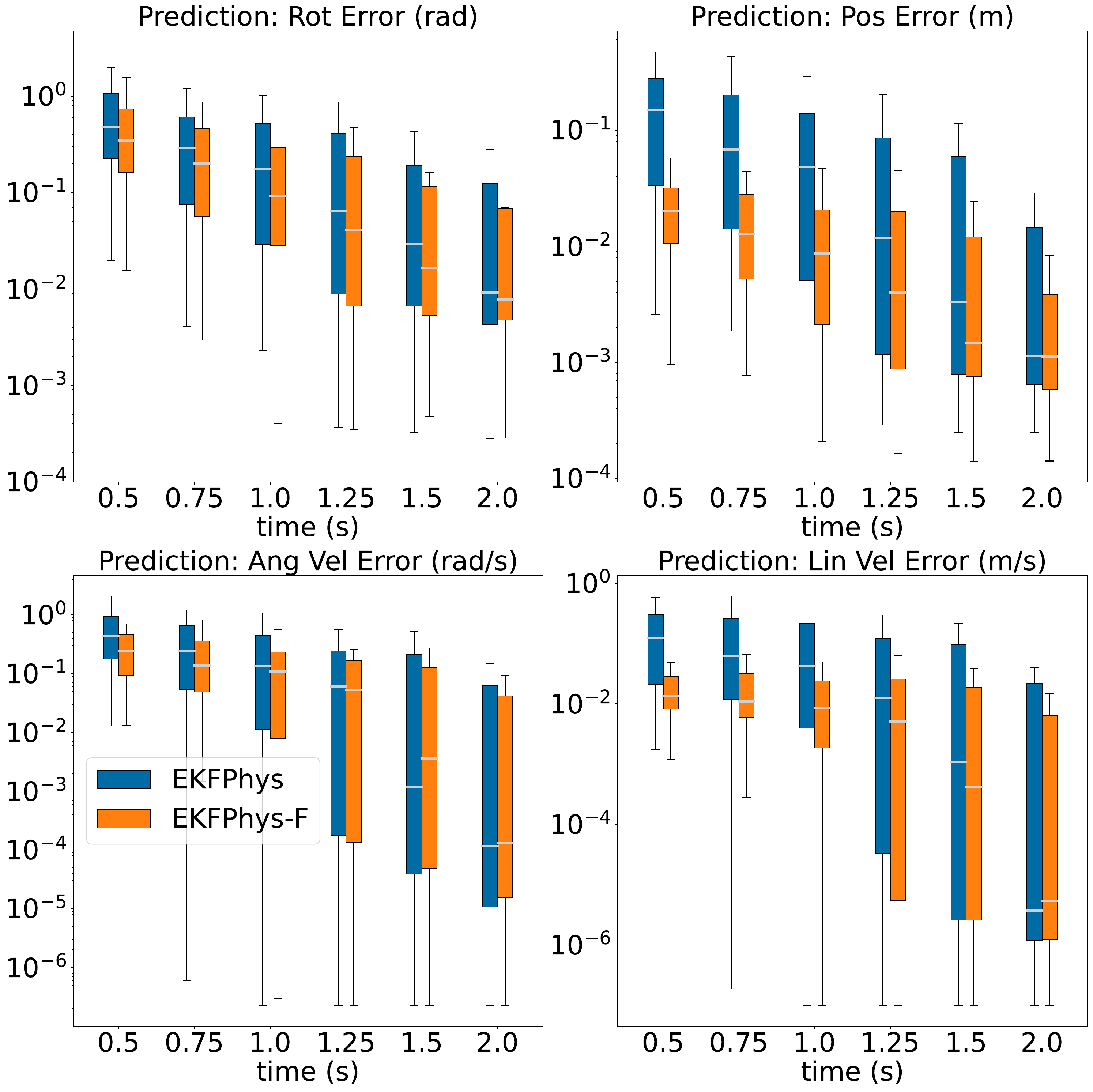}
    \caption{
    Prediction accuracy on single object sliding in synthetic sequences for non-symmetric objects.
    The box plots are arranged as in Fig.~\ref{fig:trajectory_error_filtering_single_obj_nonsymm_synthetic}.
    The x-axis denotes the time step from which we cut off observations from the detector and evaluate the average error from this point over all sequences.
    }
    \label{fig:trajectory_error_prediction_single_obj_nonsymm_synthetic}
\end{figure}
\begin{figure}
    \centering
    \includegraphics[width=.99\linewidth]{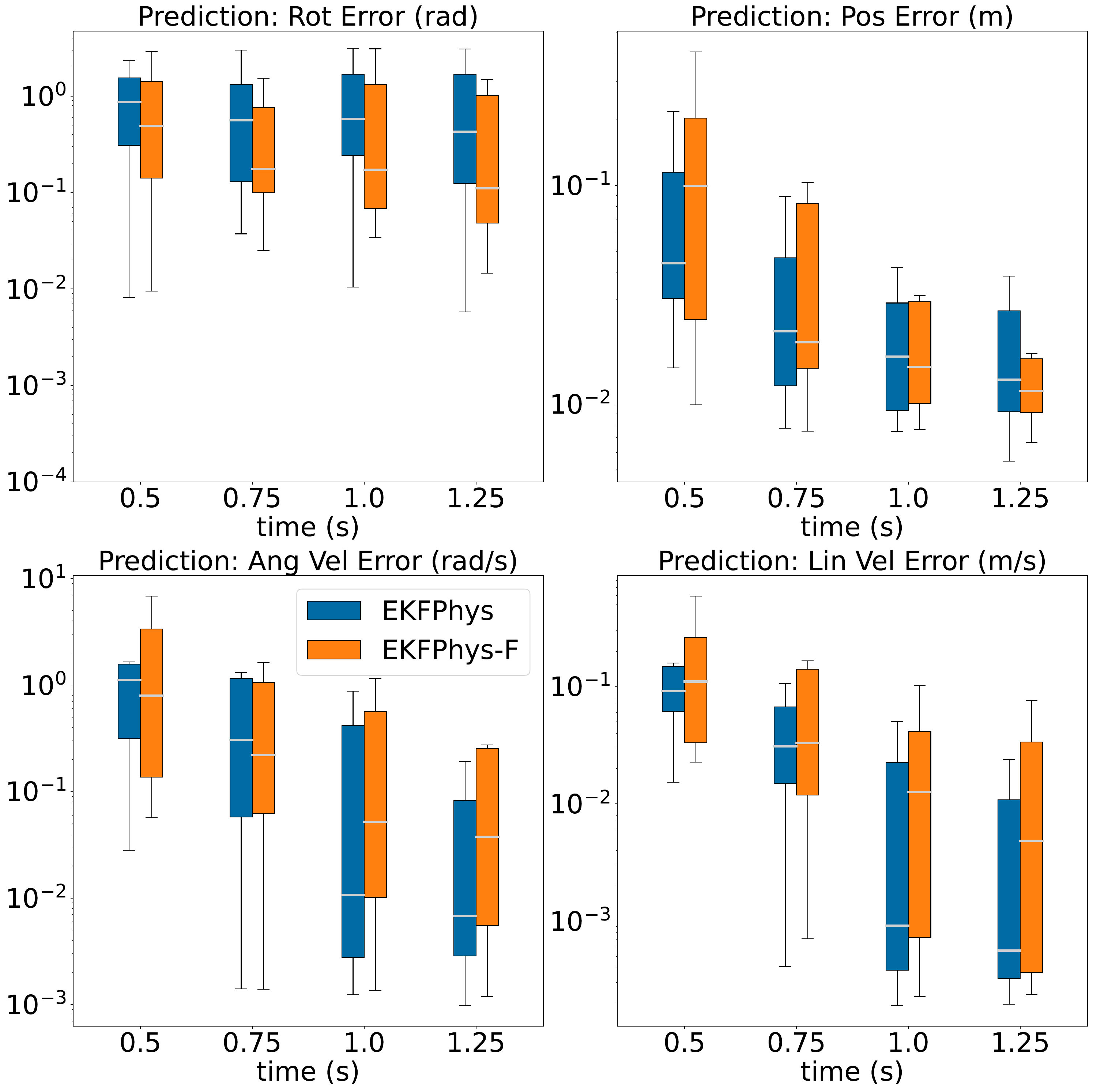}
    \caption{
    Prediction accuracy on single object sliding in real sequences for non-symmetric objects. 
    See Fig.~\ref{fig:trajectory_error_prediction_single_obj_nonsymm_synthetic} for labeling.
    }
    \label{fig:trajectory_error_prediction_single_obj_nonsymm_real}
\end{figure}
\begin{figure*}
    \centering
        \includegraphics[width=0.99\linewidth]{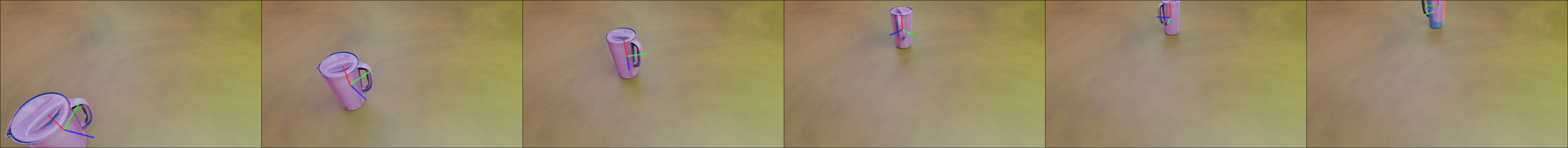}\\
        \includegraphics[width=0.99\linewidth]{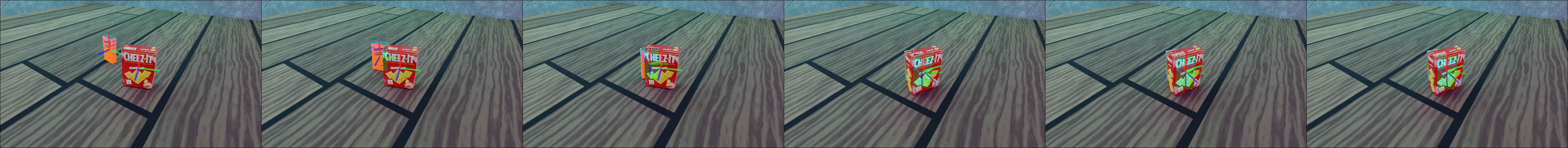}
        \includegraphics[width=0.99\linewidth]{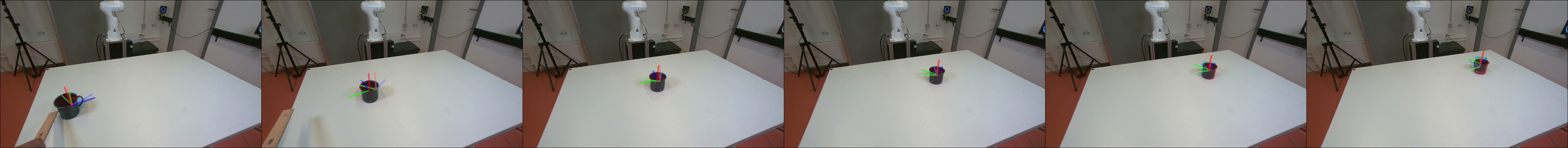}
    \caption{Qualitative results. Ground truth: transparent, estimate: saturated axes (RGB: XYZ). First row: Prediction results (EKFPhys, last two images) of a single object sliding scenario in synthetic dataset in which the poses from the detector are used as observations for 1.5s in a 2.5s trajectory.
    Second row: Prediction results (EKFPhys-F, last two images) of a two object sliding and colliding scenario
    where one object is partially occluded by the second object. Third row: Prediction results (EKFPhys, last two images) of a single object sliding scenario in real-world dataset in which the poses from the detector are used as observations for 0.5s in a 1.5s trajectory. 
    }
    \label{fig:real data sliding example}
\end{figure*}

For filtering, since the filter needs to converge from a zero mean Gaussian initialization, we show the average accuracy for several trajectory chunks running from a time after the sequence starts until the end of the sequence.
Results are shown in Figs.~\ref{fig:trajectory_error_filtering_single_obj_nonsymm_synthetic} and~\ref{fig:trajectory_error_filtering_single_obj_nonsymm_real} for synthetic and real sequences.
Our approach EKFPhys and the ablation EKFPhys-F filters the object's position and orientation with a  compromise in accuracy when compared to CIR while recovering linear and angular velocities more accurately when compared to the velocities obtained using finite differences with CIR poses.
Fig.~\ref{fig:friction_estimation_error_single_obj_nonsymm_synphys_non_damping} and Tab.~\ref{tab:fric_analysis} show the accuracy for recovering the friction value by EKFPhys.
We improve upon the baseline for median error both in synthetic and real sequences, but some of the outlier scenes in the synthetic scenes distorts the mean, as evident from the upper quantiles and the whiskers for "main" from Fig. \ref{fig:friction_estimation_error_single_obj_nonsymm_synphys_non_damping}.
In the real sequences, the differences between ground truth and estimated friction coefficient are presumably due to unmodelled physical effects that lead to estimation bias.
The improvements by EKFPhys are stronger on the real sequences for estimating linear and angular velocities, in which the noise in CIR pose estimates is larger which can be seen from Fig.~\ref{fig:trajectory_error_filtering_single_obj_nonsymm_real}. These noisy estimates of CIR also reflect the estimation of linear and angular velocities since they are calculated using finite differences. The CIR pose error decreases over time presumably due to less motion blur. CIR also has lower position recall as shown in  Fig.~\ref{fig:recall_filtering_single_obj_nonsymm_non_damping}.
Because of noisy rotation estimates on real images with many outliers, EKFPhys is optimized by Optuna to strongly smooth the rotations to be able to infer friction.
For the two-object synthetic scenes, the trajectory lengths are short before the objects collide. Thus EKFPhys doesn't get enough time to filter the friction for both the objects, hence, the drop in performance in rotation and position estimation as shown in Fig.~\ref{fig:trajectory_error_filtering_two_objs_nonsymm_synthetic}. Here, our variant EKFPhys-F performs better due to information about the ground truth friction.

Prediction is evaluated from various time steps until which the EKF filters with object pose measurements and then continues to predict without measurements until the end of the sequence.
In Figs.~\ref{fig:trajectory_error_prediction_single_obj_nonsymm_synthetic} and~\ref{fig:trajectory_error_prediction_single_obj_nonsymm_real} we provide results on synthetic and real sequences for single objects.
EKFPhys-F well predicts future motion with a small decrease in accuracy compared to filtering in this prediction horizon. EKFPhys makes larger prediction errors, especially, when starting to predict early in the sequence, since it also needs to estimate the friction coefficient from a sufficient amount of frames. Fig.~\ref{fig:real data sliding example} shows qualitative examples of prediction results.
Our methods allows to predict poses for frames in which CIR misses detections. For example, in the last three images in the second row of Fig.~\ref{fig:real data sliding example}, where the cracker box occludes the sugar box, CIR fails to detect the sugar box (see suppl. material) whereas EKFPhys-F tracks and predicts the pose based on collision dynamics with good accuracy.
We provide further results and videos in the suppl. material.

\section{Conclusions}
In summary, we have proposed a physics-based EKF filter which can track rigid body poses, velocities and friction parameters from RGB-D videos using a pre-trained 6D pose estimator as the observation model. We demonstrate that our method can recover the velocities more accurately than simple finite difference velocities from the detector poses. We also demonstrate that our method can filter scenes with occlusions and collision scenarios where the detector can miss detections. Finally, we evaluate prediction performance after several observations in synthetic and real-world experiments. Limitations of our method are that symmetries are not well handled yet with our state representation. The filter requires several frames with accurately observable object sliding motion to let the friction coefficient estimate converge. These missing detections due to occlusions, small object sizes, variations in lighting, etc., especially at the beginning of filtering can lead to inaccurate friction coefficients. Furthermore, there are unmodeled effects like damping. We add process noise to accommodate this, but these can still bias our filter like, for example, in the real-world sequences. Finally, our current python implementation is not real-time capable. We plan to address these limitations in future work, for instance, by a tighter integration of detection and filtering that can handle symmetries, using multiple cameras for more robust detections against occlusions and outliers and enhancing our filter's motion model with a learned model to capture the unmodeled effects or adding measurements such as tactile sensing. 

\paragraph{Acknowledgements}
This work was supported by Cyber Valley and the Max Planck Society. Michael Strecke has been supported by the German Federal Ministry of Education and Research (BMBF) through the Tuebingen AI Center (FKZ: 01IS18039B). This work was partially funded by the Deutsche Forschungsgemeinschaft (DFG, German Research Foundation) -- Projektnummer 466606396 (STU 771/1-1). Rama Krishna Kandukuri has been supported by Cyber Valley Research Fund project CyVy-RF-2019-06. The authors thank the International Max Planck Research School for Intelligent Systems (IMPRS-IS) for supporting Michael Strecke and Rama Krishna Kandukuri.

{
    \small
    \bibliographystyle{ieeenat_fullname}
    \bibliography{main}

\begin{thebibliography}{61}
\providecommand{\natexlab}[1]{#1}
\providecommand{\url}[1]{\texttt{#1}}
\expandafter\ifx\csname urlstyle\endcsname\relax
  \providecommand{\doi}[1]{doi: #1}\else
  \providecommand{\doi}{doi: \begingroup \urlstyle{rm}\Url}\fi

\bibitem[Akiba et~al.(2019)Akiba, Sano, Yanase, Ohta, and Koyama]{optuna_2019}
Takuya Akiba, Shotaro Sano, Toshihiko Yanase, Takeru Ohta, and Masanori Koyama.
\newblock Optuna: A next-generation hyperparameter optimization framework.
\newblock In \emph{Proceedings of the 25th {ACM} {SIGKDD} International
  Conference on Knowledge Discovery and Data Mining}, 2019.

\bibitem[Amos and Kolter(2017)]{Amos_Kolter_2017}
Brandon Amos and J.~Zico Kolter.
\newblock Optnet: Differentiable optimization as a layer in neural networks.
\newblock In \emph{Proceedings of the International Conference on Machine
  Learning (ICML)}, page 136–145, 2017.

\bibitem[Becker et~al.(2019)Becker, Pandya, Gebhardt, Zhao, Taylor, and
  Neumann]{becker2019_rkn}
Philipp Becker, Harit Pandya, Gregor H.~W. Gebhardt, Cheng Zhao, C.~James
  Taylor, and Gerhard Neumann.
\newblock Recurrent kalman networks: Factorized inference in high-dimensional
  deep feature spaces.
\newblock In \emph{Proceedings of the International Conference on Machine
  Learning (ICML)}, 2019.

\bibitem[Calli et~al.(2015{\natexlab{a}})Calli, Singh, Walsman, Srinivasa,
  Abbeel, and Dollar]{calli2015_ycb}
Berk Calli, Arjun Singh, Aaron Walsman, Siddhartha Srinivasa, Pieter Abbeel,
  and Aaron~M. Dollar.
\newblock The {YCB} object and model set: Towards common benchmarks for
  manipulation research.
\newblock In \emph{Proceedings of the International Conference on Advanced
  Robotics ({ICAR})}, 2015{\natexlab{a}}.

\bibitem[Calli et~al.(2015{\natexlab{b}})Calli, Walsman, Singh, Srinivasa,
  Abbeel, and Dollar]{calli2015_benchmarking}
Berk Calli, Aaron Walsman, Arjun Singh, Siddhartha Srinivasa, Pieter Abbeel,
  and Aaron~M. Dollar.
\newblock Benchmarking in manipulation research: Using the yale-{CMU}-berkeley
  object and model set.
\newblock \emph{{IEEE} Robotics {\&} Automation Magazine}, 22\penalty0
  (3):\penalty0 36--52, 2015{\natexlab{b}}.

\bibitem[Chang et~al.(2015)Chang, Funkhouser, Guibas, Hanrahan, Huang, Li,
  Savarese, Savva, Song, Su, Xiao, Yi, and Yu]{shapenet2015}
Angel~X. Chang, Thomas Funkhouser, Leonidas Guibas, Pat Hanrahan, Qixing Huang,
  Zimo Li, Silvio Savarese, Manolis Savva, Shuran Song, Hao Su, Jianxiong Xiao,
  Li Yi, and Fisher Yu.
\newblock {ShapeNet: An Information-Rich {3D} Model Repository}.
\newblock Technical Report arXiv:1512.03012 [cs.GR], Stanford University ---
  Princeton University --- Toyota Technological Institute at Chicago, 2015.

\bibitem[Choi and Christensen(2010)]{choi2010_real}
Changhyun Choi and Henrik~I. Christensen.
\newblock Real-time {3D} model-based tracking using edge and keypoint features
  for robotic manipulation.
\newblock In \emph{Proceedings of the {IEEE} International Conference on
  Robotics and Automation (ICRA)}, 2010.
\newblock ISSN: 1050-4729.

\bibitem[Choi and Christensen(2012)]{choi2012_3d}
Changhyun Choi and Henrik~I. Christensen.
\newblock 3d textureless object detection and tracking: An edge-based approach.
\newblock In \emph{Proceedings of the {IEEE}/{RSJ} International Conference on
  Intelligent Robots and Systems (IROS)}, 2012.

\bibitem[Cleac'h et~al.(2022)Cleac'h, Yu, Guo, Howell, Gao, Wu, Manchester, and
  Schwager]{cleach2022_differentiable}
Simon~Le Cleac'h, Hong Yu, Michelle Guo, Taylor~A. Howell, Ruohan Gao, Jiajun
  Wu, Zachary Manchester, and Mac Schwager.
\newblock Differentiable physics simulation of dynamics-augmented neural
  objects.
\newblock \emph{IEEE Robotics and Automation Letters}, 8:\penalty0 2780--2787,
  2022.

\bibitem[Cline(2002)]{Cline_2002}
Michael~Bradley Cline.
\newblock \emph{Rigid body simulation with contact and constraints}.
\newblock PhD thesis, University of British Columbia, 2002.

\bibitem[Community(2018)]{blender}
Blender~Online Community.
\newblock \emph{Blender - a {3D} modelling and rendering package}.
\newblock Blender Foundation, Stiching Blender Foundation, Amsterdam, 2018.

\bibitem[Coumans and Bai(2016--2021)]{pybullet}
Erwin Coumans and Yunfei Bai.
\newblock Pybullet, a python module for physics simulation for games, robotics
  and machine learning.
\newblock \url{http://pybullet.org}, 2016--2021.

\bibitem[de~Avila Belbute-Peres et~al.(2018)de~Avila Belbute-Peres, Smith,
  Allen, Tenenbaum, and Kolter]{lcp_physics_2d}
Filipe de Avila Belbute-Peres, Kevin Smith, Kelsey Allen, Josh Tenenbaum, and
  J.~Zico Kolter.
\newblock End-to-end differentiable physics for learning and control.
\newblock In \emph{Proceedings of the Advances in Neural Information Processing
  Systems (NIPS)}, 2018.

\bibitem[Deng et~al.(2019)Deng, Mousavian, Xiang, Xia, Bretl, and
  Fox]{deng2019_poserbpf}
Xinke Deng, Arsalan Mousavian, Yu Xiang, Fei Xia, Timothy Bretl, and Dieter
  Fox.
\newblock {PoseRBPF}: A rao-blackwellized particle filter for {6D} object pose
  estimation.
\newblock In \emph{Proceedings of the Robotics: Science and Systems (RSS)},
  2019.

\bibitem[Deng et~al.(2022)Deng, Geng, Bretl, Xiang, and Fox]{deng2022_icaps}
Xinke Deng, Junyi Geng, Timothy Bretl, Yu Xiang, and Dieter Fox.
\newblock {iCaps}: Iterative category-level object pose and shape estimation.
\newblock \emph{IEEE Robotics and Automation Letters (RAL)}, 7\penalty0
  (2):\penalty0 1784--1791, 2022.

\bibitem[Eckenhoff et~al.(2019)Eckenhoff, Yang, Geneva, and
  Huang]{eckenhoff2019_tightly}
Kevin Eckenhoff, Yulin Yang, Patrick Geneva, and Guoquan Huang.
\newblock Tightly-coupled visual-inertial localization and {3D} rigid-body
  target tracking.
\newblock \emph{IEEE Robotics and Automation Letters (RAL)}, 4\penalty0
  (2):\penalty0 1541--1548, 2019.

\bibitem[Eckenhoff et~al.(2020)Eckenhoff, Geneva, Merrill, and
  Huang]{eckenhoff2020_schmidt}
Kevin Eckenhoff, Patrick Geneva, Nathaniel Merrill, and Guoquan Huang.
\newblock Schmidt-{EKF}-based visual-inertial moving object tracking.
\newblock In \emph{Proceedings of the {IEEE} International Conference on
  Robotics and Automation ({ICRA})}, 2020.
\newblock ISSN: 2577-087X.

\bibitem[Evans(1990)]{evans1990_kalman}
R. Evans.
\newblock Kalman filtering of pose estimates in applications of the {RAPID}
  video rate tracker.
\newblock In \emph{Proceedings of the British Machine Vision Conference
  (BMVC)}, 1990.

\bibitem[Gao et~al.(2003)Gao, Hou, Tang, and Cheng]{gao2003_complete}
Xiao-Shan Gao, Xiao-Rong Hou, Jianliang Tang, and Hang-Fei Cheng.
\newblock Complete solution classification for the perspective-three-point
  problem.
\newblock \emph{IEEE Transactions on Pattern Analysis and Machine
  Intelligence}, 25\penalty0 (8):\penalty0 930--943, 2003.

\bibitem[Garon and Lalonde(2017)]{garon2017_deep}
Mathieu Garon and Jean-Francois Lalonde.
\newblock Deep 6-{DOF} tracking.
\newblock \emph{IEEE Transactions on Visualization and Computer Graphics},
  23\penalty0 (11):\penalty0 2410--2418, 2017.

\bibitem[Hafner et~al.(2019)Hafner, Lillicrap, Fischer, Villegas, Ha, Lee, and
  Davidson]{hafner2019_planet}
Danijar Hafner, Timothy Lillicrap, Ian Fischer, Ruben Villegas, David Ha,
  Honglak Lee, and James Davidson.
\newblock Learning latent dynamics for planning from pixels.
\newblock In \emph{Proceedings of the International Conference on Machine
  Learning (ICML)}, 2019.

\bibitem[Harris and Stennett(1990)]{harris1990_rapid}
C. Harris and C. Stennett.
\newblock {RAPID} - a video rate object tracker.
\newblock In \emph{Proceedings of the British Machine Vision Conference
  (BMVC)}, Oxford, 1990.

\bibitem[He et~al.(2017)He, Gkioxari, Doll\'{a}r, and Girshick]{he2017_mask}
Kaiming He, Georgia Gkioxari, Piotr Doll\'{a}r, and Ross Girshick.
\newblock Mask {R-CNN}.
\newblock In \emph{Proceedings of the {IEEE} International Conference on
  Computer Vision ({ICCV})}, 2017.

\bibitem[He et~al.(2020)He, Sun, Huang, Liu, Fan, and Sun]{he2020_pvn3d}
Yisheng He, Wei Sun, Haibin Huang, Jianran Liu, Haoqiang Fan, and Jian Sun.
\newblock {PVN}3{D}: A deep point-wise {3D} keypoints voting network for {6DoF}
  pose estimation.
\newblock In \emph{Proceedings of the {IEEE}/{CVF} Conference on Computer
  Vision and Pattern Recognition ({CVPR})}, 2020.

\bibitem[He et~al.(2021)He, Huang, Fan, Chen, and Sun]{he2021_ffb6d}
Yisheng He, Haibin Huang, Haoqiang Fan, Qifeng Chen, and Jian Sun.
\newblock {FFB}6{D}: A full flow bidirectional fusion network for {6D} pose
  estimation.
\newblock In \emph{Proceedings of the {IEEE}/{CVF} Conference on Computer
  Vision and Pattern Recognition ({CVPR})}, 2021.

\bibitem[Heiden et~al.(2022)Heiden, Liu, Vineet, Coumans, and
  Sukhatme]{heiden2022_artrigboddyn}
Eric Heiden, Ziang Liu, Vibhav Vineet, Erwin Coumans, and Gaurav~S. Sukhatme.
\newblock Inferring articulated rigid body dynamics from rgbd video.
\newblock In \emph{Proceedings of the {IEEE/RSJ} International Conference on
  Intelligent Robots and Systems (IROS)}, 2022.

\bibitem[Hoda{\v{n}} et~al.(2018)Hoda{\v{n}}, Michel, Brachmann, Kehl, Buch,
  Kraft, Drost, Vidal, Ihrke, Zabulis, Sahin, Manhardt, Tombari, Kim, Matas,
  and Rother]{hodan2018_bop}
Tom{\'{a}}{\v{s}} Hoda{\v{n}}, Frank Michel, Eric Brachmann, Wadim Kehl,
  Anders~Glent Buch, Dirk Kraft, Bertram Drost, Joel Vidal, Stephan Ihrke,
  Xenophon Zabulis, Caner Sahin, Fabian Manhardt, Federico Tombari, Tae-Kyun
  Kim, Ji{\v{r}}{\'{\i}} Matas, and Carsten Rother.
\newblock {BOP}: Benchmark for {6D} object pose estimation.
\newblock In \emph{Proceedings of the European Conference on Computer Vision
  ({ECCV})}. 2018.

\bibitem[Hoda{\v{n}} et~al.(2020)Hoda{\v{n}}, Sundermeyer, Drost, Labb{\'{e}},
  Brachmann, Michel, Rother, and Matas]{hodan2020_bop}
Tom{\'{a}}{\v{s}} Hoda{\v{n}}, Martin Sundermeyer, Bertram Drost, Yann
  Labb{\'{e}}, Eric Brachmann, Frank Michel, Carsten Rother, and
  Ji{\v{r}}{\'{\i}} Matas.
\newblock {BOP} challenge 2020 on {6D} object localization.
\newblock In \emph{Proceedings of the European Conference on Computer Vision
  Workshops ({ECCVW})}. 2020.

\bibitem[Irshad et~al.(2022)Irshad, Zakharov, Ambrus, Kollar, Kira, and
  Gaidon]{irshad2022_shapo}
Muhammad~Zubair Irshad, Sergey Zakharov, Rares Ambrus, Thomas Kollar, Zsolt
  Kira, and Adrien Gaidon.
\newblock {ShAPO}: {Implicit} {Representations} for {Multi}-object {Shape},
  {Appearance}, and {Pose} {Optimization}.
\newblock In \emph{Proceedings of the European Conference on Computer Vision
  ({ECCV})}, Cham, 2022.

\bibitem[Jongeneel et~al.(2022)Jongeneel, Bernardino, van~de Wouw, and
  Saccon]{jongeneel2022_model}
M.J. Jongeneel, Alexandre Bernardino, Nathan van~de Wouw, and Alessandro
  Saccon.
\newblock Model-{Based} {6D} {Visual} {Object} {Tracking} with {Impact}
  {Collision} {Models}.
\newblock In \emph{Proceedings of the American {Control} {Conference} ({ACC})},
  Atlanta, United States, 2022.

\bibitem[Kandukuri et~al.(2021)Kandukuri, Achterhold, Moeller, and
  Stueckler]{kandukuri2021_physical}
Rama~Krishna Kandukuri, Jan Achterhold, Michael Moeller, and Joerg Stueckler.
\newblock Physical representation learning and parameter identification from
  video using differentiable physics.
\newblock \emph{International Journal of Computer Vision (IJCV)}, 130\penalty0
  (1):\penalty0 3--16, 2021.

\bibitem[Kehl et~al.(2017)Kehl, Tombari, Ilic, and Navab]{kehl2017_real}
Wadim Kehl, Federico Tombari, Slobodan Ilic, and Nassir Navab.
\newblock Real-time 3d model tracking in color and depth on a single {CPU}
  core.
\newblock In \emph{Proceedings of the {IEEE} Conference on Computer Vision and
  Pattern Recognition ({CVPR})}, 2017.

\bibitem[Krishna~Murthy et~al.(2021)Krishna~Murthy, Macklin, Golemo, Voleti,
  Petrini, Weiss, Considine, Parent-Levesque, Xie, Erleben, Paull, Shkurti,
  Nowrouzezahrai, and Fidler]{krishnamurthy2021_gradsim}
Jatavallabhula Krishna~Murthy, Miles Macklin, Florian Golemo, Vikram Voleti,
  Linda Petrini, Martin Weiss, Breandan Considine, Jerome Parent-Levesque,
  Kevin Xie, Kenny Erleben, Liam Paull, Florian Shkurti, Derek Nowrouzezahrai,
  and Sanja Fidler.
\newblock {gradSim}: Differentiable simulation for system identification and
  visuomotor control.
\newblock In \emph{Proceedings of the International Conference on Learning
  Representations (ICLR)}, 2021.

\bibitem[Labb{\'{e}} et~al.(2020)Labb{\'{e}}, Carpentier, Aubry, and
  Sivic]{labbe2020_cosypose}
Yann Labb{\'{e}}, Justin Carpentier, Mathieu Aubry, and Josef Sivic.
\newblock {CosyPose}: Consistent multi-view multi-object {6D} pose estimation.
\newblock In \emph{Proceedings of the European Conference on Computer Vision
  ({ECCV})}. 2020.

\bibitem[Lepetit et~al.(2008)Lepetit, Moreno-Noguer, and Fua]{lepetit2008_epnp}
Vincent Lepetit, Francesc Moreno-Noguer, and Pascal Fua.
\newblock {EPnP}: An accurate o(n) solution to the {PnP} problem.
\newblock \emph{International Journal of Computer Vision (IJCV)}, 81\penalty0
  (2):\penalty0 155--166, 2008.

\bibitem[Li et~al.(2019)Li, Wang, Ji, Xiang, and Fox]{li2018_deepim}
Yi Li, Gu Wang, Xiangyang Ji, Yu Xiang, and Dieter Fox.
\newblock {DeepIM}: Deep iterative matching for {6D} pose estimation.
\newblock \emph{International Journal of Computer Vision (IJCV)}, 128\penalty0
  (3):\penalty0 657--678, 2019.

\bibitem[Lipson et~al.(2022)Lipson, Teed, Goyal, and Deng]{lipson2022_coupled}
Lahav Lipson, Zachary Teed, Ankit Goyal, and Jia Deng.
\newblock Coupled iterative refinement for {6D} multi-object pose estimation.
\newblock In \emph{Proceedings of the {IEEE}/{CVF} Conference on Computer
  Vision and Pattern Recognition ({CVPR})}, 2022.

\bibitem[Manhardt et~al.(2018)Manhardt, Kehl, Navab, and
  Tombari]{manhardt2018_deep}
Fabian Manhardt, Wadim Kehl, Nassir Navab, and Federico Tombari.
\newblock Deep model-based {6D} pose refinement in {RGB}.
\newblock In \emph{Proceedings of the European Conference on Computer Vision
  ({ECCV})}. 2018.

\bibitem[Melax et~al.(2013)Melax, Keselman, and
  Orsten]{melax2013_dynamicsbasedskeletal}
Stan Melax, Leonid Keselman, and Sterling Orsten.
\newblock Dynamics based {3D} skeletal hand tracking.
\newblock In \emph{Proceedings of the Graphics Interface}, CAN, 2013.

\bibitem[Murthy et~al.(2021)Murthy, Macklin, Golemo, Voleti, Petrini, Weiss,
  Considine, Parent{-}L{\'{e}}vesque, Xie, Erleben, Paull, Shkurti,
  Nowrouzezahrai, and Fidler]{murthy2021_gradsim}
J.~Krishna Murthy, Miles Macklin, Florian Golemo, Vikram Voleti, Linda Petrini,
  Martin Weiss, Breandan Considine, J{\'{e}}r{\^{o}}me Parent{-}L{\'{e}}vesque,
  Kevin Xie, Kenny Erleben, Liam Paull, Florian Shkurti, Derek Nowrouzezahrai,
  and Sanja Fidler.
\newblock gradsim: Differentiable simulation for system identification and
  visuomotor control.
\newblock In \emph{Proceedings of the International Conference on Learning
  Representations (ICLR)}, 2021.

\bibitem[Qiu et~al.(2019)Qiu, Qin, Gao, and Shen]{qiu2019_tracking}
Kejie Qiu, Tong Qin, Wenliang Gao, and Shaojie Shen.
\newblock Tracking 3d motion of dynamic objects using monocular visual-inertial
  sensing.
\newblock \emph{IEEE Transactions on Robotics}, 35\penalty0 (4):\penalty0
  799--816, 2019.

\bibitem[R{\"u}nz and Agapito(2017)]{ruenz2017_co}
Martin R{\"u}nz and Lourdes Agapito.
\newblock {Co-Fusion}: Real-time segmentation, tracking and fusion of multiple
  objects.
\newblock In \emph{Proceedings of the {IEEE} International Conference on
  Robotics and Automation ({ICRA})}, 2017.

\bibitem[R{\"u}nz et~al.(2018)R{\"u}nz, Buffier, and
  Agapito]{ruenz2018_maskfusion}
Martin R{\"u}nz, Maud Buffier, and Lourdes Agapito.
\newblock {MaskFusion}: Real-time recognition, tracking and reconstruction of
  multiple moving objects.
\newblock In \emph{Proceedings of the {IEEE} International Symposium on Mixed
  and Augmented Reality ({ISMAR})}, 2018.

\bibitem[Sanchez{-}Gonzalez et~al.(2020)Sanchez{-}Gonzalez, Godwin, Pfaff,
  Ying, Leskovec, and Battaglia]{sanchez-gonzalez2020_gnnphys}
Alvaro Sanchez{-}Gonzalez, Jonathan Godwin, Tobias Pfaff, Rex Ying, Jure
  Leskovec, and Peter~W. Battaglia.
\newblock Learning to simulate complex physics with graph networks.
\newblock In \emph{Proceedings of the International Conference on Machine
  Learning (ICML)}, 2020.

\bibitem[Schmidt et~al.(2015)Schmidt, Newcombe, and Fox]{schmidt2015_dart}
Tanner Schmidt, Richard~A. Newcombe, and Dieter Fox.
\newblock {DART:} dense articulated real-time tracking with consumer depth
  cameras.
\newblock \emph{Autonomous Robots}, 39\penalty0 (3):\penalty0 239--258, 2015.

\bibitem[Smith(2008)]{ode2008}
Russell Smith.
\newblock Open dynamics engine, 2008.
\newblock http://www.ode.org/.

\bibitem[Strecke and Stueckler(2019)]{strecke2019_em}
Michael Strecke and Joerg Stueckler.
\newblock {EM}-{Fusion}: Dynamic object-level {SLAM} with probabilistic data
  association.
\newblock In \emph{Proceedings of the {IEEE}/{CVF} International Conference on
  Computer Vision ({ICCV})}, 2019.

\bibitem[Strecke and Stueckler(2020)]{strecke2020_where}
Michael Strecke and Joerg Stueckler.
\newblock Where does it end? {\textendash} reasoning about hidden surfaces by
  object intersection constraints.
\newblock In \emph{Proceedings of the {IEEE}/{CVF} Conference on Computer
  Vision and Pattern Recognition ({CVPR})}, 2020.

\bibitem[Strecke and Stueckler(2021)]{strecke2021_diffsdfsim}
Michael Strecke and Joerg Stueckler.
\newblock {DiffSDFSim}: Differentiable rigid-body dynamics with implicit
  shapes.
\newblock In \emph{Proceedings of the International Conference on {3D} Vision
  (3DV)}, 2021.

\bibitem[Tan and Le(2019)]{tan2019_efficientnet}
Mingxing Tan and Quoc~V. Le.
\newblock {EfficientNet}: Rethinking model scaling for convolutional neural
  networks.
\newblock \emph{Proceedings of the International Conference on Machine Learning
  (ICML)}, 2019.

\bibitem[Teed and Deng(2020)]{teed2020_raft}
Zachary Teed and Jia Deng.
\newblock {RAFT}: Recurrent all-pairs field transforms for optical flow.
\newblock In \emph{Proceedings of the European Conference on Computer Vision
  {ECCV}}. 2020.

\bibitem[Vacchetti et~al.(2004)Vacchetti, Lepetit, and
  Fua]{vacchetti2004_combining}
L. Vacchetti, V. Lepetit, and P. Fua.
\newblock Combining edge and texture information for real-time accurate {3D}
  camera tracking.
\newblock In \emph{Proceedings of the {IEEE} and {ACM} International Symposium
  on Mixed and Augmented Reality}, 2004.

\bibitem[Varin and Kuindersma(2020)]{varin2020_constrained}
Patrick Varin and Scott Kuindersma.
\newblock A constrained kalman filter for rigid body systems with frictional
  contact.
\newblock In \emph{Springer Proceedings in Advanced Robotics}, Cham, 2020.

\bibitem[Wada et~al.(2020)Wada, Sucar, James, Lenton, and
  Davison]{wada2020_morefusion}
Kentaro Wada, Edgar Sucar, Stephen James, Daniel Lenton, and Andrew~J. Davison.
\newblock {MoreFusion}: Multi-object reasoning for {6D} pose estimation from
  volumetric fusion.
\newblock In \emph{Proceedings of the {IEEE}/{CVF} Conference on Computer
  Vision and Pattern Recognition ({CVPR})}, 2020.

\bibitem[Wang et~al.(2007)Wang, Thorpe, Thrun, Hebert, and
  Durrant-Whyte]{wang2007_simultaneous}
Chieh-Chih Wang, Charles Thorpe, Sebastian Thrun, Martial Hebert, and Hugh
  Durrant-Whyte.
\newblock Simultaneous localization, mapping and moving object tracking.
\newblock \emph{The International Journal of Robotics Research (IJRR)},
  26\penalty0 (9):\penalty0 889--916, 2007.

\bibitem[Wang et~al.(2020)Wang, Yang, Stueckler, and
  Cremers]{wang2020_directshape}
Rui Wang, Nan Yang, Joerg Stueckler, and Daniel Cremers.
\newblock {DirectShape}: Direct photometric alignment of shape priors for
  visual vehicle pose and shape estimation.
\newblock In \emph{Proceedings of the {IEEE} International Conference on
  Robotics and Automation ({ICRA})}, 2020.

\bibitem[Wen et~al.(2020)Wen, Mitash, Ren, and Bekris]{wen2020_se3}
Bowen Wen, Chaitanya Mitash, Baozhang Ren, and Kostas~E. Bekris.
\newblock se(3)-{TrackNet}: Data-driven {6D} pose tracking by calibrating image
  residuals in synthetic domains.
\newblock In \emph{Proceedings of the {IEEE}/{RSJ} International Conference on
  Intelligent Robots and Systems ({IROS})}, 2020.
\newblock ISSN: 2153-0866.

\bibitem[Wu et~al.(2015)Wu, Yildirim, Lim, Freeman, and
  Tenenbaum]{wu2015_galileo}
Jiajun Wu, Ilker Yildirim, Joseph~J Lim, Bill Freeman, and Josh Tenenbaum.
\newblock Galileo: Perceiving physical object properties by integrating a
  physics engine with deep learning.
\newblock In \emph{Proceedings of the Advances in Neural Information Processing
  Systems (NIPS)}, 2015.

\bibitem[Xiang et~al.(2018)Xiang, Schmidt, Narayanan, and
  Fox]{xiang2018_posecnn}
Yu Xiang, Tanner Schmidt, Venkatraman Narayanan, and Dieter Fox.
\newblock {PoseCNN}: A convolutional neural network for {6D} object pose
  estimation in cluttered scenes.
\newblock In \emph{Proceedings of the Robotics: Science and Systems (RSS)},
  2018.

\bibitem[Xu et~al.(2019)Xu, Li, Tzoumanikas, Bloesch, Davison, and
  Leutenegger]{xu2019_mid}
Binbin Xu, Wenbin Li, Dimos Tzoumanikas, Michael Bloesch, Andrew Davison, and
  Stefan Leutenegger.
\newblock {MID-Fusion}: Octree-based object-level multi-instance dynamic
  {SLAM}.
\newblock In \emph{Proceedings of the {IEEE} International Conference on
  Robotics and Automation ({ICRA})}, 2019.

\bibitem[Xu et~al.(2022)Xu, Papallas, and Dogar]{xu2022_physicsbasedtracking}
Zisong Xu, Rafael Papallas, and Mehmet~Remzi Dogar.
\newblock Real-time physics-based object pose tracking during non-prehensile
  manipulation.
\newblock \emph{CoRR}, abs/2211.13572, 2022.

\end{thebibliography}
}

\onecolumn
\begin{center}
    \textbf{\Large{Supplementary Material\\}}
\end{center}
\vspace{0.5cm}

\setcounter{section}{0}
\renewcommand{\thesection}{\Alph{section}}

\section{EKF Implementation}
This section describes the implementation of the EKF formulation in PyTorch as described in section 4.3 of the main paper.

\subsection{CIR Configuration}

We have used two different configurations for CIR to estimate poses on synthetic and real-world datasets. CIR's iterative refinement can be controlled with three different parameters: solver steps (Gauss-Newton iterations), inner loops (Raft-SE3 iterations) and outer loops (re-rendering iterations).

\begin{table}[tbh]
    \centering
    \begin{tabular}{ccc}
        \toprule
         & Synthetic & Real \\
         \cmidrule{2-3}
         Solver steps & 3 & 10 \\
         Inner loops & 10 & 40 \\
         Outer loops & 3 & 5 \\
         \bottomrule
    \end{tabular}
    \caption{CIR's iterative refinement configuration}
    \label{tab:my_label}
\end{table}

More iterations were used for pose estimation using CIR on real-world datasets because it's more challenging to align the meshes since the Raft-SE3 model is only trained on synthetic data.

\subsection{Initialization}
We initialize the pose ($\mathbf{x}_0 = (\mathbf{p}_0, \mathbf{R}_0)$) with the detections from CIR and the velocity ($ \boldsymbol{\xi}_0 = (\mathbf{v}_0, \boldsymbol{\omega}_0)$) using the finite differences of poses in the first two frames multiplied by the frame rate of CIR (30 Hz). The filter parameters, $\mathbf{\Sigma_0^{p}}, \mathbf{\Sigma_0^{R}}, \mathbf{\Sigma_0^{v}}, \mathbf{\Sigma_0^{\omega}}, \mathbf{\Sigma_0^{\theta}}, \mathbf{S}^{\{\mathbf{p}, \mathbf{R}, \mathbf{v}, \mathbf{\omega}\}}, \mathbf{S}^{\theta}, \mathbf{Q}^{p}, \mathbf{Q}^{R}$ and $\boldsymbol{\zeta}$ are tuned on validation set using Optuna \cite{optuna_2019}, a Bayesian hyperparameter tuning framework. The friction values are initialized at 0.0, assuming no information about the friction.
Table \ref{tab: filter initialization parameters} provides an overview of these parameters.

\begin{table}[!h]
\footnotesize
    \centering
    \begin{tabular}{ccccc}
        \toprule
        & \multicolumn{2}{c}{Synthetic} & \multicolumn{2}{c}{Real-world} \\
        \cmidrule(lr){2-3} \cmidrule(lr){4-5}
        & EKFPhys & EKFPhys-F & EKFPhys & EKFPhys-F \\
        \cmidrule(lr){2-2} \cmidrule(lr){3-3} \cmidrule(lr){4-4} \cmidrule(lr){5-5}
        
        Initial translation covariance ($\mathbf{\Sigma_0^{p}}$) & 88.253 & 3.68122 & 0.0138 & 5.13 \\
        Initial rotation covariance ($\mathbf{\Sigma_0^{R}}$) & 0.00469 & 0.26089 & 1.0902 & 47.6 \\
        Initial linear velocity covariance ($\mathbf{\Sigma_0^{v}}$) & 549.57 & 0.00011 & 546.27 & 12.93 \\
        Initial angular velocity covariance ($\mathbf{\Sigma_0^{\omega}}$) & 0.260 & 209.502 & 246.794 & 141.8 \\
        Initial friction covariance ($\mathbf{\Sigma_0^{\theta}}$) & 0.07372 & - & 0.21 & - \\
        State process noise ($\mathbf{S}^{\{\mathbf{p}, \mathbf{R}, \mathbf{v}, \mathbf{\omega}\}}$) & 0.00011 & 0.0078 & 3.549e-5 & 0.0012 \\
        Friction process noise ($\mathbf{S^{\theta}}$) & 3.49e-6 & - &  2.47e-5 & -\\
        Translation observation noise ($\mathbf{Q}^{p}$) & 0.00025 & 0.00018 & 6.91e-5 & 1.12e-5 \\
        Rotation observation noise ($\mathbf{Q}^{R}$) & 0.00092 & 0.00021 & 0.2 & 0.0005 \\
        Gating threshold ($\boldsymbol{\zeta}$) & 340 & 160 & 460 & 120 \\
        \bottomrule
    \end{tabular}
    \caption{EKF initialization parameters}
    \label{tab: filter initialization parameters}
\end{table}

\subsection{Prediction step}
The prediction step in our filter is performed at 1/60\,sec intervals, i.e., two prediction steps for each correction step (image rate). 
The state $\mathbf{s}_t$ for the EKFPhys-F model is given by $\mathbf{s}_t = (\mathbf{x}_t, \boldsymbol{\xi}_t)$ where pose, $\mathbf{x}_t = (\mathbf{p}_t, \mathbf{R}_t)$ and twist, $\boldsymbol{\xi}_t = (\mathbf{v}_t, \boldsymbol{\omega}_t)$. For the EKFPhys model, additionally we also filter the coefficient of friction to the state, $\mathbf{s}_t = (\mathbf{x}_t, \boldsymbol{\xi}_t, \boldsymbol{\theta}_t)$.\\

The twist for the next time step for EKFPhys and EKFPhys-F models \cite{Cline_2002}, \cite{lcp_physics_2d} are given by 
\begin{equation}
	\begin{pmatrix}
		0\\
		\mathbf{s}\\
		0
	\end{pmatrix}
	+
	\begin{pmatrix}
		\mathbf{M} && \mathbf{G}^\top && \mathbf{A}^\top\\
		\mathbf{G} && \mathbf{F}(\boldsymbol{\theta}_t) && 0\\
		\mathbf{A} && 0 && 0\\
	\end{pmatrix}
	\begin{pmatrix}
		-\boldsymbol{\xi}_{t+\Delta t}\\
		\mathbf{z}\\
		\mathbf{y}\\
	\end{pmatrix}
	=
	\begin{pmatrix}
		-\mathbf{q}\\
		\mathbf{m}\\
		0\\
	\end{pmatrix}\\
\label{eq: LCP formulation}
\end{equation}
\begin{equation*}
    \mathrm{subject\;to\;} \mathbf{s} \geq 0,\;\mathbf{z} \geq 0,\; \mathbf{s}^T\mathbf{z} = 0
\end{equation*}
\begin{equation*}
    \textrm{where}\;\;\mathbf{q} = \mathbf{M}\boldsymbol{\xi}_{t} + \Delta t\mathbf{f}_{ext}.
\end{equation*}

The twist for the next time step is obtained by solving equation \eqref{eq: LCP formulation} using the Primal-Dual interior point method \cite{Amos_Kolter_2017,lcp_physics_2d}. 
The pose for the next-time step is calculated as
\begin{equation}
    \mathbf{x}_{t+\Delta t} = \mathbf{x}_t \boxplus \boldsymbol{\xi}_{t+\Delta t}\Delta t,
    \label{eq: pose update}
\end{equation}
where equation \eqref{eq: pose update} can be written as
\begin{align*}
    \mathbf{p}_{t+\Delta t} = \mathbf{p}_t + \mathbf{v}_{t+\Delta t}\Delta t \\
    \mathbf{R}_{t+\Delta t} = e^{[\boldsymbol{\omega_{t+\Delta t}}\Delta t]} \mathbf{R}_t.
\end{align*}

The state-transition function for the models are given by
\begin{align*}
    \textrm{EKFPhys} : g(\mathbf{s}_t) &= (\mathbf{x}_{t+\Delta t}, \boldsymbol{\xi}_{t+\Delta t}, \boldsymbol{\theta}_{t+\Delta t}),\;\;\boldsymbol{\theta}_{t+\Delta t}:=\boldsymbol{\theta}_t\\
    \textrm{EKFPhys-F} : g(\mathbf{s}_t) &= (\mathbf{x}_{t+\Delta t}, \boldsymbol{\xi}_{t+\Delta t})
\end{align*}
with the definition of the velocity in the next time step as defined above for each model.

The predicted state mean can be calculated from equations \eqref{eq: LCP formulation}
and \eqref{eq: pose update}
\begin{equation}
    \Bar{\boldsymbol{\mu}}_{t+\Delta t} = g(\boldsymbol{\mu}_t)
    \label{eq: predicted state mean}
\end{equation}

The predicted state covariance can be calculated as
\begin{equation}
    \Bar{\boldsymbol{\Sigma}}_{t+\Delta t} = \mathbf{G}_t \mathbf{\Sigma}_t \mathbf{G}^\top_t + \mathbf{S}_t
    \label{eq: predicted state cov}
\end{equation}
\noindent
where $G_t$ is the Jacobian of the state vector, $\mathbf{s}_t$, at time, $t$, and can be calculated as,
\begin{align}
    \mathbf{G}_t &= \frac{dg(\mathbf{s}_t)}{d\mathbf{s}_t}\\
    &= \lim_{\boldsymbol{\tau} \rightarrow 0} \frac{g\left(\mathbf{s}_t \boxplus \boldsymbol{\tau}\right) \boxminus g(\mathbf{s}_t)}{\boldsymbol{\tau}}
\end{align}

For Euclidean components (in $\mathbb{R}^n$) like $\mathbf{p}_t, \mathbf{v}_t$ and $\boldsymbol{\omega}_t$, the Jacobian with respect to these components of the state can be calculated as
\begin{align}
    \mathbf{G}^{\textrm{Euc}}_t &=  \lim_{\boldsymbol{\tau} \rightarrow 0} \frac{g\left(\mathbf{s}^{\textrm{Euc}}_t + \boldsymbol{\tau}\right)- g(\mathbf{s}^{\textrm{Euc}}_t)}{\boldsymbol{\tau}}\\
    &= \left. \frac{dg(\mathbf{s}^{\textrm{Euc}}_t + \boldsymbol{\tau})}{d\boldsymbol{\tau}} \right|_{\boldsymbol{\tau} = 0}
    \label{eq: Lin jacobian}
\end{align}
where $\mathbf{s}^{\textrm{Euc}}_t$ is the linear component of the state. For rotational components (in $SO(3)$) like $\mathbf{R}_t$, the Jacobian with respect to the rotational components of the state can be calculated as
\begin{align}
    \mathbf{G}^{Rot}_t &= \lim_{\boldsymbol{\tau} \rightarrow 0} \frac{ g\left(e^{[\boldsymbol{\tau}]_\times}\mathbf{R}_t\right)\cdot g(\mathbf{R}_t)^{-1}}{\boldsymbol{\tau}}\\
    &= \left. \frac{d \log\left[g\left(e^{[\boldsymbol{\tau}]_\times}\mathbf{R}_t\right)\cdot g(\mathbf{R}_t)^{-1}\right]}{d\boldsymbol{\tau}} \right|_{\boldsymbol{\tau} = 0}
    \label{eq: Rot jacobian}
\end{align}
where $[\boldsymbol{\tau}]_\times \mathbf{y} = \boldsymbol{\tau} \times \mathbf{y}$ for $\mathbf{y} \in \mathbb{R}^3$.

The Jacobian equations are written in such a way that they can be computed using PyTorch's autograd module. From equation \eqref{eq: Rot jacobian}, we need to compute the state transition function two times; once with the state ($\mathbf{s}_t$) and once with incrementing the state with an infinitesimally small $\boldsymbol{\tau}$. For EKFPhys and EKFPhys-F models, this step is computationally expensive since it has to solve the LCP problem involving contact and friction constraints twice.
To avoid that, a vector of zeros $\boldsymbol{\tau}$, which has the same dimensions of the log map of the state vector $\mathbf{s}_t$ is created with the gradient computation flag turned on. The increment ($\boldsymbol{\tau}$) is added to the state and the incremented state ($\mathbf{\hat{s}}_t=e^{[\boldsymbol{\tau}]_\times}\mathbf{s}_t$) is passed into the motion model to get ($g(\mathbf{\hat{s}}_t)$). Since the increment is a vector of zeros, the result of the operation will not change even without the increment. Therefore to get $g(\mathbf{s}_t)$, we simply detach $g(\mathbf{\hat{s}}_t)$ from the computational graph, compute the difference in rotation and calculate the log map.

Since the autograd module only computes the gradients with respect to a scalar quantity, we use Functorch's vmap module to parallelize the calculation of gradients of all the output entries with respect to $\boldsymbol{\tau}$.

\subsection{Correction step}
Since we observe poses ($\mathbf{z}_{t+\Delta t}$) with CIR \cite{lipson2022_coupled}, the observation function $h(\mathbf{s}_{t+\Delta t})$ only returns the pose component ($\mathbf{x}_{t+\Delta t}$) and the measurement state covariance $\mathbf{H}_t$ for EKFPhys model is given by
\setcounter{MaxMatrixCols}{20}
\begin{equation}
    \mathbf{H}_t = 
    \begin{bmatrix}
        1 & 0 & 0 & 0 & 0 & 0 & 0 & 0 & 0 & 0 & 0 & 0 & 0\\
        0 & 1 & 0 & 0 & 0 & 0 & 0 & 0 & 0 & 0 & 0 & 0 & 0\\
        0 & 0 & 1 & 0 & 0 & 0 & 0 & 0 & 0 & 0 & 0 & 0 & 0\\
        0 & 0 & 0 & 1 & 0 & 0 & 0 & 0 & 0 & 0 & 0 & 0 & 0\\
        0 & 0 & 0 & 0 & 1 & 0 & 0 & 0 & 0 & 0 & 0 & 0 & 0\\
        0 & 0 & 0 & 0 & 0 & 1 & 0 & 0 & 0 & 0 & 0 & 0 & 0\\
    \end{bmatrix}
\end{equation}
where the rotational components are again computed on the Lie algebra of $SO(3)$.

For the EKFPhys-F model, $\mathbf{H}_t$ remains the same but without the last column, as friction is not part of the state for this model.
We implement outlier measurement rejection by gating with threshold $\zeta$ using
\begin{equation}
    (\mathbf{z}_{t+\Delta t} \ominus h(\boldsymbol{\bar{\mu}}_{t+\Delta t}))^\top (\mathbf{H}_t\Bar{\boldsymbol{\Sigma}}_{t+\Delta t}\mathbf{H}^\top_t + \mathbf{Q}_t)^{-1} (\mathbf{z}_{t+\Delta t} \ominus h(\boldsymbol{\bar{\mu}}_{t+\Delta t})) > \zeta
    \label{eq: gating}
\end{equation}
where $(\mathbf{z}_{t+\Delta t} \ominus h(\boldsymbol{\bar{\mu}}_{t+\Delta t}))$ is a vector composed of the difference between the positions and the log map of the difference rotation. The Kalman gain $\mathbf{K}_t$ is calculated as 
\begin{equation}
    \mathbf{K}_t = \Bar{\boldsymbol{\Sigma}}_{t+\Delta t}\mathbf{H}^\top_t(\mathbf{H}_t\Bar{\boldsymbol{\Sigma}}_{t+\Delta t}\mathbf{H}^\top_t + \mathbf{Q}_t)^{-1}
    \label{eq: kalman gain}
\end{equation}

The state mean and covariance are updated as
\begin{align}
    \boldsymbol{\mu}_{t+\Delta t} &= \bar{\boldsymbol{\mu}}_{t+\Delta t} \boxplus \mathbf{K}_t(\mathbf{z}_{t+\Delta t} \ominus h(\boldsymbol{\bar{\mu}}_{t+\Delta t})\\
    \boldsymbol{\Sigma}_{t+\Delta t} &= (\mathbf{I} - \mathbf{K}_t\mathbf{H}_t)\bar{\boldsymbol{\Sigma}}_{t+\Delta t}
\end{align}

\section{Dataset Details}
\subsection{Synthetic Scene Setup}
\paragraph{Room size and center offset.}
The room in the synthetic scenes is rectangular with a fixed height of $2\,m$ and uniformly sampled extents in $x$ and $y$ in the range $[3.0,7.0]\,m$.
The center of the room (reference point for camera and initial object position) is varied by a uniformly sampled shift of $20\%$ of the previously sampled room size.

In the detector training dataset for the real-world experiments, we move the floor down by $1m$ and model the table as a cuboid dimensions $(0.815, 1.200, 1.00)^\top m$ at position $(0, 0, -0.5)^\top m$.
This setup mimics the size of the table in our lab with the table surface aligned with the horizontal plane through the origin.

\paragraph{Camera pose.}
The camera position is sampled uniformly on a half sphere around the aforementioned center of the room.
The half sphere has a radius of $0.85m$ for the detector training and $0.75m$ in the sliding objects dataset, both with a uniform random jitter of $\pm 0.05m$ on the camera distance.
The lower distance to the center in the sliding experiments accounts for objects moving further away than observed during detector training in some of the scenes.
We exclude the top and bottom $20\%$ of the half sphere (i.e.~re-sample the camera position if it's $z$-coordinate outside the range $[0.2,0.8]$ times camera distance) to avoid degenerate camera poses (i.e.~the camera looking at the scene directly from above or the camera being so low it looks below the floor).
The camera is set to focus on a point in the center, placed at $[0, 0, 0.1] \pm 0.05m$ relative to the aforementioned room center.

\paragraph{Object poses.}
For the detector training dataset, object positions are sampled uniformly in the range $[-0.5,0.5]$ in $x$ and $y$ and $[0.0,0.5]$ in $z$.
For rotations, we uniformly sample 3 Euler angles in the range $[-\pi,\pi]$.
In the part of the dataset, where objects are placed in the air, these are the final object poses, while in the part where objects are placed on the table, we simulated them from these initial poses until they are static (defined as linear and angular velocity below $10^{-3}$) and then render the scene with a camera trajectory in a circle around the center.

In the detector training dataset for the real-world experiments, the objects are placed in the air above the table (i.e. range $[-0.4, 0.4]m$ in $x$ and $[-0.6, 0.6]m$ in $y$).
To further improve robustness, we place additional objects from ShapeNet \cite{shapenet2015} on the floor (lower than the table), upright with random rotation around $z$.

For the sliding objects dataset, the initial object position is set in polar coordinates with a distance from the scene center uniformly sampled in the range $[0.3,0.4]m$ and an angle in the range $[-\pi,\pi]$.
In this dataset, all objects are placed upright, i.e.~the vertical axis in the object coordinate frame as stored in the models in the BOP dataset \cite{hodan2018_bop,hodan2020_bop} is aligned with the world $z$-axis and we only randomly sample orientations around $z$ in the range $[-\pi,\pi]$.
For the 2-objects scenes, we additionally sample a second object and place it at the center with random orientation around $z$.
The target point $\mathbf{c}_v$ is sampled on a circle with radius $0.02m$ for these sequences.
This way, it will be hit by the sliding objects with the mentioned sampling settings in most cases.

\subsection{Velocities and Physical parameters in the sliding objects dataset}
\begin{figure}
    \centering
    \includegraphics[width=\linewidth]{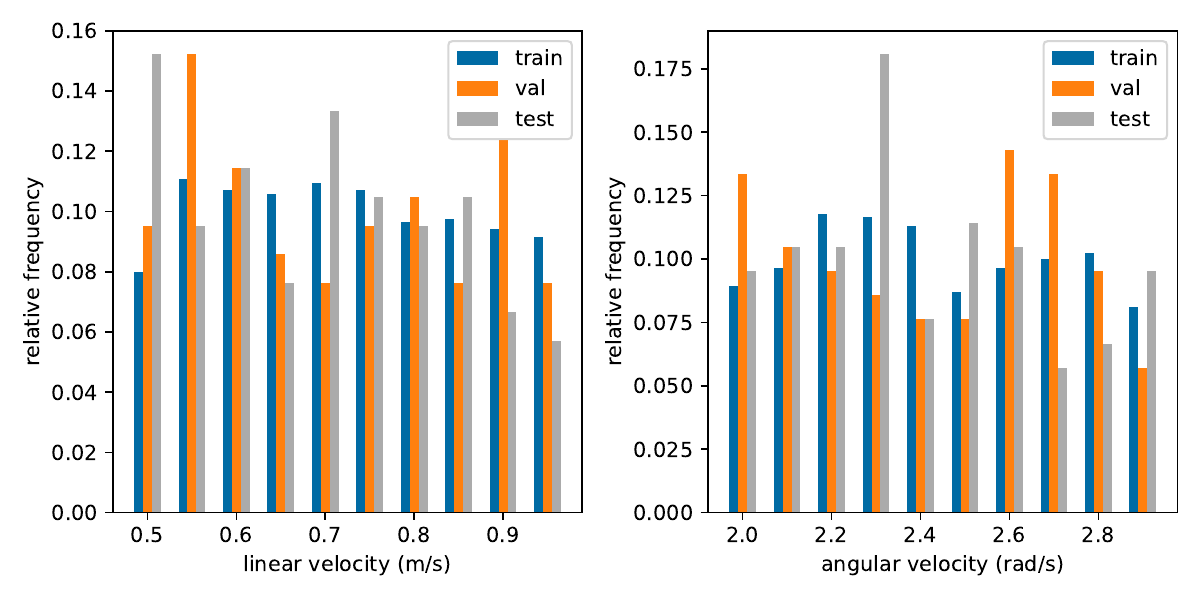}\\
    \includegraphics[width=\linewidth]{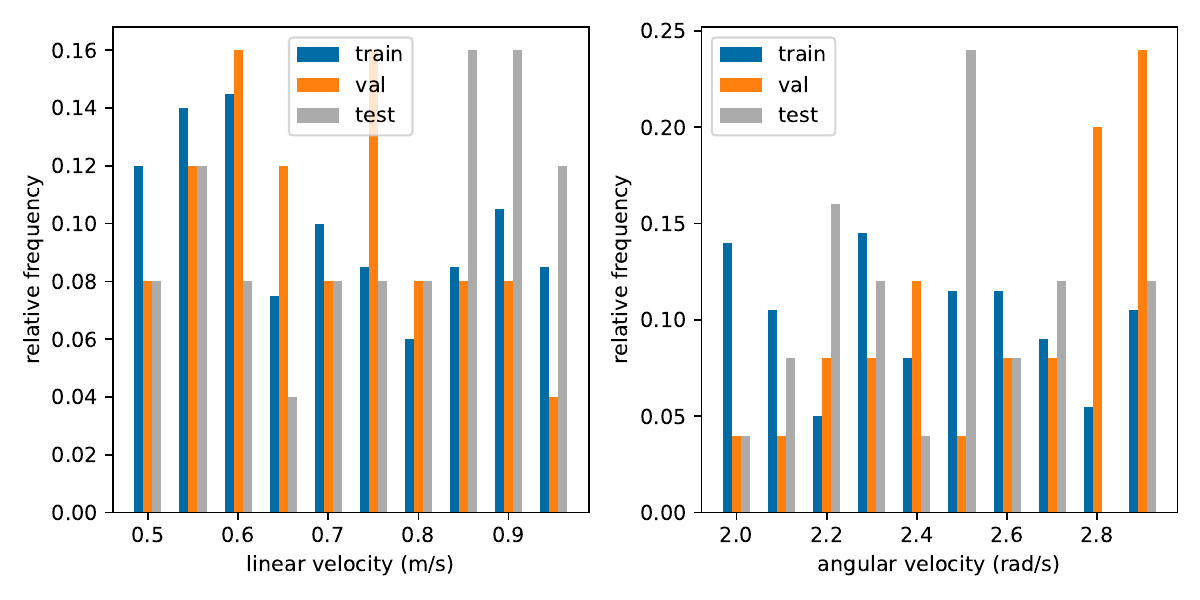}
    \caption{
    Initial velocity statistics on the different splits for sliding sequences. Top: single objects sequences, bottom: 2-object sequences (velocity of moving object).
    We plot histograms for each split with 10 bins in the sampling range. Equal distribution would thus be achieved at a relative frequency of $0.1$ for all plots.
    }
    \label{fig:vel_statistic_syn}
\end{figure}

In our sliding objects dataset, the initial object velocity magnitudes are sampled uniformly in the range $[2.0, 3.0]rad/s$ for angular velocity $\omega_0$ and in the range $[0.5, 1.0]m/s$ for linear velocity $v_0$.
The direction of movement is determined by sampling a point $\mathbf{c}_v$ on a circle with a radius of $0.1m$ around the scene center and setting the linear velocity to $\mathbf{v}_0 = v_0\frac{\mathbf{c}_v - \mathbf{p}_0}{\|\mathbf{c}_v - \mathbf{p}_0\|}$.
For angular velocity we first sample $\omega_z$ from the range $[-\pi,\pi]$, and then set $\boldsymbol{\omega}_0 = \omega_0 \frac{[0, 0, \omega_z]^\top}{\|[0, 0, \omega_z]^\top\|}$.
This procedure allows for future extension to arbitrary rotations.
\cref{fig:vel_statistic_syn} shows statistics on the sampled linear and angular velocities in the synthetic sliding datasets.

Masses for each object are set as reported in \cite{calli2015_benchmarking} (see \cref{tab:obj_id_mapping}).
As we do not know the ground truth friction coefficients for the objects, we set it to $0.2$ for the background and sample friction values uniformly in the range $[0.1, 0.5]$ for the objects.
We report the sampled friction values for the objects in \cref{tab:obj_id_mapping}.
Note that only the combined friction value of background and object can be observed.
This is the product of the values reported in \cref{tab:obj_id_mapping} and the background friction coefficient $0.2$.
The mean (center of the sampling range) for the object friction coefficients is $0.3$, thus the mean combined friction coefficient as used in the experiments for initialization is $0.06$.

\subsection{Detector training dataset splits}
For detector training, we create two training splits of 80 and two validation splits of 15 sequences each, where each sequence contains 200 frames.
The difference between the two splits (for each, training and validation) is the object placement.
In the first set of splits, ``train'' and ``val'', objects are placed randomly in the air (with random position and orientation) and the camera moves around them in a circle.
For the second set of splits, ``train stable'' and ``val stable'', we first simulate dropping the objects until they are stable (i.e.~angular and linear velocity below $10^{-3}$), and then render the resulting arrangement with the camera moving around the objects.
These ``stable'' splits might exhibit less variation in object orientation as not all orientations are stable, but might capture visual effects such as casting shadows on the floor better.

For detector performance on real-world sequences, we found it beneficial to train the detector on different synthetic splits replicating the table found in the real-world sequences and placing ShapeNet \cite{shapenet2015} objects in the background for additional clutter with the same number of sequences and frames in the splits.
 
\subsection{Object sampling statistics}
\begin{figure}
    \centering
    \includegraphics[width=.49\linewidth]{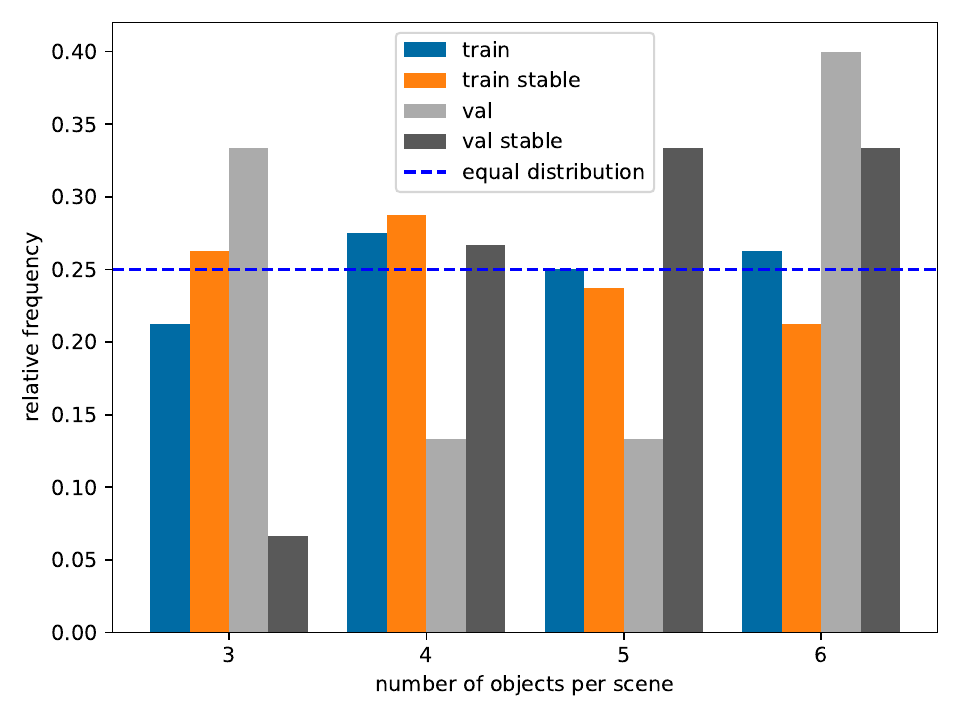}
    \includegraphics[width=.49\linewidth]{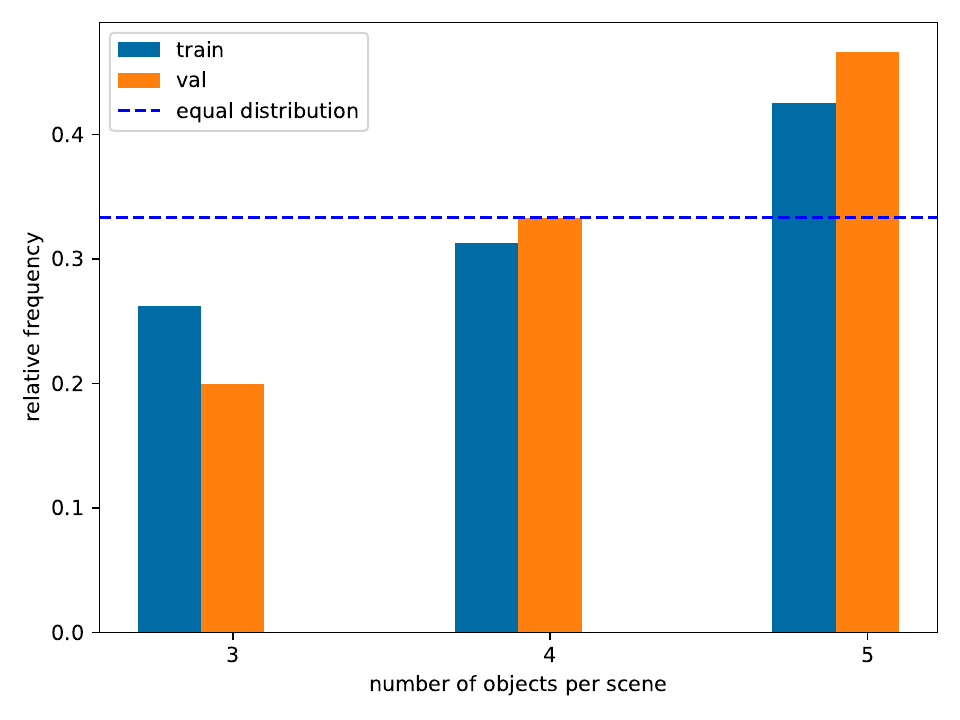}
    \caption{Distribution of number of objects in the detector training dataset. Left: detector training for synthetic experiments. Equal distribution of the 4 possible values would result in relative frequencies of $0.25$.
    Right: detector training for real-world experiment. Equal distribution of the 3 possible values would result in relative frequencies of $0.33$.
    The training splits in all cases are very close to the equal distribution, while the validation splits exhibit a bit more deviation due to the overall smaller number of scenes.}
    \label{fig:obj_count_detector}
\end{figure}

\begin{figure}
    \centering
    \includegraphics[width=\linewidth]{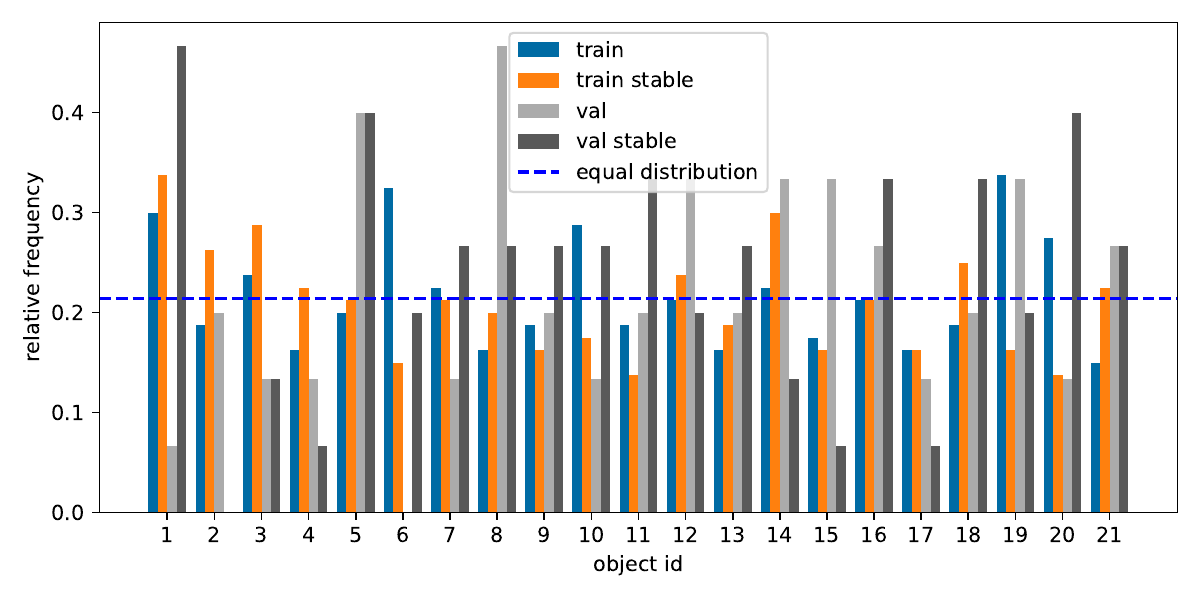}
    \caption{Frequency distribution of each object in the detector training dataset. At an average of $4.5$ objects per scene, equal distribution would result in a relative frequency of $0.21$. Similar as in \cref{fig:obj_count_detector}, the training splits are close to this, while the validation splits show more variation.}
    \label{fig:obj_frequency_detector}
\end{figure}

\begin{figure}
    \centering
    \includegraphics[width=\linewidth]{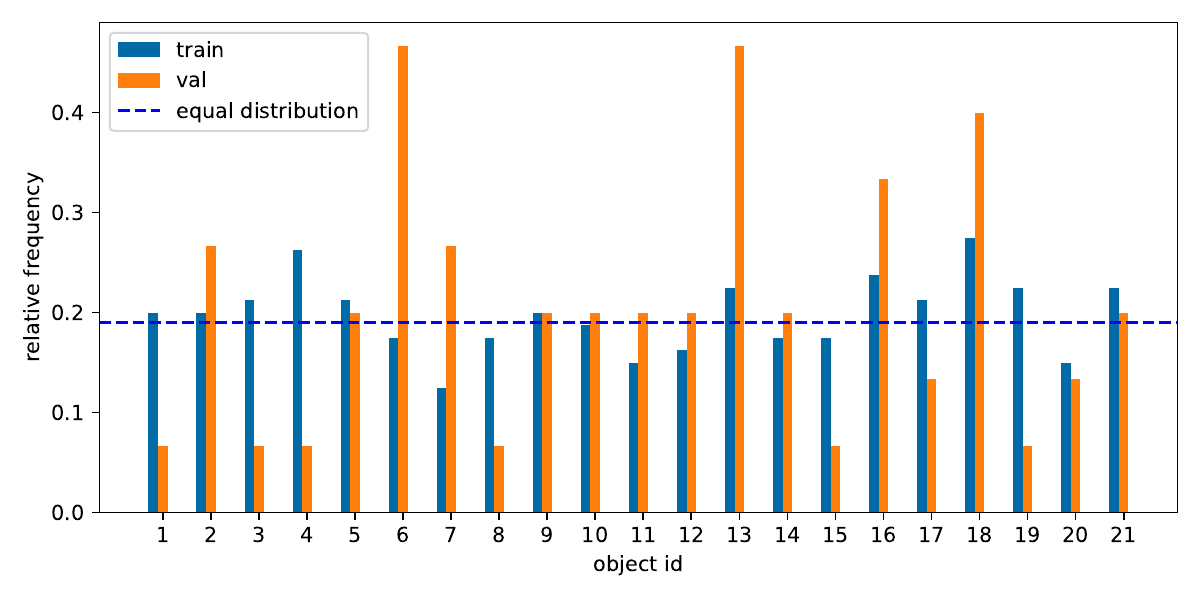}
    \caption{Frequency distribution of each object in the detector training dataset for real-world sequences. At an average of $4$ objects per scene, equal distribution would result in a relative frequency of $0.19$. Similar as in \cref{fig:obj_count_detector}, the training splits are close to this, while the validation splits show more variation.}
    \label{fig:obj_frequency_detector_real}
\end{figure}

For the detector training dataset, we sample between 3 and 6 objects of the 21 objects present in the YCB Video dataset \cite{xiang2018_posecnn}.
The detector training dataset for the real-world experiments contains 3 to 5 objects per scene.
Please see \cref{fig:obj_count_detector} for statistics on the numbers of objects in the scenes and \cref{fig:obj_frequency_detector,fig:obj_frequency_detector_real} for the relative frequencies of the objects.
\cref{tab:obj_id_mapping} shows the mapping between BOP object ids and the names of the YCB objects.
The figures show that the uniform sampling set up in the rendering script is followed quite closely for the training dataset, while the validation dataset statistics seem less uniform due to the small sample size (80 training and 15 validation scenes each).

\begin{table}
\footnotesize
    \centering
    \begin{tabular}{cccc}\toprule
         Id & Name & Mass (kg) & Friction \\\midrule
         1 & Master Chef Can & $0.414$ & $0.292$ \\
         2 & Cracker Box & $0.411$ & $0.178$ \\
         3 & Sugar Box & $0.514$ & $0.335$ \\
         4 & Tomato Soup Can & $0.349$ & $0.246$ \\
         5 & Mustard Bottle & $0.603$ & $0.273$ \\
         6 & Tuna Fish Can & $0.171$ & $0.205$ \\
         7 & Pudding Box & $0.187$ & $0.214$ \\
         8 & Gelatin Box & $0.097$ & $0.377$ \\
         9 & Potted Meat Can & $0.370$ & $0.123$ \\
         10 & Banana & $0.066$ & $0.355$ \\
         11 & Pitcher Base & $0.244$ & $0.185$ \\\bottomrule
     \end{tabular}
    \begin{tabular}{cccc}\toprule
         Id & Name & Mass (kg) & Friction \\\midrule
         12 & Bleach Cleanser & $1.131$ & $0.394$ \\
         13 & Bowl & $0.147$ & $0.317$ \\
         14 & Mug & $0.118$ & $0.327$ \\
         15 & Power Drill & $0.895$ & $0.439$ \\
         16 & Wood Block & $0.729$ & $0.272$ \\
         17 & Scissors & $0.082$ & $0.433$ \\
         18 & Large Marker & $0.0158$ & $0.439$ \\
         19 & Large Clamp & $0.125$ & $0.467$ \\
         20 & Extra Large Clamp & $0.202$ & $0.460$ \\
         21 & Foam Brick & $0.028$ & $0.182$ \\
         & & & \\\bottomrule
    \end{tabular}
    \caption{Mapping between object ids, names, masses and friction values.}
    \label{tab:obj_id_mapping}
\end{table}

\begin{figure}
    \centering
    \includegraphics[width=\linewidth]{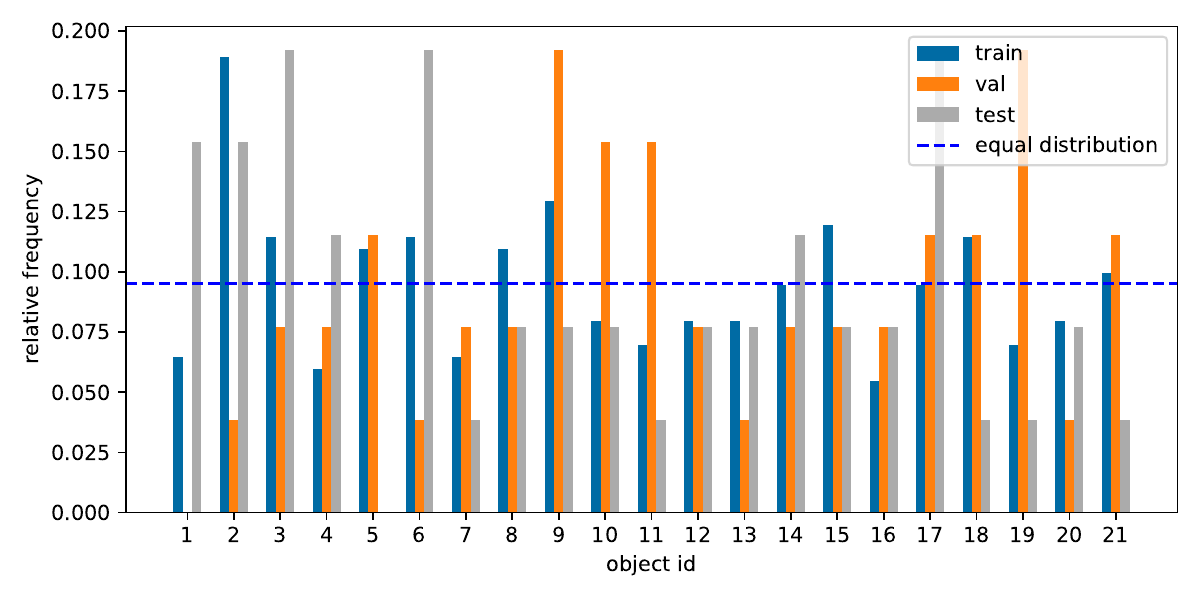}
    \caption{Frequency distribution of each object in the 2-object sliding splits. At $2$ objects per scene, equal distribution would result in a relative frequency of $0.095$. The training split is close to this, while the validation and test splits show more variation.}
    \label{fig:obj_frequency_2_objects}
\end{figure}

As mentioned in the main paper, the dataset of single-object sliding sequences contains 40 sequences per object for training and 5 sequences per object for validation (i.e.~hyperparameter tuning in our case) and testing (i.e.~in total 840 sequences for training and 105 sequences for validation and test).
Statistics on relative frequencies for the 2-objects sequences with 200 training and 25 validation and test sequences are plotted in \cref{fig:obj_frequency_2_objects}.

\subsection{Real-world object properties}
We report object properties for the real-world objects in \cref{tab:obj_properties_real}.
We fill the objects ``Mustard Bottle'', ``Pitcher'' and ``Bleach Cleanser'' partially with sand to stabilize the sliding and measure the mass of all objects using a scale.

\begin{figure}
    \centering
    \includegraphics[width=\linewidth]{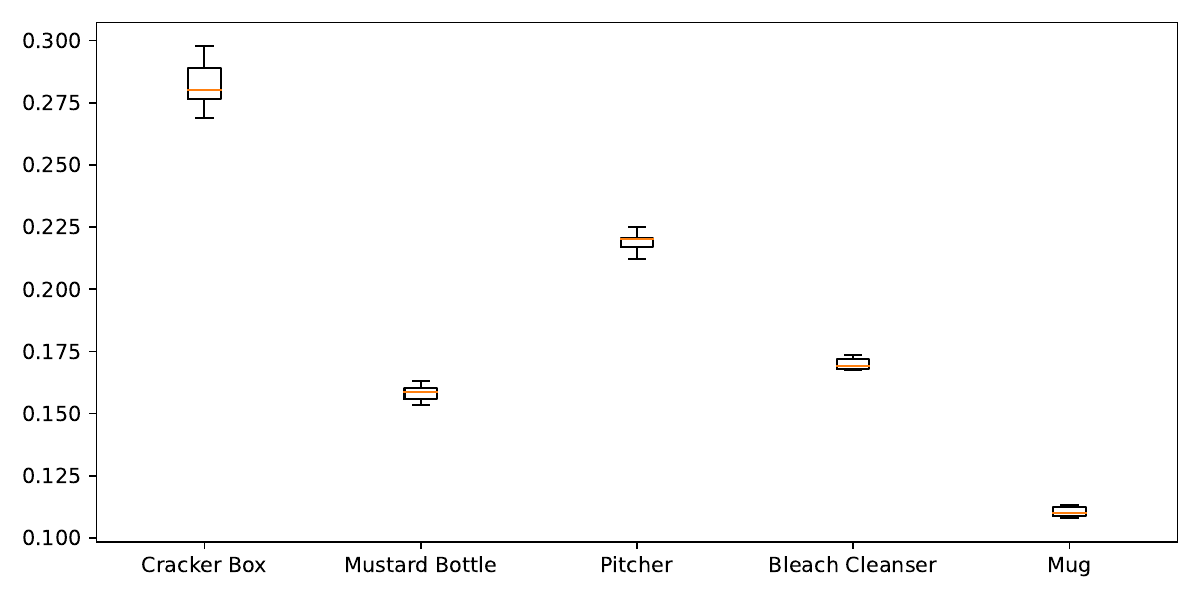}
    \caption{Statistics on the real-world friction value estimation. We get consistent estimates across the 10 runs per object. For Mustard Bottle and Bleach Cleanser we estimate similar friction parameters, as they are made from similar materials.}
    \label{fig:friction_statistics_real}
\end{figure}

To estimate the combined friction values between the objects and the table, we equip the table with MoCap-markers and tilt it at an angle, which we can measure using the MoCap pose then.
We then put each object on the table and let it slide 10 times, recording its trajectory with the MoCap-system.
Afterwards, we compute linear velocities from the trajectory with finite differences and fit a line to the resulting noisy velocity graph.
The slope $a$ of this line gives us the acceleration of the object, which allows us to compute the combined friction coefficient by the formula $\mu = \tan(\alpha) - \frac{a}{g}$, where $\alpha$ is the angle at which the table is tilted and $g = 9.81\frac{m}{s^2}$ is the gravity constant.
In \cref{tab:obj_properties_real}, we report the median values over the 10 recordings.
Additionally, we provide box plots with the statistics of friction values for each object in \cref{fig:friction_statistics_real}.
We observe that the friction values are very consistent across the 10 runs, i.e.~the estimation range for each object is small.
Furthermore, we get similar friction values for ``Mustard Bottle'' and ``Bleach Cleaser'' which makes sense as they are made from similar materials.
The range for ``Cracker Box'' shows more variation, as this object has low mass and anisotropic friction (i.e.~different friction in different directions as the cardboard surface is not perfectly flat and wears off over time).

\begin{table}
    \centering
    \begin{tabular}{cccc}\toprule
         Id & Name & Mass (kg) & Friction (combined) \\\midrule
         2 & Cracker Box & $0.0565$ & $0.280$ \\
         5 & Mustard Bottle & $0.3180$ & $0.159$ \\
         11 & Pitcher & $0.7120$ & $0.220$ \\
         12 & Bleach Cleanser & $0.4030$ & $0.169$ \\
         14 & Mug & $0.1030$ & $0.110$ \\\bottomrule
    \end{tabular}
    \caption{Object properties for the real-world objects.}
    \label{tab:obj_properties_real}
\end{table}

\subsection{Real-world velocity statistics}

\begin{figure}[tbh]
    \centering
    \includegraphics[width=\linewidth]{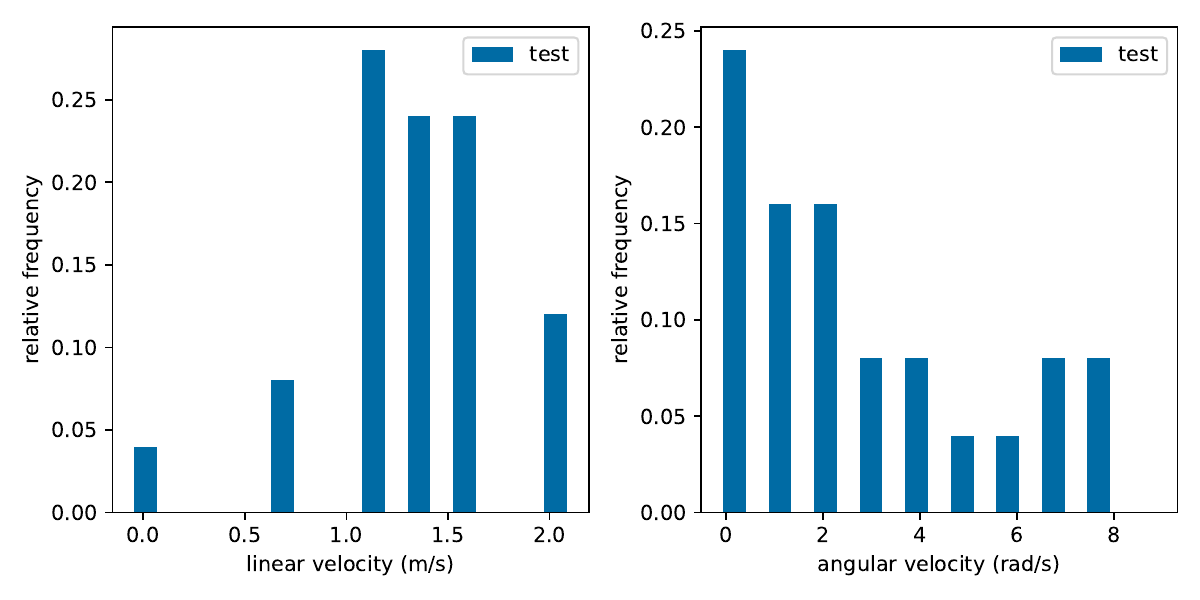}
    \caption{Initial velocity statistics on the real-world sliding sequences. Each plot shows a histogram with 10 bins between the minimum and maximum velocity norm.}
    \label{fig:vel_statistic_real}
\end{figure}

We estimate real-world ground-truth velocities by finite differences from the MoCap poses in the first two frames.
For linear velocities, we compute the magnitude by $\|\mathbf{v}\| = \left\|\frac{\mathbf{t}_1 - \mathbf{t}_0}{\Delta t}\right\|$, where $\mathbf{t}_0$ and $\mathbf{t}_1$ are the positions of the object in the first and second frame, respectively.
Similarly for angular velocities we estimate the magnitude $\|\boldsymbol{\omega}\| = \left\|\frac{\log(\mathbf{R}_1\mathbf{R}_0^\top)}{\Delta t}\right\|$, where $\mathbf{R}_0$ and $\mathbf{R}_1$ are the rotations in the first and second frames, respectively and $\log$ computes the matrix logarithm, i.e.~the axis-angle vector of the difference rotation in this case.
Histograms over the velocity magnitudes with 10 bins over the 20 real-world sequences are shown in \cref{fig:vel_statistic_real}.

\clearpage
\section{Additional Results and Ablations}

\subsection{Prediction and Recall Plots for Two Object Scenes}

\begin{figure}[tbh]
    \centering
    \includegraphics[width=.49\linewidth]{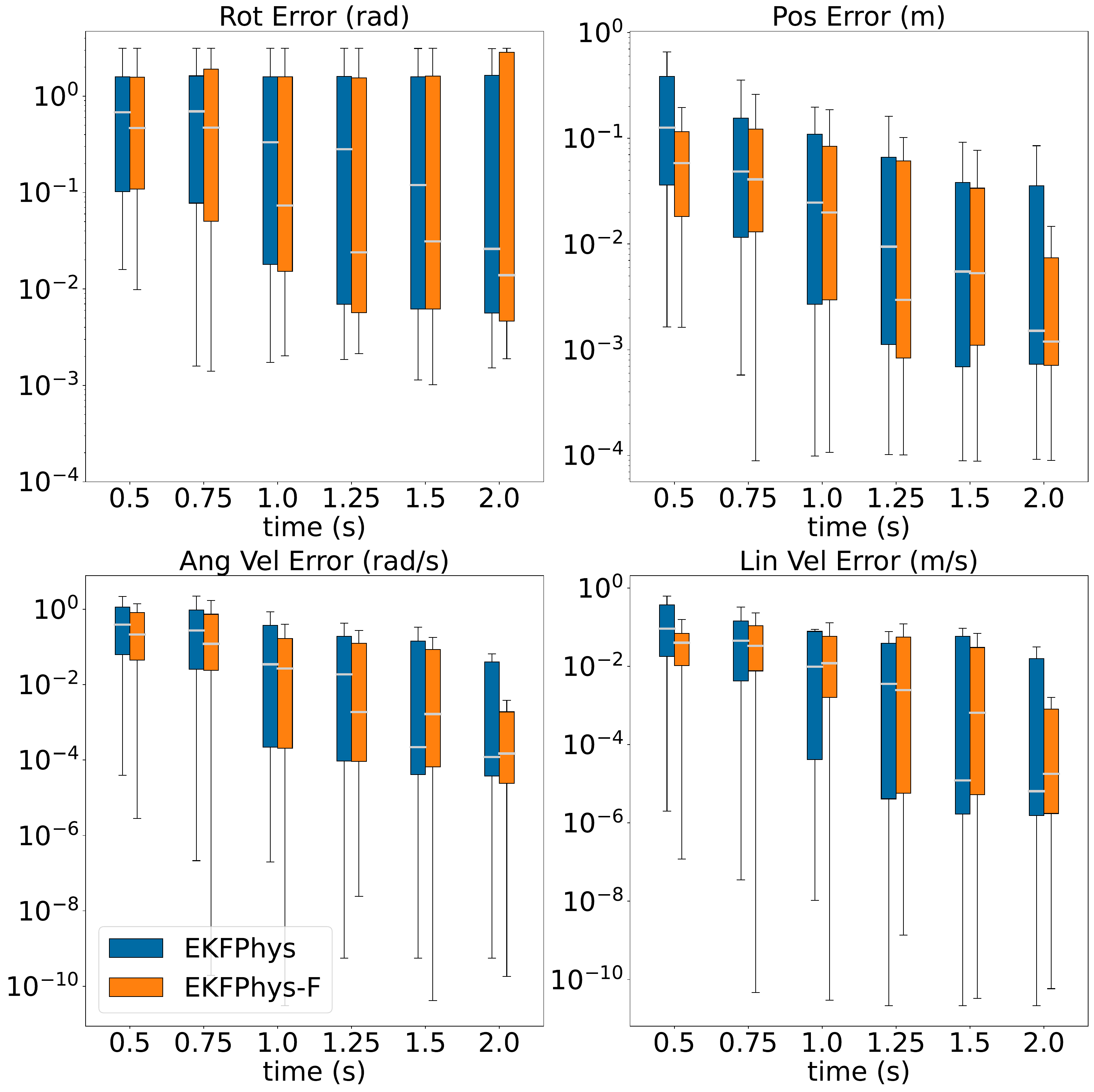}
    \caption{Prediction accuracy on single object sliding in synthetic sequences for symmetric objects. The box plot shows the median and quartiles of average rotation error (top-left), average position error (top-right), average angular velocity error (bottom-left) and average linear velocity error (bottom-right) of the trajectory over all the scenes in the dataset.}
    \label{fig:trajectory_error_prediction_two_objs_nonsymm_synphys_non_damping}
\end{figure}

\begin{figure}[tbh]
    \centering
    \includegraphics[width=.49\linewidth]{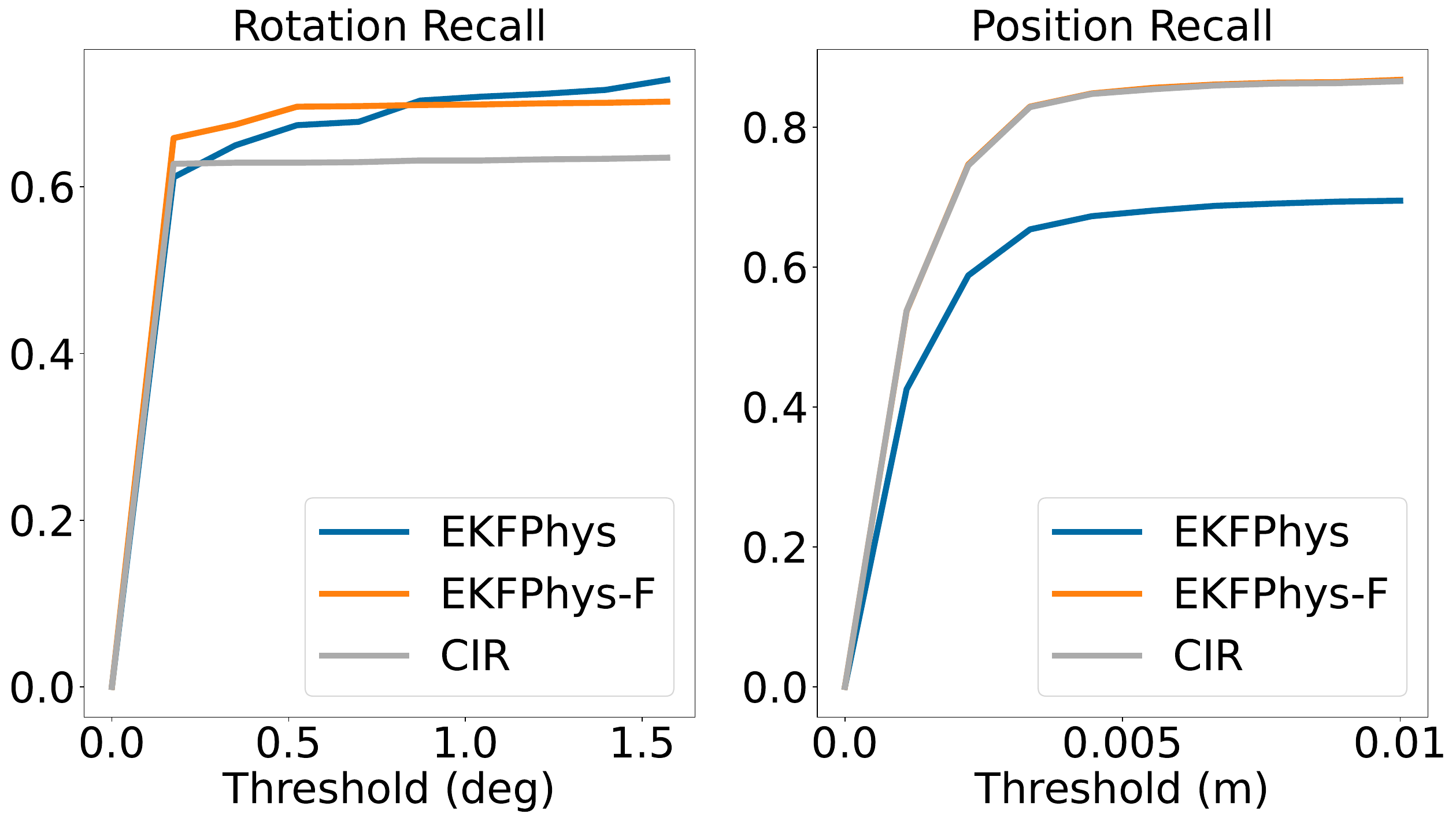}
    \caption{Recall of rotation and position for two objects sliding in synthetic sequences for non-symmetric objects.
    The x-axis denotes the threshold (in degrees for rotation and m for position) under which the detection/estimation is considered accurate.
    We achieve higher recall rates for rotation than CIR by filtering poses for frames without detections. For positions, EKFPhys-F achieves similar recall rate as CIR.}
    \label{fig:recall_two_objs_nonsymm_synphys_non_damping}
\end{figure}

In this section, we present the prediction plots (Fig. \ref{fig:trajectory_error_prediction_two_objs_nonsymm_synphys_non_damping}) and the recall curves (Fig. \ref{fig:recall_two_objs_nonsymm_synphys_non_damping}) for two object scenarios on synthetic dataset for non symmetric objects.

From Fig. \ref{fig:trajectory_error_prediction_two_objs_nonsymm_synphys_non_damping} we can see that the prediction accuracy of EKFPhys is sightly worse in estimating rotations and translations, comparable in estimating angular velocity and better at estimating linear velocities. The lower accuracy in rotations and translations is because the objects do not move long enough for EKFPhys to filter the friction accurately. This is also evident in the position recall curves in Fig. \ref{fig:recall_two_objs_nonsymm_synphys_non_damping}.

\subsection{Full Friction Analysis}

\begin{figure}[tbh]
    \centering
    \includegraphics[width=.49\linewidth]{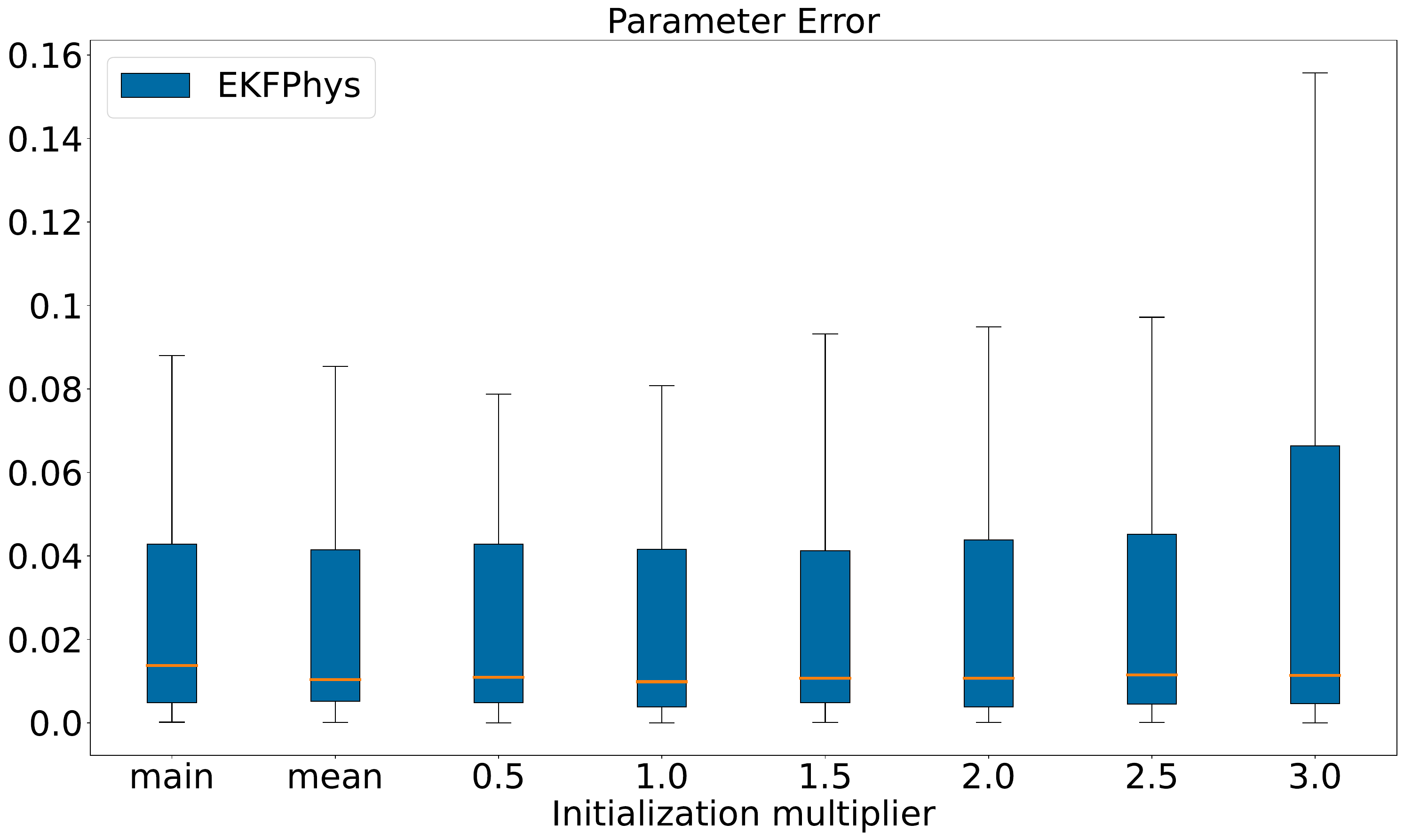}
    \includegraphics[width=.49\linewidth]{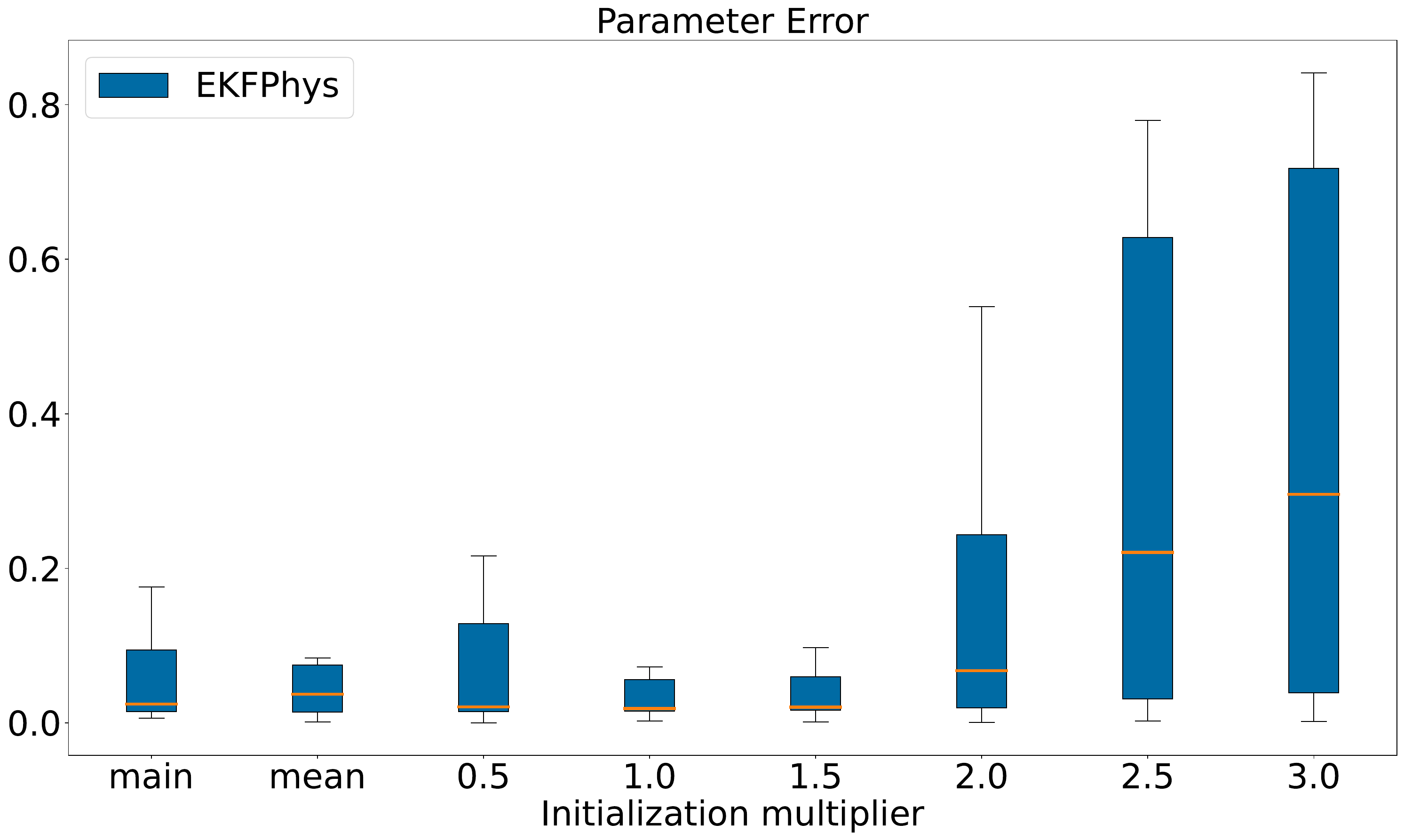}\\
    \caption{
    Friction coefficient error on single object sliding in synthetic sequences for non-symmetric objects. 
    The x-axis denotes the initialization multiplier, i.e., the friction value is initialized with factor * gt-friction. For "mean" we initialize with the average value of all the friction values in the validation dataset, which is 0.062 (synthetic) and 0.164 (real) in our case.
    The entry "main" indicates the value 0.0, at which our model EKFPhys is initialized for filtering and prediction experiments.
    Our approach recovers the friction coefficients with a good accuracy in the median.
    }
    \label{fig:friction_estimation_full_error_single_obj_nonsymm_syn_ycbphys_non_damping}
\end{figure}

In the main paper, we have shown the friction analysis for EKFPhys on synthetic and real-world scenes for reduced number of initialization multipliers. In this section, we show the full range of friction analysis in Fig. \ref{fig:friction_estimation_full_error_single_obj_nonsymm_syn_ycbphys_non_damping}

From Fig. \ref{fig:friction_estimation_full_error_single_obj_nonsymm_syn_ycbphys_non_damping}, we see that for synthetic scenes the median error increases very slowly as we move away from gt initialization, i.e., initialization multiplier 1.0. 
In the case of real-world scenes both median and quartile errors increase more strongly with deviation of the initialization from the gt value. 

In Fig. \ref{fig:friction_estimation_full_error_single_obj_nonsymm_syn_ycbphys_non_damping}, we show the absolute friction error ($||\theta^{gt} - \theta^{est}||$), but when we consider the signed error ($\theta^{gt} - \theta^{est}$), we observe that the median bias is 0.0067 for synthetic sequences and -0.0203 for real world sequences. 
We hypothesize that the larger bias on the real world sequences is due to model inaccuracies and noisy CIR poses.

\subsection{Evaluation on Cracker Box on Real-World Dataset}

\begin{figure}[tbh]
    \centering
    \includegraphics[width=.49\linewidth]{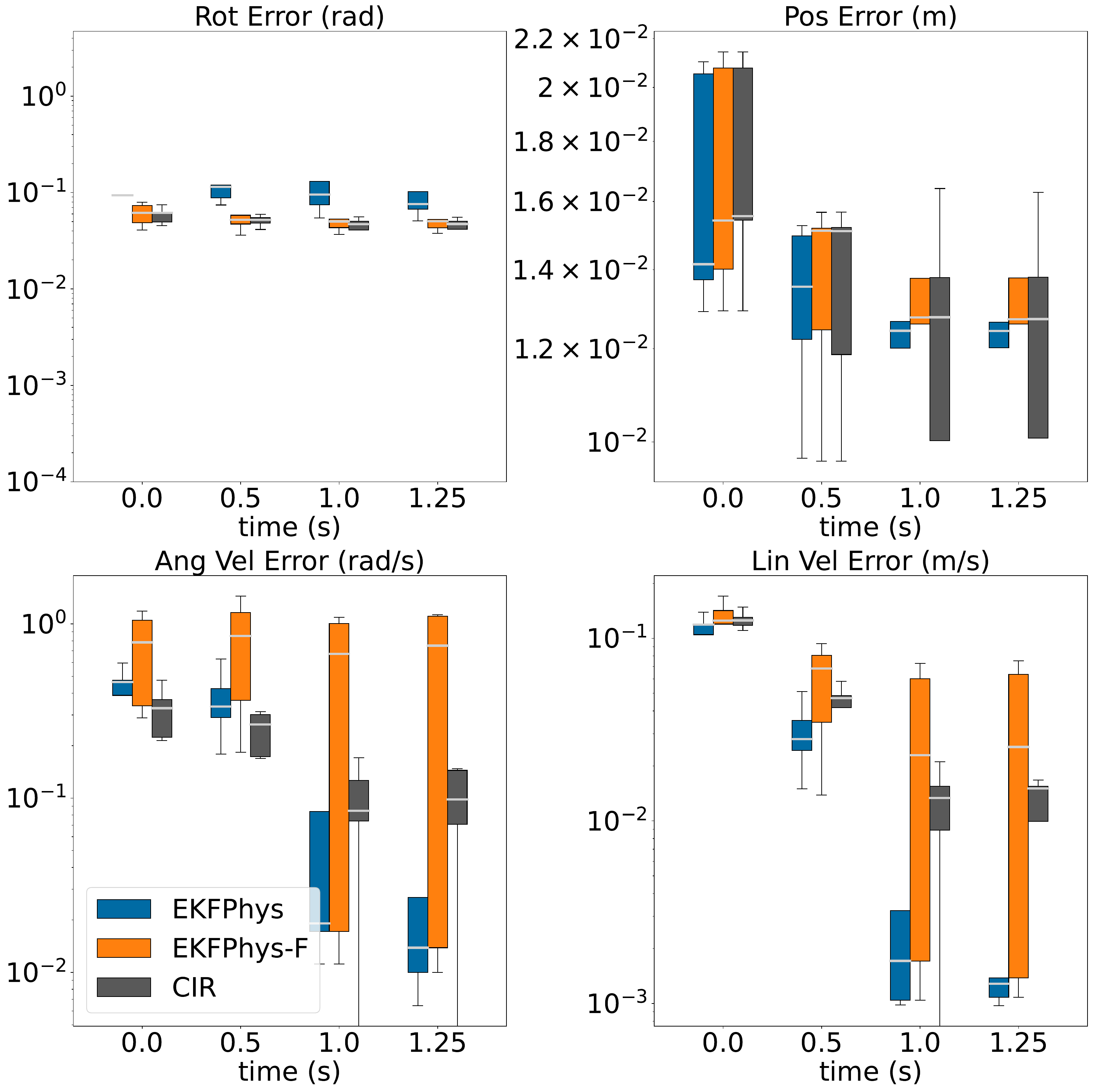}
    \includegraphics[width=.49\linewidth]{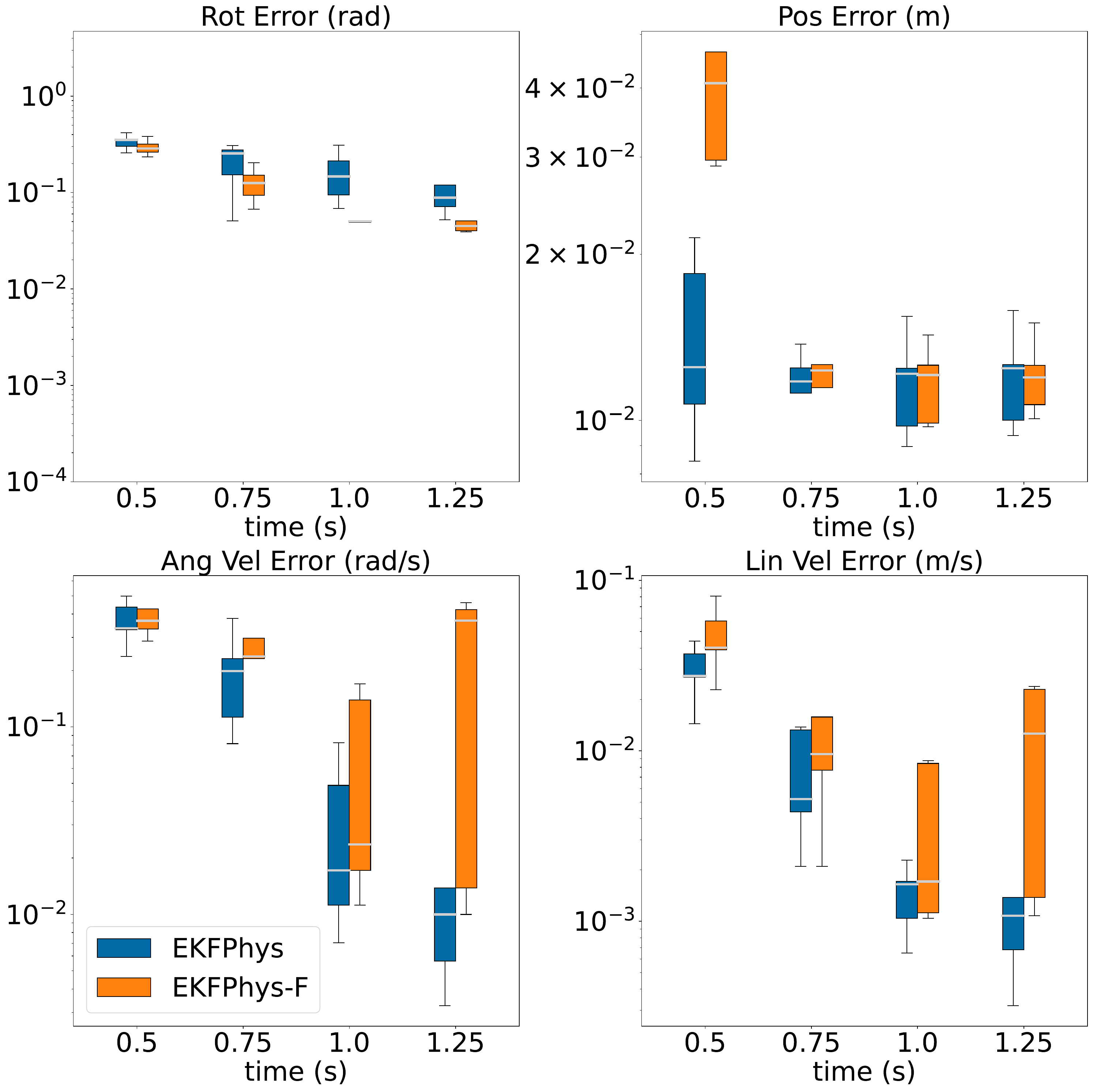}\\
    \caption{Filtering (left) and prediction (right) accuracy on cracker box sliding in real-world sequences. The box plot shows the median and quartiles of average rotation error (top-left), average position error (top-right), average angular velocity error (bottom-left) and average linear velocity error (bottom-right) of the trajectory over all the scenes in the dataset.
    From the plots we can see that EKFPhys performs better than CIR in estimating positions, linear and angular velocities. But the variant EKFPhys-F is much worse in comparison. This is because cracker box is made of cardboard, and the surface is not completely flat. This means that the object doesn't touch the table evenly and thus has unequal friction at different places (anisotropic friction), which has heavy non-linear frictional effects that are not captured by our model. Since EKFPhys-F gets the gt friction, we hypothesize that it has a strong inductive bias towards the wrong friction, whereas EKFPhys can adjust friction to adapt to model inconsistencies.}
    \label{fig:trajectory_error_filt_prediction_one_obj_nonsymm_ycbphys_cb}
\end{figure}

\begin{figure}[tbh]
    \centering
    \includegraphics[width=.49\linewidth]{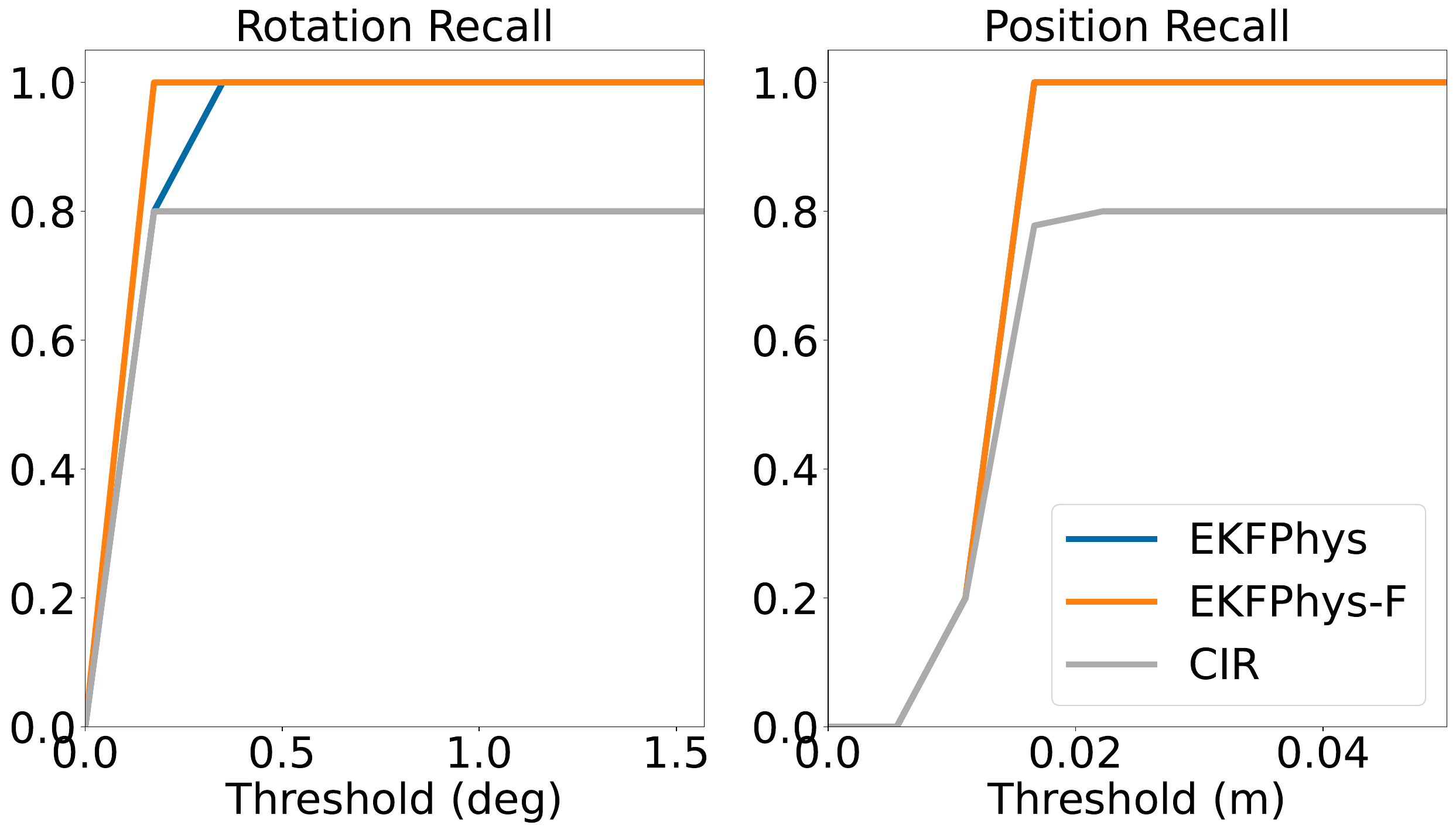}
    \caption{Recall of rotation and position for cracker box sliding in real-world sequences.
    The x-axis denotes the threshold (in degrees for rotation and m for position) under which the detection/estimation is considered accurate.
    We achieve higher recall rates than CIR by filtering poses for frames without detections. We can see that the recall rates are significantly higher than the other objects in real-world dataset (see Fig. 6 in the main paper). This is possibly because the cracker box is larger with distinguishable textures.}
    \label{fig:recall_one_obj_nonsymm_ycbphys_cb}
\end{figure}

\begin{figure}[tbh]
    \centering
    \includegraphics[width=.49\linewidth]{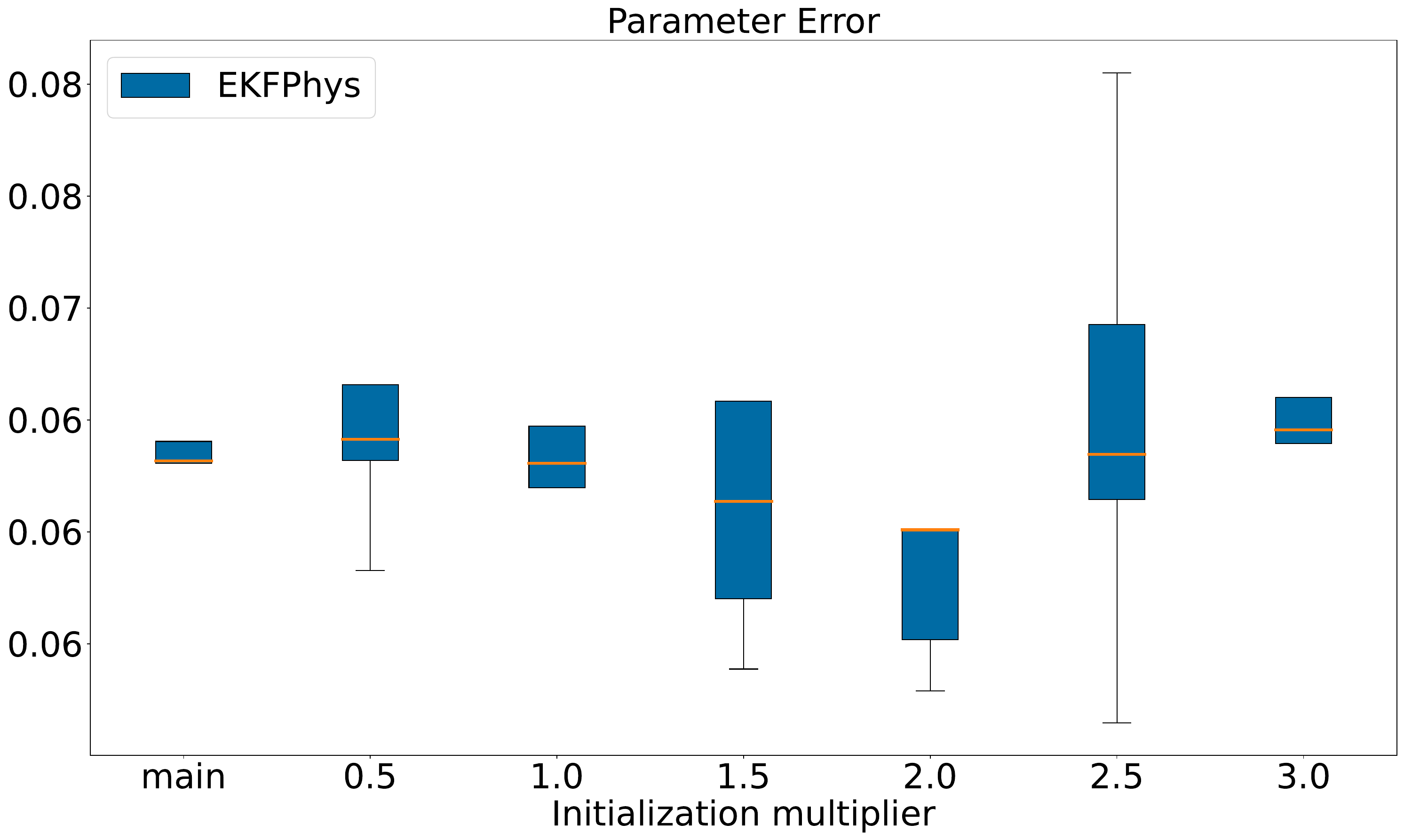}
    \caption{Friction coefficient error on cracker box sliding in real-world sequences. 
    The entry "main" indicates the value 0.0, at which our model EKFPhys is initialized for filtering and prediction experiments.
    The median friction error is around 0.06 (gt friction value is 0.275) which is high when compared to the median error on the other objects (0.0256; see Table 1 in the main paper) in the real-world dataset. This is because cracker box is made of cardboard, and the surface is not completely flat. This means that the object doesn't touch the table evenly and thus has unequal friction at different places (anisotropic friction), which has heavy non-linear frictional effects that are not captured by our model.
    }
    \label{fig:friction_estimation_full_error_single_obj_nonsymm_ycbphys_cb}
\end{figure}

In this section we present the filtering and prediction errors (Fig. \ref{fig:trajectory_error_filt_prediction_one_obj_nonsymm_ycbphys_cb}), recall curves (Fig. \ref{fig:recall_one_obj_nonsymm_ycbphys_cb}) and friction analysis plots (Fig. \ref{fig:friction_estimation_full_error_single_obj_nonsymm_ycbphys_cb}) for the cracker box on real-world dataset. We did not include it with the analysis in the main paper because the surface of the cracker box is made of cardboard, and the surface is not completely flat. This means that the object doesn't touch the table evenly and thus has unequal friction at different places (anisotropic friction), which has heavy non-linear frictional effects that are not captured by our model. It is still interesting to know how our model performs in the scenes with this object.

\subsection{Evaluation on Symmetric Objects on Synthetic Dataset}

\begin{figure}[tbh]
    \centering
    \includegraphics[width=.49\linewidth]{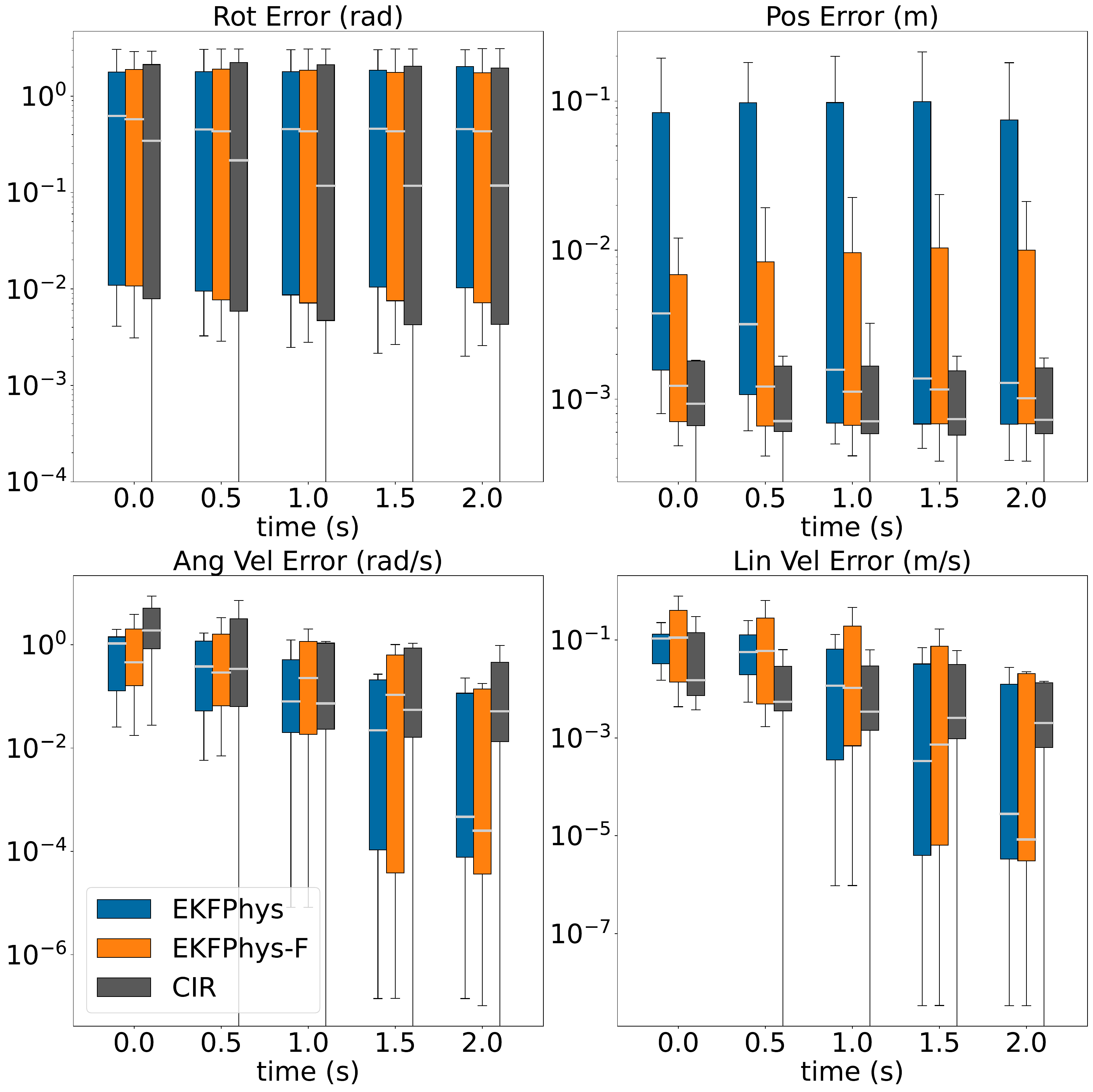}
    \includegraphics[width=.49\linewidth]{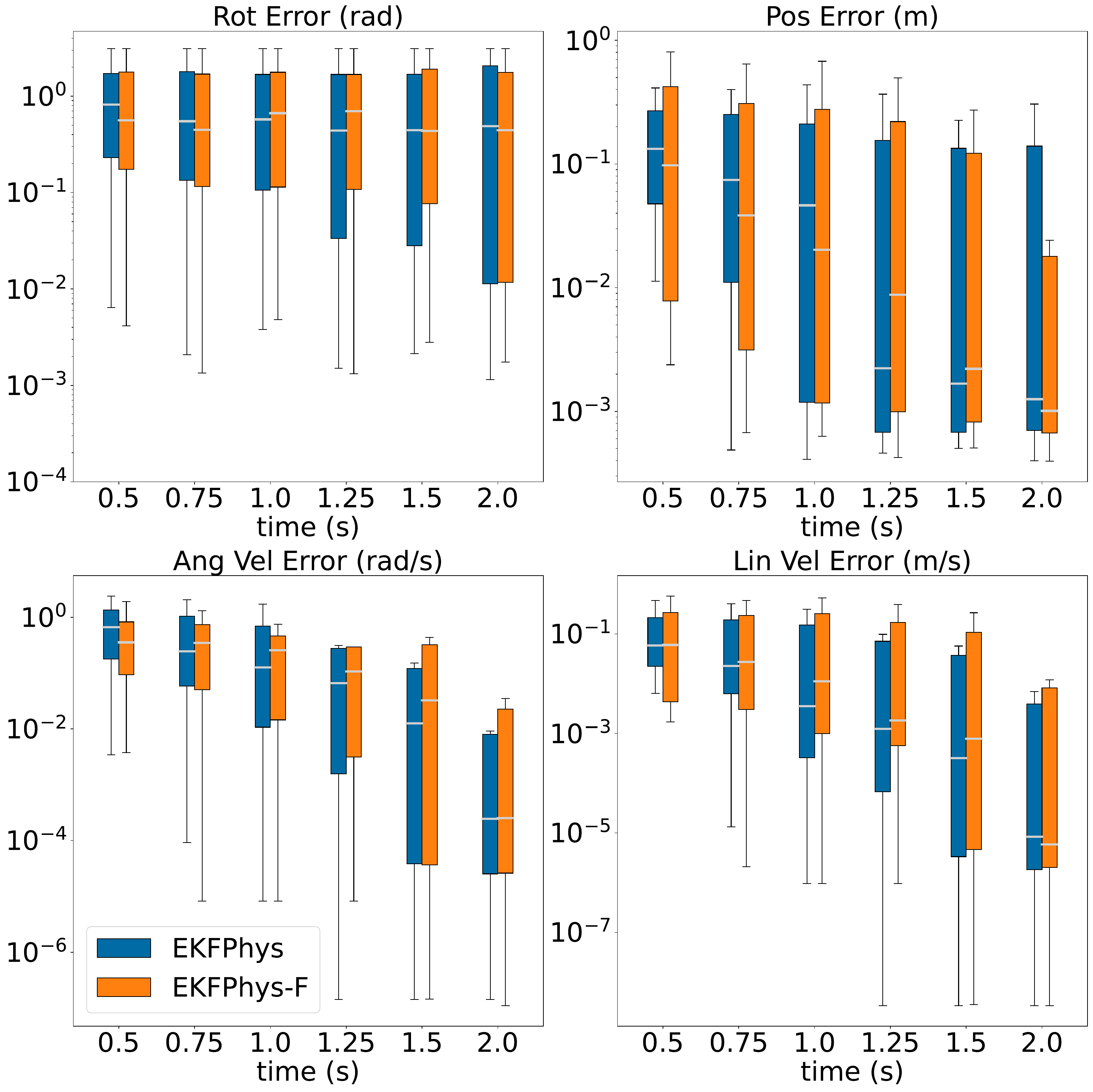}
    \caption{Filtering (left) and prediction (right) accuracy on single object sliding in synthetic sequences for symmetric objects. The box plot shows the median and quartiles of average rotation error (top-left), average position error (top-right), average angular velocity error (bottom-left) and average linear velocity error (bottom-right) of the trajectory over all the scenes in the dataset.}
    \label{fig:trajectory_error_filtering_prediction_single_obj_symm_synphys_non_damping}
\end{figure}

\begin{figure}
    \centering
    \includegraphics[width=.49\linewidth]{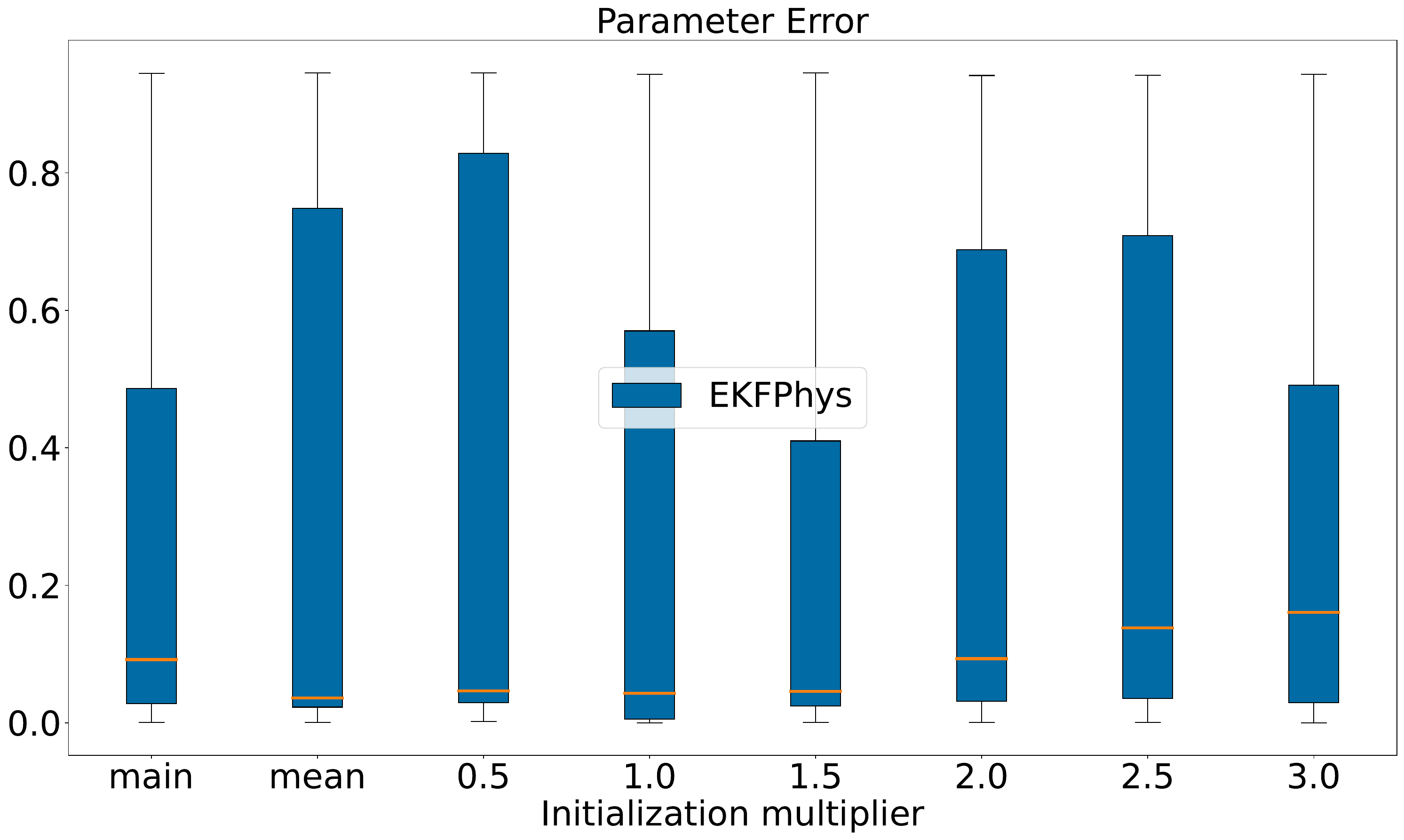}
    \caption{
    Friction coefficient error on single object sliding in synthetic sequences for symmetric objects. 
    The x-axis denotes the initialization multiplier, i.e., the friction value is initialized with factor * gt-friction. For "mean" we initialize with the average value of all the friction values in the validation dataset, which is 0.062 (synthetic) and 0.164 (real) in our case.
    The entry "main" indicates the value 0.0, at which our model EKFPhys is initialized for filtering and prediction experiments.
    The median errors increase as we move away from gt initialization, i.e., multiplier 1.0
    }
    \label{fig:friction_estimation_full_error_single_obj_symm_synphys_non_damping}
\end{figure}

In this section, we present the evaluation, i.e., filtering and prediction (Fig. \ref{fig:trajectory_error_filtering_prediction_single_obj_symm_synphys_non_damping}) and friction analysis (Fig. \ref{fig:friction_estimation_full_error_single_obj_symm_synphys_non_damping}) on symmetric objects on the synthetic dataset.

\section{Qualitative Results}

In this section, we provide qualitative results of prediction in synthetic (Fig. \ref{fig:pudding box syn prediction} - \ref{fig:Mug - tuna fish can syn prediction}) and real-world (Fig. \ref{fig:pitcher real prediction} - \ref{fig:mustard bottle real prediction}) scenes.

\subsection{Single Object: Synthetic}

\begin{figure}[tbh]
    \centering
    \includegraphics[width=0.99\linewidth]{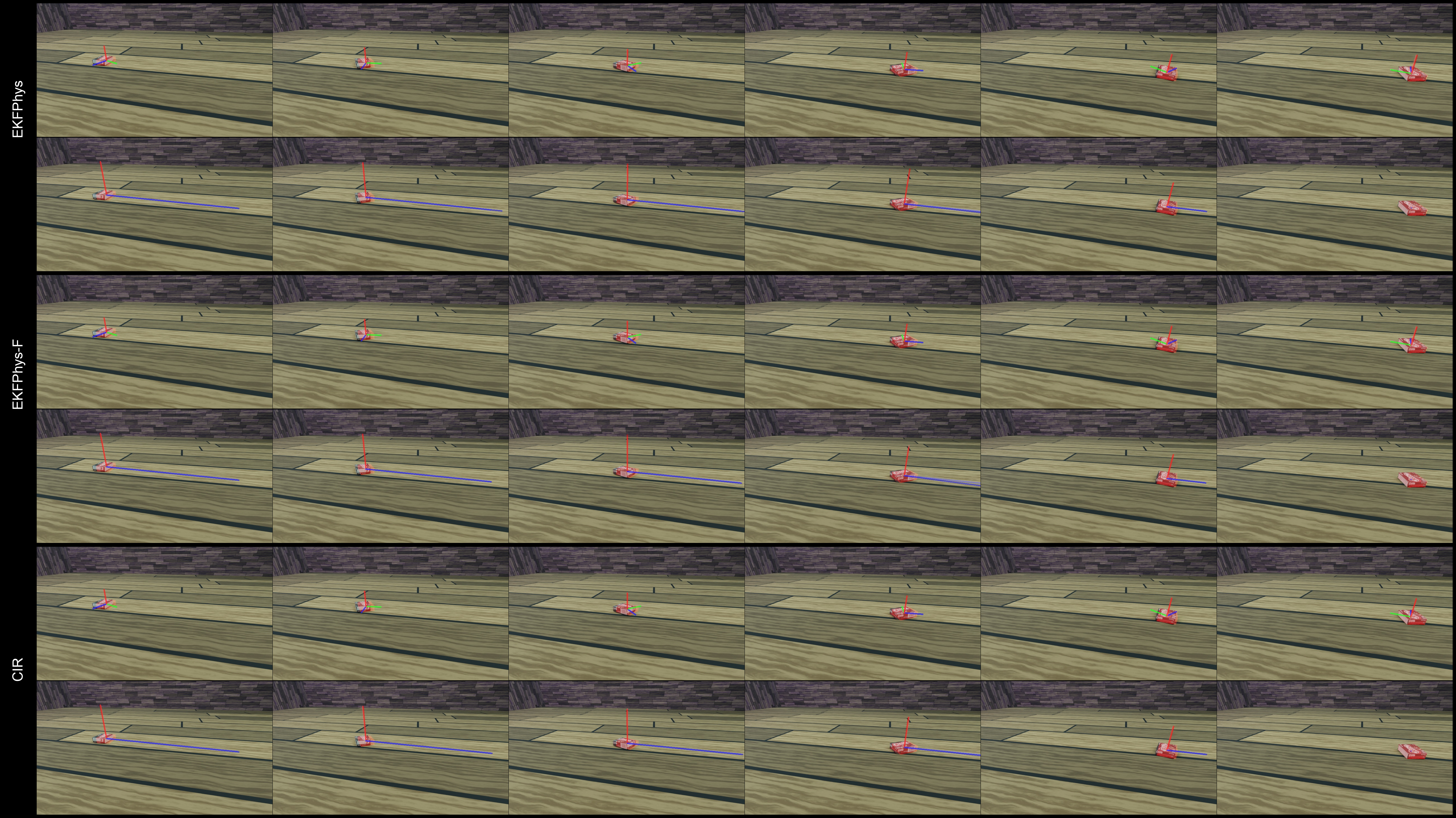}
    \caption{Prediction (Synthetic): Pudding box sliding scene. Left 4 images: filtering. Right 2 images: prediction. Poses and velocities are predicted with good accuracy.}
    \label{fig:pudding box syn prediction}
\end{figure}

\begin{figure}[tbh]
    \centering
    \includegraphics[width=0.99\linewidth]{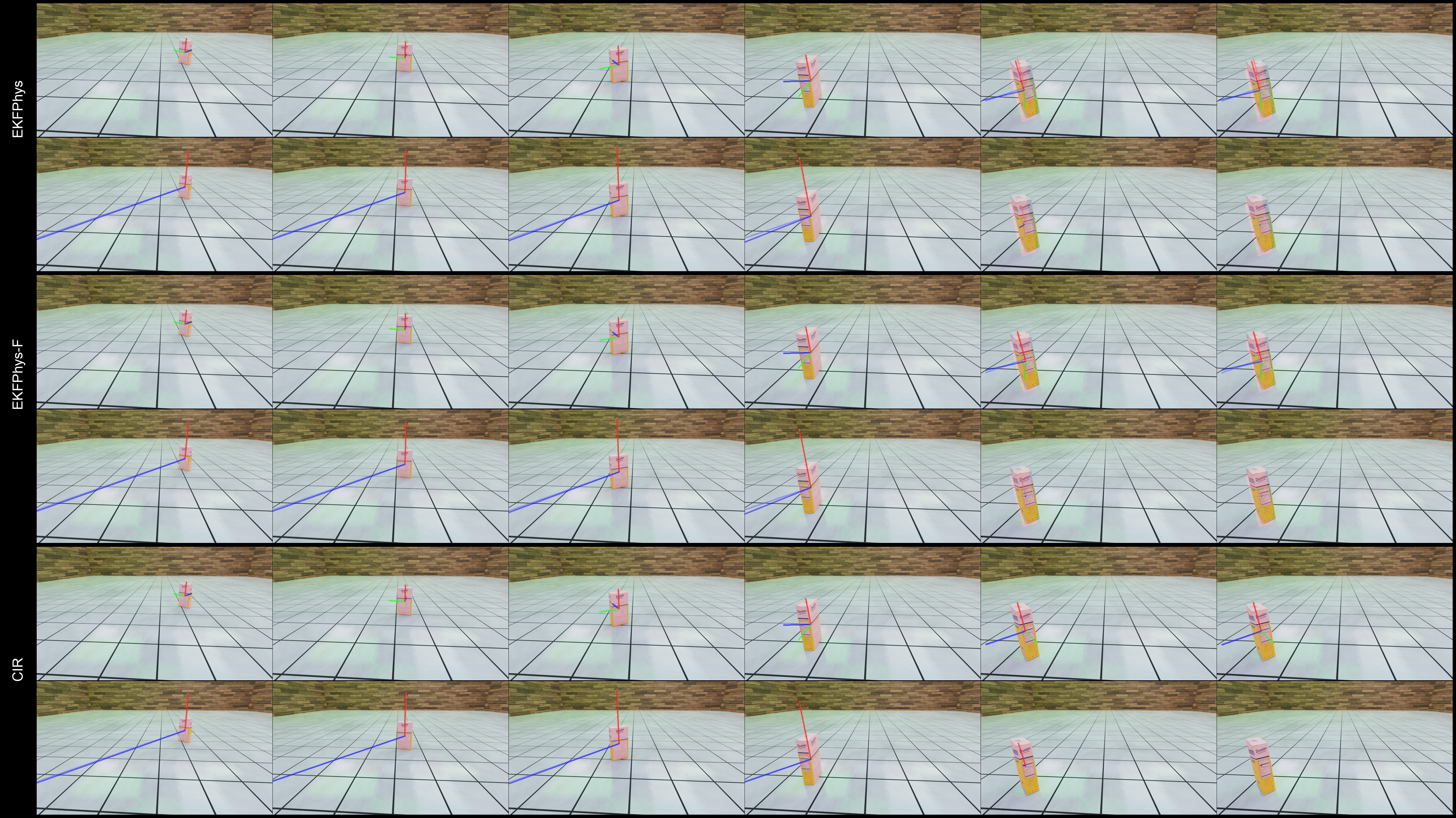}
    \caption{Prediction (Synthetic): Sugar box sliding scene. Left 3 images: filtering. Right 3 images: prediction. Poses and velocities are predicted with good accuracy except for a small error in z-rotation (red axis).}
    \label{fig:sugar box syn prediction}
\end{figure}

\begin{figure}[tbh]
    \centering
    \includegraphics[width=0.99\linewidth]{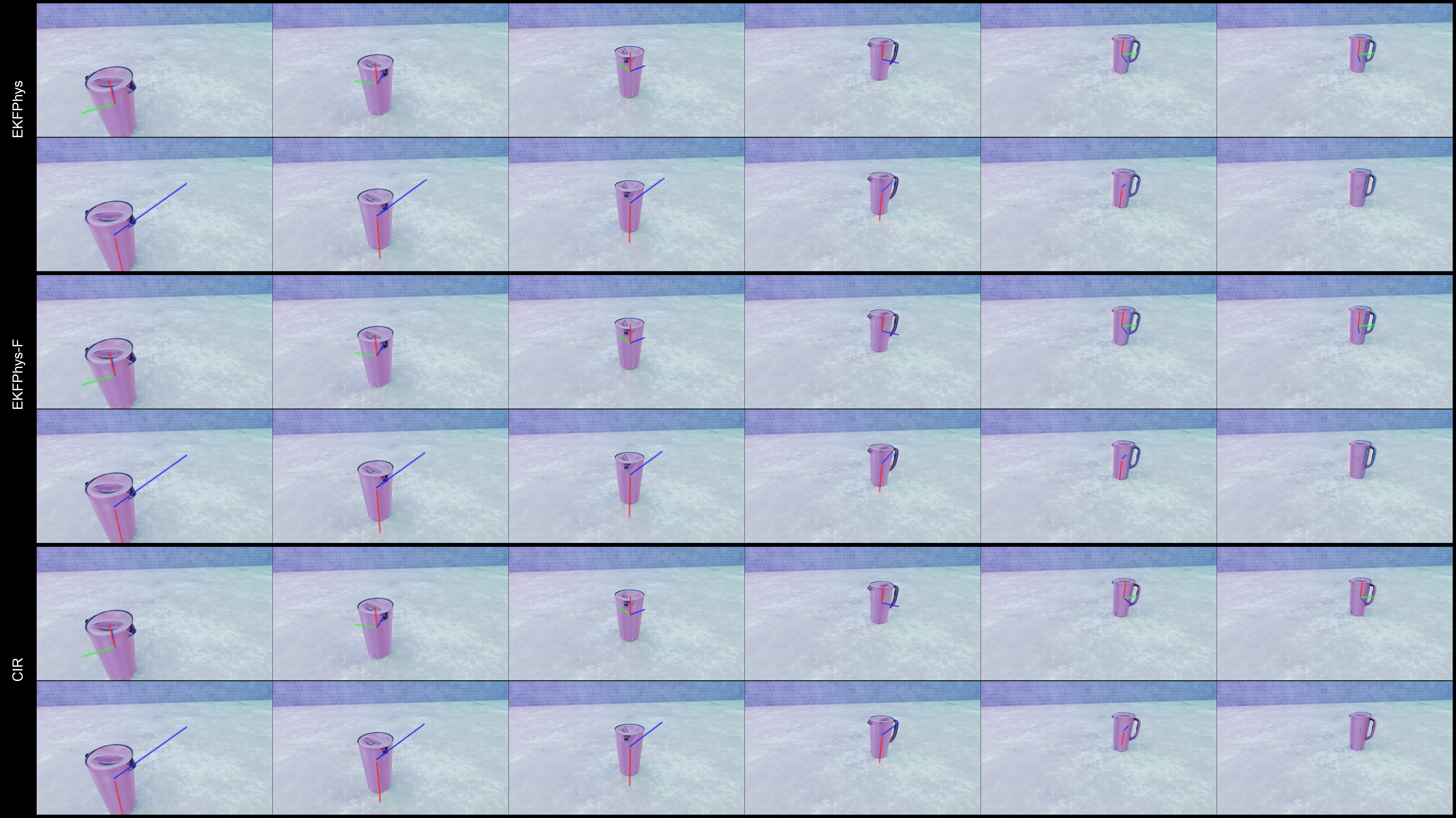}
    \caption{Prediction (Synthetic): Pitcher sliding scene. Left 3 images: filtering. Right 3 images: prediction. Poses and velocities are predicted with good accuracy except for a small error in z-rotation (red axis).}
    \label{fig:pitcher syn prediction}
\end{figure}

\begin{figure}[tbh]
    \centering
    \includegraphics[width=0.99\linewidth]{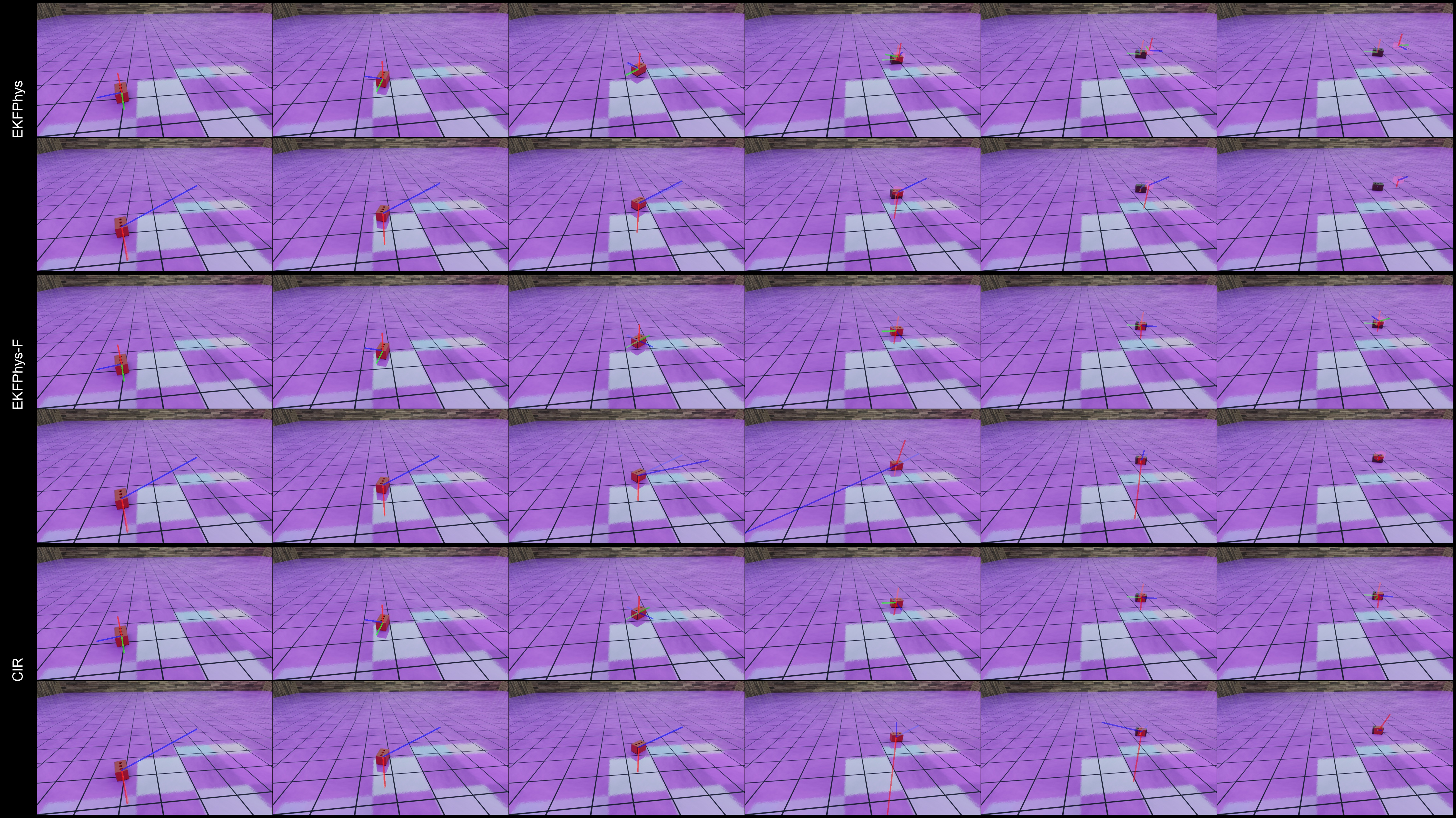}
    \caption{Failed prediction (Synthetic): Foam brick sliding scene. Left 4 images: filtering. Right 2 images: prediction. The rotation estimates from CIR got extremely noisy quite early in the trajectory (starting from frame 20 of 75) and because of that EKFPhys rejected the observations with gating. Since only 20 frames were observed well from the beginning, friction did not converge to the right value. This makes EKFPhys predict trajectories with low friction and thus moves the body far away.}
    \label{fig:foam brick syn prediction}
\end{figure}

Fig. \ref{fig:pudding box syn prediction},  \ref{fig:sugar box syn prediction},  \ref{fig:pitcher syn prediction} and \ref{fig:foam brick syn prediction} (failure case), show the prediction scenes for single object sliding scenes on synthetic dataset.

\subsection{Two Objects: Synthetic}

\begin{figure}[tbh]
    \centering
    \includegraphics[width=0.99\linewidth]{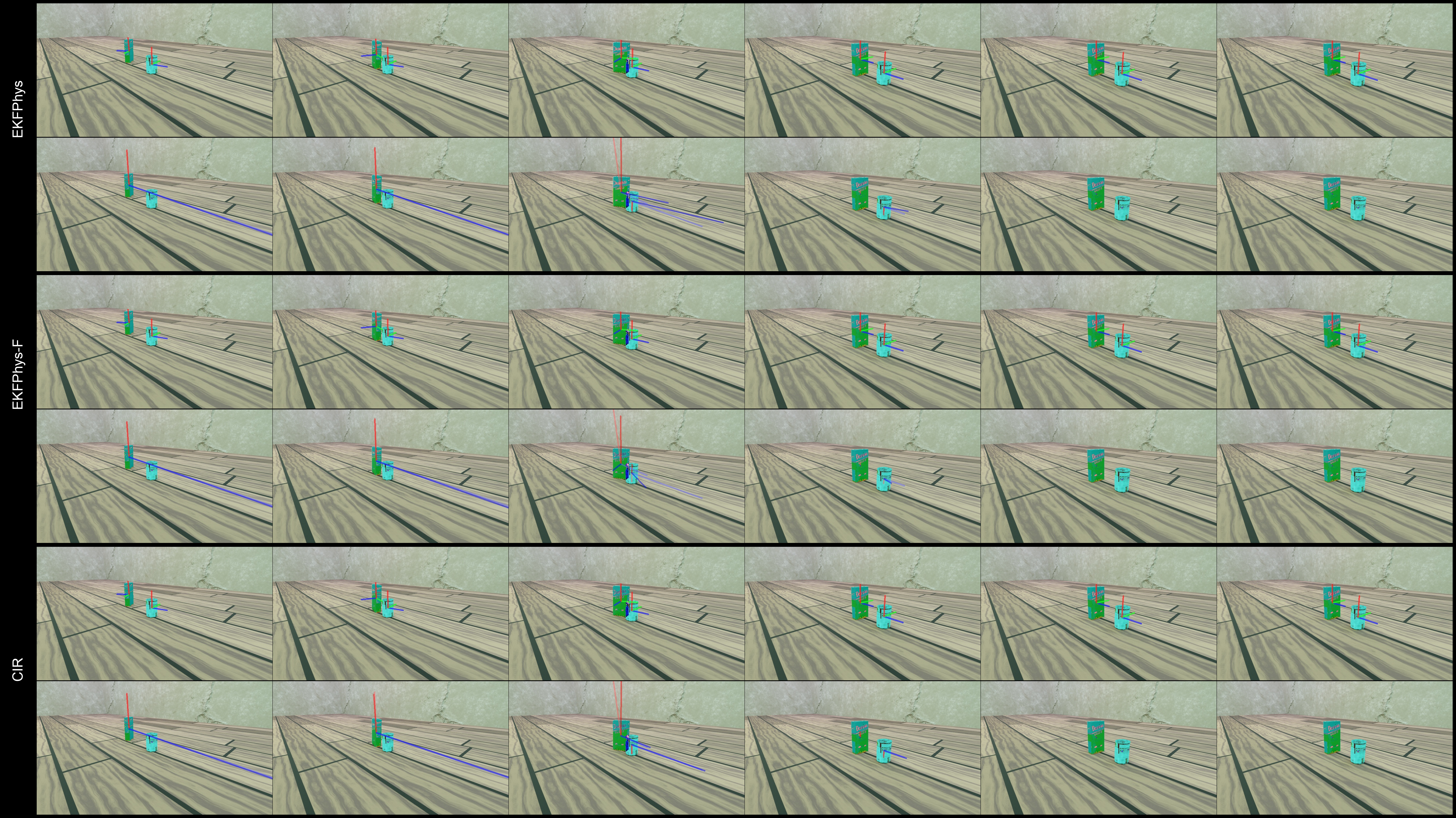}
    \caption{Prediction (Synthetic): Tomato soup can and sugar box sliding and colliding with each other. Left 4 images: filtering. Right 2 images: prediction.
    For each variant: upper row: estimated (saturated colored axes) and ground-truth (light colored axes) poses; lower row: estimated velocities (blue linear, red angular) and ground-truth velocities (light colors). 
    Both the objects are observed in each frame. Poses and velocities are predicted with good accuracy.}
    \label{fig:tomato soup - sugar box syn prediction}
\end{figure}

\begin{figure}[tbh]
    \centering
    \includegraphics[width=0.99\linewidth]{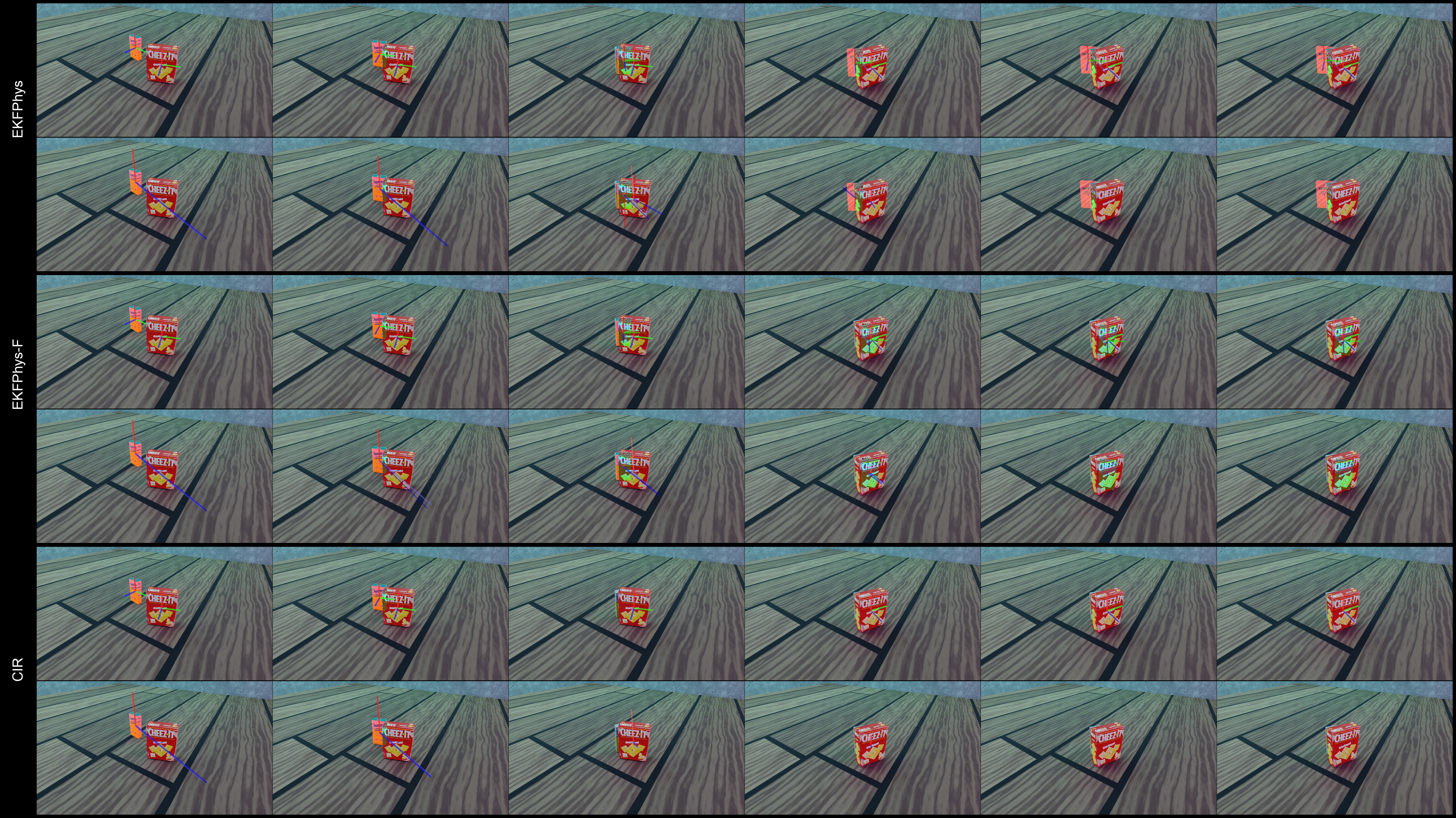}
    \caption{Prediction (Synthetic): Cracker box and sugar box sliding and colliding with each other. Left 4 images: filtering. Right 2 images: prediction.
    For each variant: upper row: estimated (saturated colored axes) and ground-truth (light colored axes) poses; lower row: estimated velocities (blue linear, red angular) and ground-truth velocities (light colors). 
    In this scene, CIR loses pose estimate of sugar box from the fourth image from the left (before prediction). Since the observations are lost quite early, EKFPhys doesn't filter the friction and thus wrongly estimates the final position and rotation of the sugar box although it models the collision correctly. Here the variant EKFPhys-F which has gt friction information predicts positions and rotations with good accuracy.}
    \label{fig:cracker box - sugar box syn prediction}
\end{figure}

\begin{figure}[tbh]
    \centering
    \includegraphics[width=0.99\linewidth]{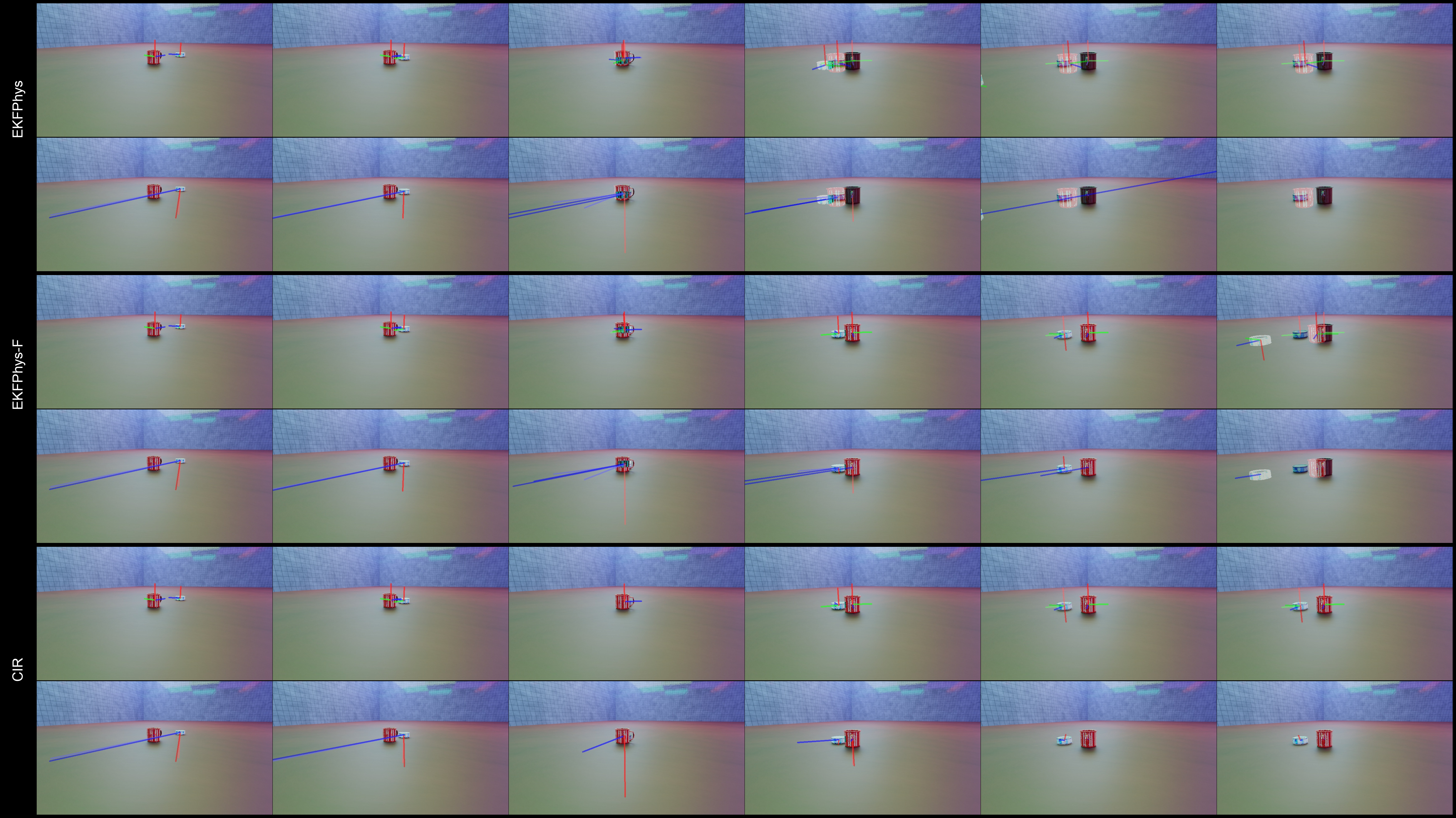}
    \caption{Failure Prediction (Synthetic): Tuna fish can and mug sliding and colliding with each other. Left 4 images: filtering. Right 2 images: prediction.
    For each variant: upper row: estimated (saturated colored axes) and ground-truth (light colored axes) poses; lower row: estimated velocities (blue linear, red angular) and ground-truth velocities (light colors). 
    The tuna fish can is observed in the initial frames by CIR and it gets quickly occluded by the mug. The EKFPhys couldn't observe enough frames to filter friction and thus moves the tuna  can at a much higher velocity and leaves the scene in frame 4. Tuna can has collided with the mug and displaced it much further with this velocity as the prediction phase starts. The estimate of EKFPhys-F is also quite off because just before CIR loses tuna can and soon after it reappears, it gives out an erroneous rotation estimate which throws the filter off.}
    \label{fig:Mug - tuna fish can syn prediction}
\end{figure}

Fig. \ref{fig:tomato soup - sugar box syn prediction}, \ref{fig:cracker box - sugar box syn prediction} and \ref{fig:Mug - tuna fish can syn prediction} (failure case), show the prediction scenes for two objects sliding and colliding scenes on synthetic dataset.

\subsection{Single Object: Real}
\begin{figure}[tbh]
    \centering
    \includegraphics[width=0.99\linewidth]{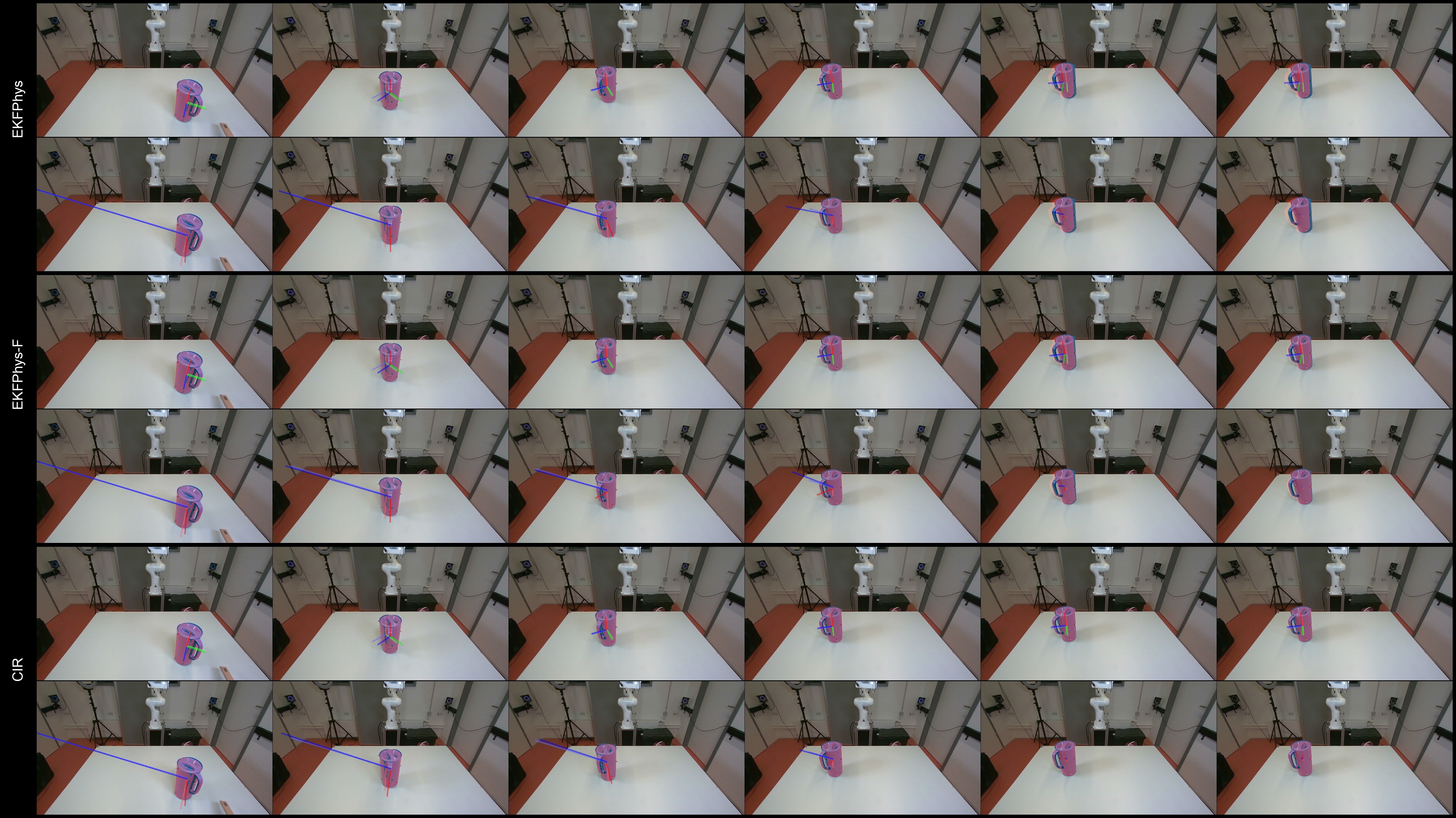}
    \caption{Prediction (Real): Pitcher sliding scene. Left 4 images: filtering. Right 2 images: prediction.
    For each variant: upper row: estimated (saturated colored axes) and ground-truth (light colored axes) poses; lower row: estimated velocities (blue linear, red angular) and ground-truth velocities (light colors). Poses and velocities are predicted close to the ground truth. EKFPhys-F predicts the poses and velocities with good accuracy and EKFPhys slightly overshoots in pose. Note that the sequence is short and the friction coefficient has been filtered slightly (0.209) below gt (0.220) before the prediction phase.}
    \label{fig:pitcher real prediction}
\end{figure}

\begin{figure}[tbh]
    \centering
    \includegraphics[width=0.99\linewidth]{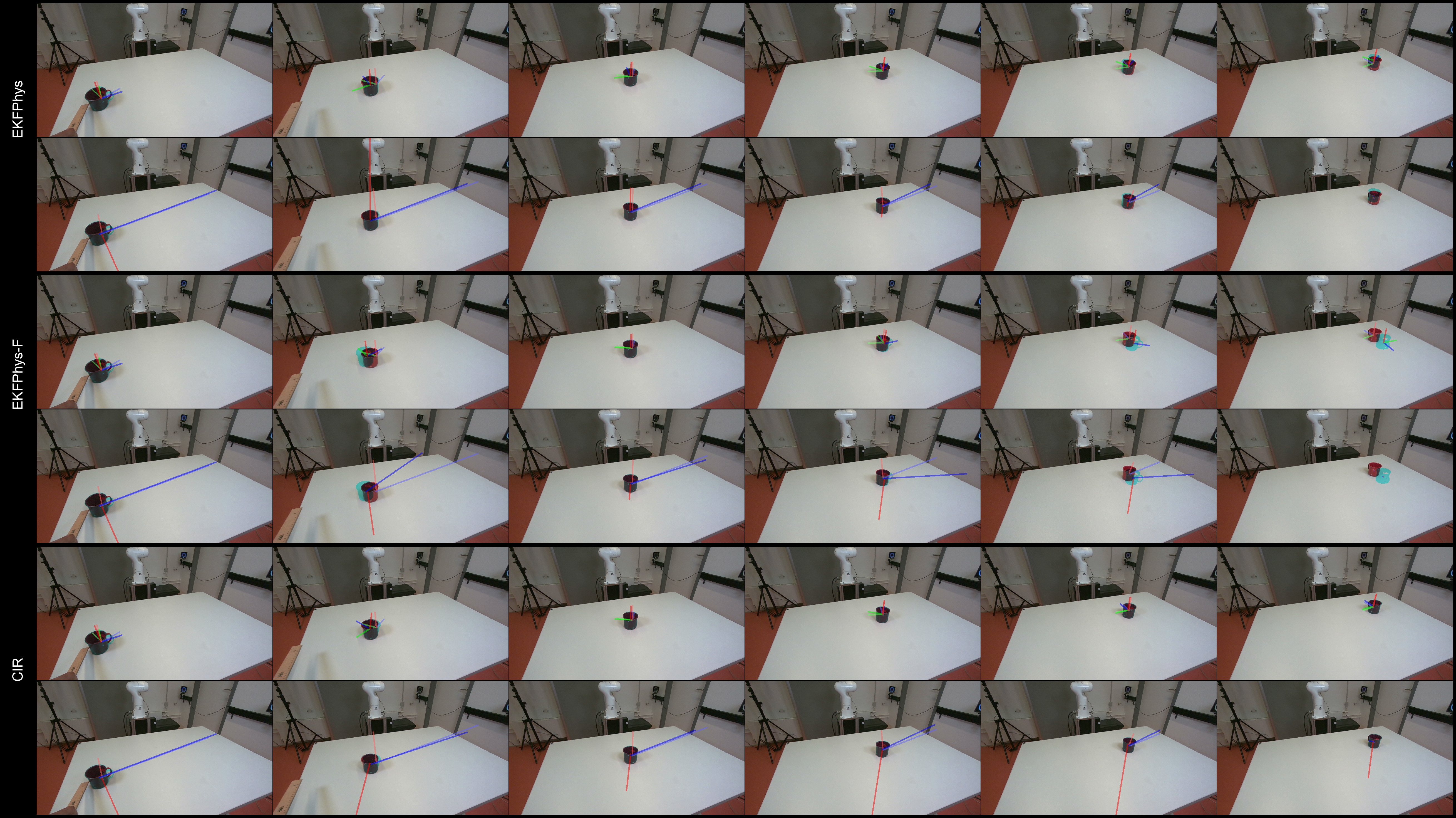}
    \caption{Prediction (Real): Mug sliding scene. Left 4 images: filtering. Right 2 images: prediction.
    For each variant: upper row: estimated (saturated colored axes) and ground-truth (light colored axes) poses; lower row: estimated velocities (blue linear, red angular) and ground-truth velocities (light colors). Poses and velocities are predicted close to the ground truth. The rotation estimates from CIR are quite noisy. Since EKFPhys-F has low observation covariance in rotations, it confidently uses those measurements which brings the object into penetration and thus filters incorrect velocities. When the prediction starts, it predicts the pose of the object incorrectly. EKFPhys on the other hand has high observation covariance in rotations which makes it smooth out the rotations and filter velocities more accurately. Also, since the simulator is an approximate model for the real world, even with gt friction coefficient information, there will be a deviation in the final estimate.}
    \label{fig:mug real prediction}
\end{figure}

\begin{figure}[tbh]
    \centering
    \includegraphics[width=0.99\linewidth]{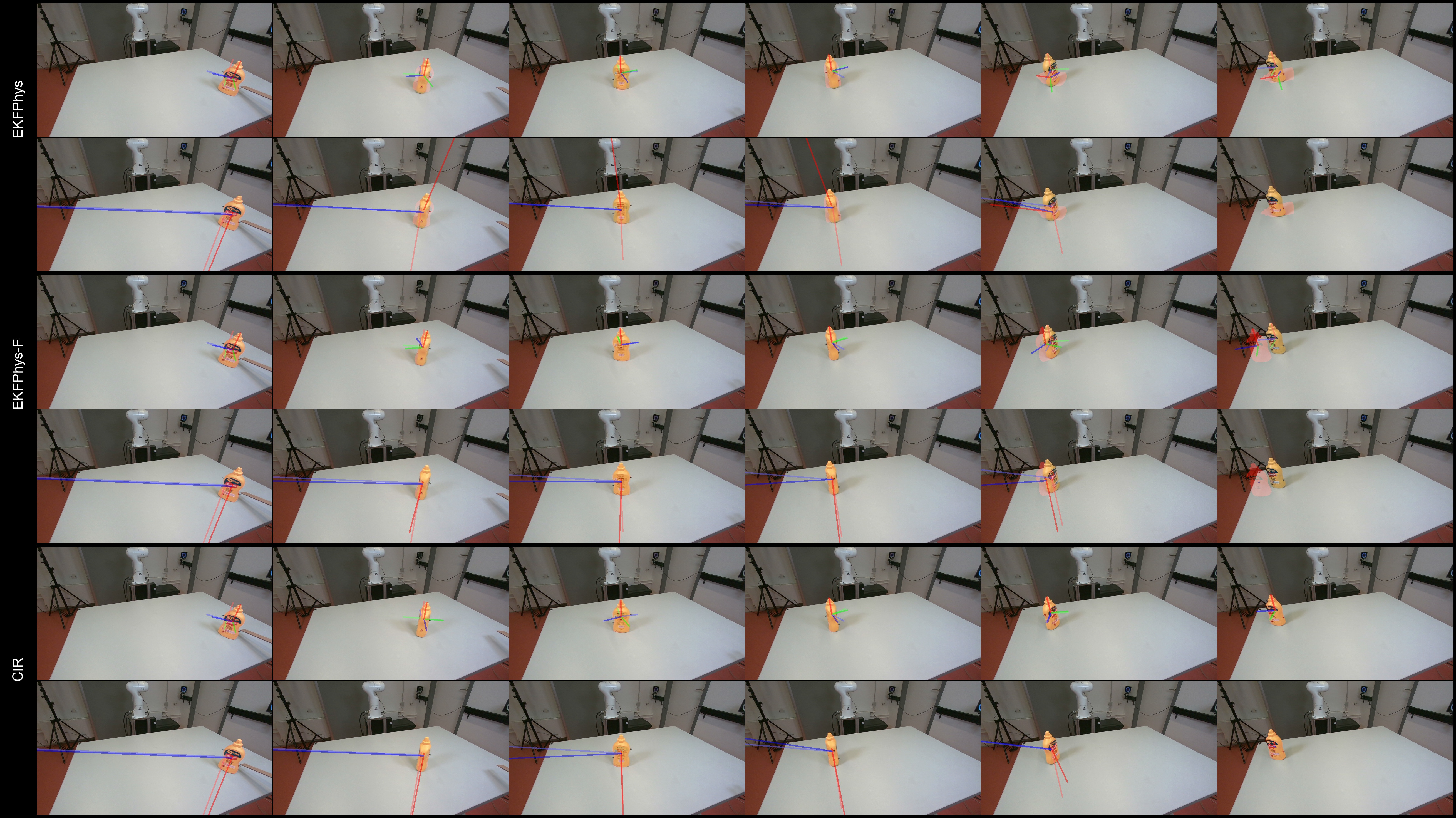}
    \caption{Failed Prediction (Real): Mustard bottle sliding scene. Left 4 images: filtering. Right 2 images: prediction.
    For each variant: upper row: estimated (saturated colored axes) and ground-truth (light colored axes) poses; lower row: estimated velocities (blue linear, red angular) and ground-truth velocities (light colors). The rotation estimates from CIR are quite noisy and has erroneous detections in rotation just before the prediction starts. This makes EKFPhys topple the mustard bottle before prediction and it never recovers. Because of these erroneous estimates EKFPhys-F rejects measurements before prediction and thus doesn't accurately filter the velocities. Also, since the simulator is an approximate model for the real world, even with gt friction coefficient information, there will be a deviation in the final estimate.}
    \label{fig:mustard bottle real prediction}
\end{figure}

Fig. \ref{fig:pitcher real prediction}, \ref{fig:mug real prediction} and \ref{fig:mustard bottle real prediction} (failure case), show the prediction scenes for single object sliding scenes on real-world dataset.

\end{document}